\documentclass{article}
\usepackage{kore,times}

\usepackage{amsmath,amssymb,amsfonts,amsthm,bm,mathtools}
\usepackage{graphicx}
\usepackage{booktabs}
\usepackage{multirow}
\usepackage{algorithm}
\usepackage{algpseudocode}
\usepackage{float}
\usepackage[section]{placeins}
\usepackage{microtype}
\usepackage{xcolor}
\definecolor{korelink}{HTML}{1A4E8A}
\usepackage{url}
\usepackage[colorlinks=true,linkcolor=korelink,citecolor=korelink,urlcolor=korelink]{hyperref}
\usepackage{tabularx}
\usepackage{array}
\usepackage{siunitx}
\usepackage[font=small,labelfont=bf,labelsep=period,justification=justified,singlelinecheck=false]{caption}

\DeclareMathOperator*{\argmin}{arg\,min}

\newcommand{\EE}{\mathbb{E}}
\newcommand{\Err}{\mathrm{Err}}

\newcommand{\KORE}{\textsc{KORE}}
\newcommand{\add}{\mathrm{add}}
\newcommand{\pair}{\mathrm{pair}}
\newcommand{\LoO}{\mathrm{LOO}}

\newtheorem{proposition}{Proposition}
\newtheorem{theorem}{Theorem}
\newtheorem{corollary}{Corollary}
\newtheorem{remark}{Remark}
\newtheorem{lemma}{Lemma}

\korefinalcopy

\title{{\fontsize{16pt}{19pt}\selectfont Solve for the Hyperparameter, Skip the Search:} \\Kolmogorov-Optimal Scaling Laws for Spline Regression}

\author{
Yong Yi Bay\thanks{Equal contribution. Correspondence: \texttt{\{yongyibay, kallie.a.yearick\}@gmail.com}.} \hspace{1cm} Kathleen A.\ Yearick\footnotemark[1] \\[3pt]
{\normalfont PhD, University of Illinois at Urbana-Champaign}
}

\begin{document}

\maketitle

\begin{abstract}
Hyperparameter tuning almost always means search: fit the model at every value on a grid, score each by cross-validation, and keep the winner. For spline regression that search is unnecessary. The optimal resolution can be \emph{solved for} in closed form, to the accuracy an exhaustive search would reach and at a fraction of the compute. Three facts make this possible. Classical approximation theory pins the squared bias to $G^{-2\beta}$ in the resolution $G$, which is exactly the Kolmogorov $n$-width of the smoothness class, so the spline family is approximation-optimal among linear methods of its size; the basis dimension is an explicit polynomial in $G$; and the leave-one-out (LOO) error of any linear smoother follows from a single fit through the PRESS identity. Balancing the two known curves gives the minimizer analytically. We carry the same calculus from one coordinate to many by replacing ambient input dimension with \emph{interaction order}, the number of active low-order components in an ANOVA decomposition, and obtain a scaling law in which the optimal resolution and the optimal error are power functions of the effective density $n / s_r$, with the input dimension absent from the exponent. The law becomes an algorithm. \KORE{} (\textbf{K}olmogorov-optimal \textbf{O}rder-aware \textbf{R}esolution \textbf{E}stimation) fits two pilot resolutions, solves a leverage-calibrated $2\times2$ system for the bias and noise/variance scales, and evaluates the closed-form plug-in resolution with a tiny LOO certificate: a fixed dozen fits in place of a full grid sweep, with a consistency guarantee as $n$ grows. Across additive and sparse pairwise targets up to $80$ input dimensions, \KORE{} matches exhaustive $3$-fold cross-validation and the entire classical full-grid ladder of generalized cross-validation, Mallows' $C_p$, AIC, and BIC in accuracy while fitting roughly $8\times$ fewer models; on $36$ real tabular datasets it ranks first among $21$ methods in accuracy delivered per unit of compute, ahead of tuned gradient boosters and kernel machines. When a target's complexity lives in low interaction order, solving for the resolution beats searching for it.
\end{abstract}

\section{Introduction}
\label{sec:intro}

Hyperparameter selection is usually done by search: a model is trained at every candidate setting, validation error is recorded, and the best score wins. Cross-validation, the canonical realization of that workflow, is reliable but uninformative; it treats every hyperparameter as an opaque knob and offers no closed-form guidance for which values are worth trying or why one value beats its neighbors. The compute cost scales with the size of the grid multiplied by the number of folds and grows with every new model family. An attractive alternative is to \emph{solve} for the optimal hyperparameter directly, in the same sense that the minimum of a known function is found by calculus rather than by tabulation.

\paragraph{Why spline regression.}
Solving rather than searching requires a model class where prediction error has an analytically tractable dependence on the hyperparameter of interest. Three ingredients are needed: (i)~an approximation theory that specifies how bias scales with the hyperparameter, (ii)~an explicit formula for the number of free parameters, which controls variance, and (iii)~a closed-form estimate of prediction error, so the bias-variance curve can be pinned down from a small number of fits rather than from an exhaustive grid. Spline regression is the natural home for this program. Classical B-spline approximation theory gives precise bias rates as a power of the knot resolution $G$ \citep{deboor2001,schumaker2007}; these rates are exactly the Kolmogorov $n$-widths of the smoothness class, so among all linear methods of the same size the spline family is approximation-optimal \citep{kolmogorov1936,melkman1978,pinkus1985}. The basis dimension is an explicit, countable polynomial in $G$. Because a spline fit is a linear smoother, exact leave-one-out error is available at $O(1)$ per point through the PRESS identity \citep{allen1974} with no refitting. Together, these three properties express the entire bias-variance curve as a closed-form function of a single integer, and they make solving for the minimum a calculus problem rather than a search problem.

Other popular model classes lack this combination. Neural-network hyperparameters (width, depth, learning rate) interact through a non-convex loss landscape, and no finite-sample power law links architecture to approximation error; the strength of deep models lies instead in extrapolating beyond the training support, a regime where they can outperform classical smoothers \citep{bay2024generalization}. Tree ensembles (random forests, gradient boosting) have multiple interacting knobs whose joint effect on bias is data-dependent and resists closed-form treatment. Kernel methods come closest: bandwidth controls a bias-variance tradeoff, but the effective dimensionality of the smoothing problem is implicit in the kernel's eigenspectrum and cannot be read off a design matrix. In each of these cases search is the only available option. Splines are the mature classical model family in which solving is tractable, and this paper studies how far that solvability extends into multiple coordinates.

In spline regression, the key hyperparameter is the resolution $G$, the number of knot intervals per coordinate. It controls model capacity directly: too few intervals and the spline is too stiff to follow the true signal; too many and it starts fitting noise. For a grid of $20$ candidate additive resolutions and $10$ candidate sparse pairwise resolutions with $3$-fold CV, exhaustive search costs $(20+10)\times 3 + 1 = 91$ model fits in total, plus a final refit on the full data. When multiple families are compared the cost multiplies. \KORE{} replaces that entire search with a constant number of fits by deriving a closed-form scaling law for the optimal resolution $G^\star$ and fitting at just two pilot resolutions to identify the law's unknown constants. The question this paper answers is whether $G^\star$ can be predicted instead of searched. The scope is also explicit: the contribution is resolution selection \emph{once a structured spline family has been chosen}. Penalized generalized additive models tune a continuous roughness penalty for a fixed basis by criteria such as GCV, AIC, or REML \citep{wood2017}. \KORE{} targets the complementary discrete question of selecting the right basis resolution when comparing low-order spline dictionaries. The closed-form selector applies inside the spline-ANOVA function class with bounded post-one-hot dimension; on signals dominated by high-order interactions or by deep categorical structure, tuned boosters retain the upper hand, and Section~\ref{sec:failure_modes} delineates the boundary explicitly.

\paragraph{The bias-variance tradeoff in resolution.}
Selecting the best resolution requires a quantitative description of how prediction error depends on $G$. Write $\Err(G)$ for the expected squared error of a spline predictor fitted at resolution $G$ on a new test point. This error decomposes as
\begin{equation}
    \Err(G) \;-\; \sigma^{2}
    \;=\;
    \underbrace{A\, G^{-2\beta}\vphantom{\dfrac{p(G)}{n}}}_{\displaystyle\text{bias}^{2}}
    \;+\;
    \underbrace{\tau\;\dfrac{p(G)}{n}}_{\displaystyle\text{variance}},
\end{equation}
where
\begin{itemize}
    \item $\sigma^{2}$ is the irreducible noise in the data, which no model can remove;
    \item $G$ is the resolution, the number of knot intervals per coordinate;
    \item $\beta$ is the smoothness exponent of the target function, set by the spline degree $k$;
    \item $p(G)$ is the basis size, the number of spline basis functions at resolution $G$;
    \item $n$ is the sample size;
    \item $A > 0$ is the bias scale and $\tau > 0$ is the noise/variance scale.
\end{itemize}
The bias term $A\,G^{-2\beta}$ falls as $G$ grows because a finer grid approximates the true function more closely. The variance term $\tau\,p(G)/n$ rises as $G$ grows because more basis functions imply more parameters to estimate from $n$ noisy observations. The sum traces a U-shaped curve. The resolution at the bottom of that U is the desired one. The entire paper is built on making this decomposition precise for structured multidimensional splines and then solving for its minimizer.

\paragraph{One coordinate: a clean classical answer.}
In one coordinate, $G$ is the only knob and the answer is closed-form. A spline with $G$ knot intervals and degree $k$ has $G + k$ basis functions. The squared approximation bias decays as $G^{-2\beta}$ with $\beta$ the smoothness of the target. The variance grows as $(G + k) / n$, since each basis function adds one degree of freedom to be estimated from $n$ data points. Balancing the two forces gives the optimal resolution:
\[
\underbrace{A\, G^{-2\beta}\vphantom{\dfrac{G + k}{n}}}_{\substack{\text{bias}^2 \\ \text{(falling)}}}
\;=\;
\underbrace{\tau\;\dfrac{G + k}{n}}_{\substack{\text{variance} \\ \text{(rising)}}}
\qquad\Longrightarrow\qquad
G^\star \;\propto\; n^{1 / (2\beta + 1)}.
\]
The exponent depends only on smoothness. The two constants $A$ (bias scale) and $\tau$ (noise/variance scale) can be identified from fits at two different resolutions. For a single coordinate, the closed-form solution replaces search entirely.

\paragraph{Multiple coordinates: the naive tensor product fails.}
The naive multivariate spline is the tensor product, the Cartesian product of $d$ univariate bases. It produces $(G + k)^d$ basis functions, a count that explodes even in moderate dimension:
\smallskip
\begin{center}
\small
\begin{tabular}{@{}lccccc@{}}
\toprule
 & $d=1$ & $d=2$ & $d=5$ & $d=10$ & $d=20$ \\
\midrule
$(G+k)^d$ at $G=5,\,k=3$ & 8 & 64 & 32{,}768 & $\sim 10^9$ & $\sim 10^{18}$ \\
\bottomrule
\end{tabular}
\end{center}
\smallskip

\noindent With $10^9$ or more basis functions the variance term dominates, and any closed-form estimate of the optimum loses its practical edge over cross-validation. This is the curse of dimensionality. The escape is structural: most regression problems of practical interest do not need the full tensor product.

\paragraph{Low-order structure: the escape.}
Consider a function of ten coordinates that is purely additive,
\[
f(x_1, \dots, x_{10}) = f_1(x_1) + f_2(x_2) + \cdots + f_{10}(x_{10}).
\]
The natural spline basis is ten univariate blocks rather than a tensor product. The total basis size is $10(G + k)$ instead of $(G + k)^{10}$. At $G = 5$ with cubic splines that is $80$ basis functions instead of about a billion. A function with sparse pairwise interactions has univariate terms plus a few bivariate terms $f_{ij}(x_i, x_j)$. If $s$ pairs are active, the basis adds $s(G+k)^2$ tensor-product columns for those pairs, giving a total of roughly $d(G + k) + s(G + k)^2$. While $s$ is moderate, the basis remains tractable. In both cases the variance is governed by the number of active low-order components, not by the input dimension. Balancing bias against variance produces a different scaling law per structure family:
\smallskip
\begin{center}
\small
\begin{tabular}{@{}lcc@{}}
\toprule
Structure & Variance scales with & Optimal $G^\star$ scales as \\
\midrule
Full tensor product & $(G+k)^d / n$ & $(n)^{1/(2\beta+d)}$ \\[2pt]
Additive & $d(G+k) / n$ & $(n/d)^{1/(2\beta+1)}$ \\[2pt]
Sparse pairwise & $s(G+k)^2 / n$ & $(n/s)^{1/(2\beta+2)}$ \\
\bottomrule
\end{tabular}
\end{center}
\smallskip

\noindent For the full tensor product the exponent $1 / (2\beta + d)$ shrinks with $d$ and quickly becomes useless. For the additive family the exponent is $1 / (2\beta + 1)$, the same as in one coordinate. For sparse pairwise families it is $1 / (2\beta + 2)$. In neither case does the input dimension $d$ appear in the exponent. The quantity that matters is the effective density: $n / d$ for additive models, $n / s$ for sparse pairwise models with $s$ active pairs.

\paragraph{From law to algorithm.}
The effective-density law fixes the shape of the error curve. Its two unknown constants $A$ (bias scale) and $\tau$ (noise/variance scale) must still be identified from data. \KORE{} fits at two pilot resolutions per family. Each fit yields an exact LOO error through the PRESS identity, with no refitting. Two LOO measurements at two resolutions form a $2\times2$ leverage-calibrated linear system in $(A, \tau)$, solvable in closed form. The continuous plug-in resolution $\widehat{G}^\dagger$ is then the unique positive root of the derivative of the fitted excess-risk curve, and a small symmetric integer neighborhood around the rounded plug-in serves as a finite-sample certificate. The entire procedure costs about a dozen model fits. Exhaustive $3$-fold cross-validation on the candidate grids ($20$ additive resolutions and $10$ pairwise resolutions, three folds each, plus a final refit) costs $(20 + 10) \times 3 + 1 = 91$.

\begin{figure}[t]
\centering
\includegraphics[width=0.58\textwidth]{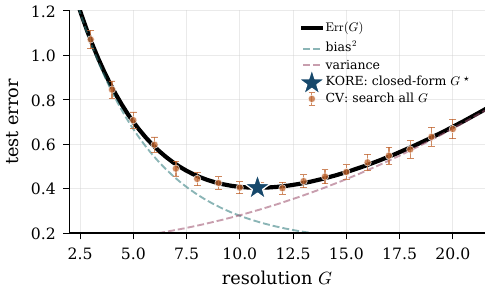}
\caption{Law-driven versus search-driven resolution selection. Cross-validation evaluates every grid candidate (clay dots). \KORE{} fits two pilot resolutions, identifies the bias and variance scale constants, and reports the closed-form optimum (navy star). The dashed curves show how squared bias (falling) and variance (rising) compose the U-shaped error curve.}
\label{fig:conceptual}
\end{figure}

\paragraph{Contributions.}
The paper makes four contributions.
\begin{enumerate}
    \item An intrinsic-order test-error law for structured spline regression (Proposition~\ref{prop:error_law}). The optimal resolution scales with the count of active highest-order interactions, not exponentially with input dimension (Theorem~\ref{thm:resolution_law}). The bias term of this law is the Kolmogorov $n$-width of the smoothness class, so the spline family is approximation-theoretically width-optimal and the selected resolution attains the Stone minimax rate (Remarks~\ref{rem:minimax} and~\ref{rem:kolmogorov_width}).
    \item A search-free plug-in algorithm, \KORE{}, that estimates $(A_f, \tau_f)$ from two pilot fits via a leverage-calibrated $2\times2$ system, evaluates the closed-form plug-in resolution $\widehat{G}_f^\dagger$, and certifies it with a small LOO neighborhood (Algorithm~\ref{alg:kore}).
    \item A consistency theorem for the plug-in (Theorem~\ref{thm:plugin_guarantee}): the estimated constants and the rounded plug-in converge to their population counterparts as $n$ grows, and the integer selector matches exhaustive risk minimization once the oracle margin exceeds the plug-in error. An empirical companion (Section~\ref{sec:consistency}) confirms the convergence at $d = 20$.
    \item An empirical accuracy-compute study up to $d = 80$. Across six controlled frontier tasks, \KORE{} matches the entire classical full-grid ladder of $3$-fold CV, GCV, Mallows' $C_p$, AIC, and BIC at $8.1\times$ fewer model fits, and the effective-density collapse holds at every dimension and density tested (Section~\ref{sec:collapse}, Section~\ref{sec:frontier}). On nine smooth named benchmarks the geometric-mean RMSE ratio against CV is $0.918$ at $8.7\times$ fewer fits (Section~\ref{sec:bench_main}).
\end{enumerate}
Code is available at \url{https://github.com/bay-yearick-lab/kore}.

\section{From bias-variance to search-free resolution laws}
\label{sec:law}

The title of the paper asks for more than a faster heuristic. It asks when a hyperparameter can be treated as a \emph{statistical estimand}: a quantity determined by the training distribution, the model family, and the sample size, rather than by a blind validation sweep. For spline regression, the estimand is the resolution
\begin{equation}
    G_f^\bullet \;:=\; \argmin_{G\in\mathcal{G}_f^{\mathrm{stab}}} \Bigl\{\Err_f(G)-\sigma^2\Bigr\},
    \label{eq:g_estimand}
\end{equation}
where $f$ denotes the chosen structured spline family and $\mathcal{G}_f^{\mathrm{stab}}$ is the set of resolutions whose design matrices are numerically estimable. Cross-validation estimates \eqref{eq:g_estimand} by evaluating every candidate. \KORE{} estimates the constants in the risk curve and then solves for \eqref{eq:g_estimand}. This section gives the mathematical backbone for that claim.

\subsection{Structured spline families}
\label{sec:setup}

The data are independent training samples $(x_i,y_i)$ with $x_i\in[0,1]^d$ and
\begin{equation}
    y_i = f(x_i)+\varepsilon_i,
    \qquad
    \EE[\varepsilon_i\mid x_i]=0,
    \qquad
    \EE[\varepsilon_i^2\mid x_i]=\sigma^2.
    \label{eq:data_model}
\end{equation}
The analysis is stated for random design with a density bounded above and below on $[0,1]^d$; the same formulas apply conditionally on a fixed design whenever the empirical Gram matrices of the spline bases have eigenvalues bounded away from zero and infinity on the stable range of $G$. This is the usual condition under which least-squares spline estimates behave like their population projections.

\paragraph{Low-order ANOVA structure.}
The escape from the curse of dimensionality is structural rather than numerical. The signal $f$ is assumed to decompose into centered low-order components,
\begin{equation}
    f(x) = f_0 + \sum_{t=1}^{r}\;\sum_{u\in\mathcal{U}_t} f_u(x_u),
    \qquad
    \int f_u(z_u)\,dz_j =0\quad\text{for every }j\in u,
    \label{eq:anova}
\end{equation}
where $u$ is a subset of coordinates, $x_u$ is the corresponding subvector, $\mathcal{U}_t$ is the set of active $t$-way components, $s_t=|\mathcal{U}_t|$, and $r$ is the highest active interaction order. The centering constraints make the decomposition identifiable: constants live in $f_0$, one-dimensional effects live in the main-effect blocks, and interactions are not allowed to re-create lower-order terms.

The two families used throughout the paper are special cases of \eqref{eq:anova}. Additive models have $r=1$ and $s_1=d$,
\[
    f(x)=f_0+\sum_{j=1}^d f_j(x_j),
\]
while sparse pairwise models have $r=2$ and an interaction graph $\mathcal{E}$ with $s=|\mathcal{E}|$ active edges,
\[
    f(x)=f_0+\sum_{j=1}^d f_j(x_j)+\sum_{(i,j)\in\mathcal{E}} f_{ij}(x_i,x_j).
\]

\paragraph{The resolution-indexed spaces.}
Fix spline degree $k$ and let $G$ be the number of knot intervals per coordinate. A univariate B-spline block has $G+k$ columns before centering and
\begin{equation}
    m(G)=G+k-1
    \label{eq:mG}
\end{equation}
centered columns after the constant direction is removed. An additive design therefore has one intercept plus $d$ centered univariate blocks,
\[
    B_{\add}(G)=\bigl[\mathbf{1}\mid B_1\mid\cdots\mid B_d\bigr],
    \qquad
    p_{\add}(G)=1+d\,m(G)=d(G+k)-(d-1).
\]
A sparse pairwise design adds one row-wise tensor block for each active edge,
\[
    B_{\pair}(G)=\bigl[\mathbf{1}\mid B_1\mid\cdots\mid B_d\mid (B_i\otimes B_j)_{(i,j)\in\mathcal{E}}\bigr],
    \qquad
    p_{\pair}(G)=1+d\,m(G)+s\,m(G)^2.
\]
In general, if the highest active interaction order is $r$, the structured basis dimension is
\begin{equation}
    p_r(G)=1+\sum_{t=1}^{r}s_t\,m(G)^t.
    \label{eq:prg}
\end{equation}
This count is the first key mathematical fact: the ambient dimension $d$ enters only through the number of active components $s_t$. The dense tensor-product count $(G+k)^d$ has disappeared.

\subsection{The risk curve}
\label{sec:closed_form}

Let $\mathcal{S}_r(G)$ be the structured spline space with dimension $p_r(G)$, and let $\hat f_G$ be the ridge-stabilized least-squares fit in that space, with ridge small enough that it only regularizes the linear solve. The expected test MSE is
\begin{equation}
    \Err(G)=\EE_{(X,Y),\mathcal{D}}\!\left[(Y-\hat f_G(X))^2\right].
    \label{eq:test_err}
\end{equation}
Since $Y=f(X)+\varepsilon$ and the test noise is independent of the training data,
\begin{equation}
    \Err(G)=\sigma^2+\EE_{\mathcal{D}}\!\left[\|\hat f_G-f\|_{L_2(P_X)}^2\right].
    \label{eq:bvd}
\end{equation}
The selection-relevant part is therefore
\begin{equation}
    \widetilde{\Err}(G):=\Err(G)-\sigma^2.
    \label{eq:selection_err}
\end{equation}
The irreducible floor $\sigma^2$ matters for estimating the curve from data, but it does not affect the minimizer once the excess curve is known.

\paragraph{Bias: approximation by a resolution-$G$ space.}
Let $\Pi_G f$ be the $L_2(P_X)$ projection of $f$ onto $\mathcal{S}_r(G)$. Assume each active component $f_u$ has smoothness $\beta\le k+1$ on its $|u|$-dimensional domain. Tensor-product B-spline approximation theory gives
\begin{equation}
    \|f_u-\Pi_{G,u}f_u\|_{L_2}^2 \le C_{u,k,\beta}\,G^{-2\beta},
    \label{eq:component_bias}
\end{equation}
because the mesh width in every active coordinate is $h=1/G$ \citep{deboor2001,schumaker2007,stone1982}. Summing over finitely many active components yields
\begin{equation}
    \|f-\Pi_G f\|_{L_2(P_X)}^2 \le A_r\,G^{-2\beta},
    \label{eq:bias_bound}
\end{equation}
where $A_r>0$ is a bias-scale constant determined by the component functions and the design distribution, but not by $G$ or $n$. If the leading approximation term is nonzero, \eqref{eq:bias_bound} is sharp up to lower-order terms.

\paragraph{The bias rate is a Kolmogorov $n$-width.}
The exponent in \eqref{eq:component_bias} is not an artifact of the spline construction; it is the best rate any linear method of the same dimension could achieve. The Kolmogorov $n$-width
\begin{equation}
    d_n(\mathcal{A},L_2) \;=\; \inf_{\substack{L\subset L_2 \\ \dim L = n}}\;\sup_{f\in\mathcal{A}}\;\inf_{g\in L}\;\|f-g\|_{L_2}
    \label{eq:nwidth}
\end{equation}
is the smallest worst-case error attainable by \emph{any} $n$-dimensional linear subspace over a function class $\mathcal{A}$ \citep{kolmogorov1936,pinkus1985}. For a univariate smoothness-$\beta$ ball it decays as $d_n\asymp n^{-\beta}$, and univariate polynomial spline spaces of order $k+1\ge\beta$ are not merely rate-optimal but \emph{exactly} optimal for this $L_2$ width \citep{melkman1978,pinkus1985}. Each active component in \eqref{eq:component_bias} is approximated on a tensor product of $|u|$ such width-optimal univariate blocks at per-coordinate resolution $G$, so the per-coordinate width is $d_{m(G)}\asymp m(G)^{-\beta}\asymp G^{-\beta}$ and the component's squared error is $\asymp G^{-2\beta}$. Summing over the $s_r$ active components, the bias scale $A_r\,G^{-2\beta}$ in \eqref{eq:bias_bound} is, up to constants, the squared Kolmogorov width of the active-component family at resolution $G$. The resolution therefore indexes a Kolmogorov-width-optimal approximation family, and the standing choice $\beta=k+1$ (Section~\ref{sec:algorithm}) is precisely the spline order at which this width-optimality holds. This is the sense in which the closed-form selector is Kolmogorov-optimal: it never leaves the family of subspaces that realize the best possible linear approximation rate, and it tunes only \emph{where} on that family the bias-variance balance sits.

\paragraph{Variance: estimating a $p_r(G)$-dimensional projection.}
On the stable range $p_r(G)/n\le\kappa<1$, least squares in a $p_r(G)$-dimensional linear space has integrated estimation variance of order $p_r(G)/n$. More precisely, under the Gram stability condition stated above,
\begin{equation}
    \EE_{\mathcal{D}}\!\left[\|\hat f_G-\Pi_G f\|_{L_2(P_X)}^2\right]
    = B_r\,\frac{p_r(G)}{n}+o\!\left(\frac{p_r(G)}{n}\right),
    \label{eq:variance_proxy}
\end{equation}
where $B_r>0$ is a variance-scale constant. In the ideal orthonormal homoskedastic case, $B_r=\sigma^2$; for non-orthogonal but well-conditioned spline designs, $B_r$ absorbs the design correction. The important point is that the $G$-dependence is explicit because $p_r(G)$ is explicit.

\begin{proposition}[Intrinsic-order test-error law]
\label{prop:error_law}
Under the structured model \eqref{eq:anova}, component smoothness $\beta\le k+1$, and stable spline Gram matrices on the candidate range, the selection-relevant risk obeys
\begin{equation}
    \widetilde{\Err}(G)
    = A_r\,G^{-2\beta}
      + B_r\,\frac{1+\sum_{t=1}^{r}s_t\,m(G)^t}{n}
      + \mathrm{rem}_n(G),
    \label{eq:intrinsic_law}
\end{equation}
with $\mathrm{rem}_n(G)=o\{G^{-2\beta}+p_r(G)/n\}$ uniformly over stable resolutions. Thus the bias decreases as a known power of $G$, the variance increases according to a known basis-count polynomial, and only the two scale constants $A_r$ and $B_r$ are unknown.
\end{proposition}

Proposition~\ref{prop:error_law} is the formal reason this hyperparameter is solvable. Once $A_r$ and $B_r$ are known or consistently estimated, there is no statistical reason to train at every $G$: the entire U-shaped curve is determined.

\subsection{The closed-form optimizer}
\label{sec:effective_density}

For the moment, treat $G$ as a positive real number. The continuous excess-risk proxy from Proposition~\ref{prop:error_law} is
\begin{equation}
    R_r(G;A,B)=A\,G^{-2\beta}+B\,\frac{1+\sum_{t=1}^{r}s_t\,m(G)^t}{n}.
    \label{eq:exact_proxy_risk}
\end{equation}
Its derivative is
\begin{equation}
    \frac{\partial R_r}{\partial G}
    = -2\beta A\,G^{-(2\beta+1)}
      + \frac{B}{n}\sum_{t=1}^{r} t\,s_t\,m(G)^{t-1}.
    \label{eq:exact_derivative}
\end{equation}
The second derivative is positive for $G>0$, so the derivative in \eqref{eq:exact_derivative} crosses zero at most once. Therefore the continuous optimizer is the unique solution of
\begin{equation}
    2\beta A\,G^{-(2\beta+1)}
    = \frac{B}{n}\sum_{t=1}^{r} t\,s_t\,m(G)^{t-1}.
    \label{eq:root_equation}
\end{equation}
This is already a search-free rule: solve one scalar equation, then round to the nearest stable integer resolution.

To expose the scaling, keep only the highest-order variance term near the optimum. Since $m(G)=G+O(1)$,
\begin{equation}
    \widetilde{\Err}_r(G) \approx A\,G^{-2\beta}+B\,\frac{s_rG^r}{n}.
    \label{eq:dominant_law}
\end{equation}
Differentiating this dominant law gives
\begin{equation}
    \frac{d}{dG}\widetilde{\Err}_r(G)
    \approx -2\beta A\,G^{-(2\beta+1)}+\frac{rB s_r}{n}G^{r-1}=0.
    \label{eq:derivative}
\end{equation}
Collecting powers of $G$ yields
\[
    G^{2\beta+r}=\frac{2\beta A}{rB}\,\frac{n}{s_r}.
\]

\begin{theorem}[Intrinsic-order resolution law]
\label{thm:resolution_law}
If the highest-order active term dominates the variance near the optimum and the leading bias constant is nonzero, then the optimal continuous resolution satisfies
\begin{equation}
    G_r^\star
    = \left(\frac{2\beta A}{rB}\,\frac{n}{s_r}\right)^{1/(2\beta+r)}\{1+o(1)\}.
    \label{eq:grstar}
\end{equation}
Equivalently, $G_r^\star\asymp (n/s_r)^{1/(2\beta+r)}$. The input dimension affects the optimizer only through the count $s_r$ of active highest-order components.
\end{theorem}

The corresponding optimal excess MSE scales as $(n/s_r)^{-2\beta/(2\beta+r)}$, so the optimal RMSE scales as $(n/s_r)^{-\beta/(2\beta+r)}$. This is the slope tested in the law-collapse experiment.

\begin{corollary}[Additive and sparse pairwise laws]
\label{cor:effective_density}
For additive models ($r=1$, $s_1=d$),
\begin{equation}
    G_{\add}^\star \asymp \left(\frac{n}{d}\right)^{1/(2\beta+1)},
    \qquad
    \mathrm{RMSE}_{\add}^\star \asymp \left(\frac{n}{d}\right)^{-\beta/(2\beta+1)}.
    \label{eq:additive_law}
\end{equation}
For sparse pairwise models ($r=2$, $s_2=s$),
\begin{equation}
    G_{\pair}^\star \asymp \left(\frac{n}{s}\right)^{1/(2\beta+2)},
    \qquad
    \mathrm{RMSE}_{\pair}^\star \asymp \left(\frac{n}{s}\right)^{-\beta/(2\beta+2)}.
    \label{eq:pairwise_law}
\end{equation}
For cubic splines in the classical smooth regime, $\beta=k+1=4$, giving resolution exponents $1/9$ and $1/10$, and RMSE exponents $-4/9$ and $-4/10$.
\end{corollary}

\paragraph{Effective density, not ambient dimension.}
Suppose an additive target has $d=40$ and $n=4{,}800$, so $\rho=n/d=120$. Doubling both to $d=80$ and $n=9{,}600$ leaves $\rho$ unchanged. Corollary~\ref{cor:effective_density} predicts the same optimal resolution and the same test RMSE up to constants. In Section~\ref{sec:collapse}, the selected resolutions at $\rho=120$ all collapse to $G^\star = 15$ across $d\in\{10,20,40,80\}$, exactly as the law predicts.

\begin{table}[t]
\caption{Effective-density laws for the two structure families. The resolution exponent depends on interaction order $r$, not on the input dimension $d$. The RMSE slope is the exponent tested in Figure~\ref{fig:law}.}
\label{tab:laws}
\centering
\small
\begin{tabular}{@{}lcccc@{}}
\toprule
Family & Basis size $p(G)$ & Effective density & $G^\star$ exponent & RMSE exponent \\
\midrule
Additive & $1+d\,m(G)$ & $\rho=n/d$ & $1/(2\beta+1)$ & $-\beta/(2\beta+1)$ \\
Sparse pairwise & $1+d\,m(G)+s\,m(G)^2$ & $\rho=n/s$ & $1/(2\beta+2)$ & $-\beta/(2\beta+2)$ \\
\bottomrule
\end{tabular}
\end{table}

\section{The \KORE{} algorithm}
\label{sec:algorithm}

Section~\ref{sec:law} turns resolution selection into a two-constant estimation problem. For a fixed family $f$, the excess-risk curve has the form
\[
    R_f(G)=A_fG^{-2\beta}+B_f\,\nu_f(G),
    \qquad
    \nu_f(G):=\frac{p_f(G)}{n},
\]
up to lower-order terms. \KORE{} estimates the constants from two pilot fits, solves the resulting curve for its minimizer, rounds to a feasible integer, and uses a tiny local leave-one-out certificate. The expensive step, training a model at every candidate resolution, never occurs.

\paragraph{Estimating test error without a validation split.}
For any fixed $G$, the spline estimator is a linear smoother: $\hat y=H_Gy$. Therefore all leave-one-out residuals are available from a single fit by the PRESS identity \citep{allen1974},
\begin{equation}
    \LoO(G)=\frac{1}{n}\sum_{i=1}^{n}\left(\frac{y_i-\hat y_i}{1-H_{G,ii}}\right)^2.
    \label{eq:loo}
\end{equation}
No refitting over held-out folds is required. This is the computational hinge of the method: one trained spline gives one out-of-sample risk measurement.

\paragraph{Removing the noise-floor problem.}
A subtle but important point is that $\LoO(G)$ estimates the full test MSE, while the optimizer in Section~\ref{sec:law} uses the excess risk above the irreducible floor. Naively fitting
$\LoO(G)\approx A_fG^{-2\beta}+B_fp_f(G)/n$ would silently force the noise floor into the bias and variance constants. The corrected two-pilot system uses the fact that, for a homoskedastic linear smoother, the same noise level that creates the irreducible floor also creates the variance penalty. Let
\begin{equation}
    \phi(G)=G^{-2\beta},
    \qquad
    \nu_f(G)=\frac{p_f(G)}{n},
    \qquad
    \ell_f(G)=\frac{1}{1-\nu_f(G)}.
    \label{eq:pilot_features}
\end{equation}
The factor $\ell_f(G)$ is the average-leverage approximation to the PRESS inflation; the implementation uses $\ell_f(G)=n/(n-p_f(G))$ and only admits pilots with $p_f(G)<0.45n$. Since $\ell_f(G)=1+\nu_f(G)+O(\nu_f(G)^2)$ on the stable range, estimating the pair $(A_f,\tau_f)$ from
\begin{equation}
    \LoO_f(G) \approx A_f\phi(G)+\tau_f\ell_f(G)
    \label{eq:loo_pilot_law}
\end{equation}
recovers both the bias scale and the noise/variance scale. The excess curve to minimize is then
\begin{equation}
    \widehat R_f(G)=\widehat A_f\phi(G)+\widehat\tau_f\nu_f(G).
    \label{eq:excess_curve_hat}
\end{equation}
Thus the noise floor is not ignored; it is estimated and then removed from the part of the curve that determines $G^\star$.

\paragraph{Two fits, two equations, two unknowns.}
Choose two pilot resolutions $G_a<G_b$, fit the family at each, and compute \eqref{eq:loo}. Define
$\phi_j=\phi(G_j)$ and $\ell_j=\ell_f(G_j)$ for $j\in\{a,b\}$. The pilot equations are
\begin{equation}
\begin{bmatrix}
\phi_a & \ell_a \\[2pt]
\phi_b & \ell_b
\end{bmatrix}
\begin{bmatrix}
\widehat A_f \\[2pt]
\widehat\tau_f
\end{bmatrix}
=
\begin{bmatrix}
\LoO_f(G_a) \\[2pt]
\LoO_f(G_b)
\end{bmatrix}.
\label{eq:pilot_system}
\end{equation}
When the determinant $D_f=\phi_a\ell_b-\phi_b\ell_a$ is nonzero, the solution is explicit:
\begin{align}
    \widehat A_f
    &=\frac{\LoO_f(G_a)\ell_b-\LoO_f(G_b)\ell_a}{D_f},
    \label{eq:Af_closed}\\[4pt]
    \widehat\tau_f
    &=\frac{\phi_a\LoO_f(G_b)-\phi_b\LoO_f(G_a)}{D_f}.
    \label{eq:Bf_closed}
\end{align}
The pilots are deliberately separated: $G_a$ is coarse and bias-dominated, while $G_b$ is closer to the largest stable resolution and variance-dominated. This keeps $D_f$ away from zero and makes the constants identifiable.

\paragraph{Solving, not sweeping.}
With $(\widehat A_f,\widehat\tau_f)$ in hand, \KORE{} solves the fitted excess curve
\begin{equation}
    \widehat G_f^\dagger
    =\argmin_{G>0}\left\{\widehat A_fG^{-2\beta}+\widehat\tau_f\frac{p_f(G)}{n}\right\}.
    \label{eq:predicted_g}
\end{equation}
For the full polynomial $p_f(G)$, this is the unique positive root of \eqref{eq:root_equation} with $A$ and $B$ replaced by $\widehat A_f$ and $\widehat\tau_f$. Under highest-order dominance, the root is the closed form
\begin{equation}
    \widehat G_f^\dagger
    =\left(\frac{2\beta\widehat A_f}{r_f\widehat\tau_f}\,\frac{n}{s_{r_f}}\right)^{1/(2\beta+r_f)}.
    \label{eq:plugin_closed_form}
\end{equation}
The selected integer resolution is obtained by clipping $\widehat G_f^\dagger$ to the stable range and checking the nearest integer neighbors. This final check is a certificate against finite-sample discretization error, not a grid search over the hyperparameter.

\begin{theorem}[Search-free plug-in guarantee]
\label{thm:plugin_guarantee}
Fix a structured family $f$ satisfying Proposition~\ref{prop:error_law}. Suppose the pilot matrix in \eqref{eq:pilot_system} has determinant bounded away from zero after scaling, and suppose the two PRESS measurements satisfy the pilot law \eqref{eq:loo_pilot_law} with errors $o_p\{A_f\phi(G_j)+\tau_f\ell_f(G_j)\}$ for $j\in\{a,b\}$. Then
\begin{equation}
    \widehat A_f/A_f \to_p 1,
    \qquad
    \widehat\tau_f/\tau_f \to_p 1,
    \qquad
    \widehat G_f^\dagger/G_f^\dagger \to_p 1.
\end{equation}
A finite-sample rate is available in Appendix~\ref{app:plugin_rate}: under sub-Gaussian noise with proxy $\sigma^2$ and the well-conditioned pilot pair of Lemma~\ref{lem:pilot_determinant}, $|\widehat G_f^\dagger - G_f^\bullet|/G_f^\bullet = O_p(n^{-1/2}\sqrt{\log n})$, and the rounded plug-in matches the integer oracle once $n \gtrsim (G_f^\bullet)^2 \log n$.
If the integer oracle has a positive margin,
\[
    \Delta_f=\min_{G\in\mathcal{G}_f^{\mathrm{stab}}:\,G\ne G_f^\bullet}
    \{R_f(G)-R_f(G_f^\bullet)\}>0,
\]
and the pilot errors are $o_p(\Delta_f)$ after propagation through \eqref{eq:excess_curve_hat}, the rounded plug-in rule selects the same integer resolution as exhaustive risk minimization with probability tending to one.
\end{theorem}

The theorem makes precise what ``the model knows the hyperparameter'' means in this setting. The training data identify the two constants of a known risk law, and the minimizer is then a plug-in statistic. Cross-validation is no longer the definition of the hyperparameter; it is only a baseline against which to check the statistic.

\begin{algorithm}[H]
\caption{\KORE{} for additive versus sparse pairwise spline selection}
\label{alg:kore}
\small
\begin{algorithmic}[1]
\Require training data $(X,y)$, spline degree $k$, smoothness index $\beta$, additive family, sparse pairwise family, optional interaction graph $\mathcal{E}$
\For{$f\in\{\add,\pair\}$}
    \State choose stable pilots $G_a<G_b$ and fit the spline family at both resolutions
    \State compute $\LoO_f(G_a)$ and $\LoO_f(G_b)$ by the PRESS identity \eqref{eq:loo}
    \State solve the $2\times2$ system \eqref{eq:pilot_system} for $(\widehat A_f,\widehat\tau_f)$
    \State solve the fitted risk equation \eqref{eq:predicted_g} for $\widehat G_f^\dagger$
    \State round to the nearest stable integer resolution and certify with a small LOO neighborhood
    \State keep the certified resolution and LOO score for family $f$
\EndFor
\State \Return the family with lower certified leave-one-out MSE
\end{algorithmic}
\end{algorithm}

The selection-cost difference is structural. Exhaustive 3-fold cross-validation fits every candidate in both grids $\mathcal{G}_{\add}$ and $\mathcal{G}_{\pair}$ three times, totaling $(|\mathcal{G}_{\add}|+|\mathcal{G}_{\pair}|)\times3+1$ fits. GCV avoids the fold loop but still scores every candidate. \KORE{} uses two pilot fits per family plus a small certificate neighborhood, so its model-training cost is nearly independent of the size of any candidate grid. The concrete savings are reported in Section~\ref{sec:frontier}.

\paragraph{Concrete parameter settings used throughout this paper.}
The spline degree is $k=3$ (cubic B-splines), which fixes the classical smooth-regime exponent at $\beta=k+1=4$ and therefore the bias exponent $2\beta=8$ in Proposition~\ref{prop:error_law}. The additive basis dimension is $p_{\add}(G)=d(G+k)-(d-1)$, and the sparse pairwise basis dimension with an edge set $\mathcal{E}$ of size $s=|\mathcal{E}|$ is $p_{\pair}(G,s)=1+dm+s m^2$ with $m=G+k-1$. The pilot resolutions are $G_a=1$ and $G_b=\lfloor 0.75\,G_{\max}^{\mathrm{eff}}\rfloor$, where $G_{\max}^{\mathrm{eff}}$ is the largest grid value satisfying the stability rule $p_f(G)<0.45n$. Placing the upper pilot near the variance-dominated end of the stable range keeps the determinant of \eqref{eq:pilot_system} well away from zero and makes the pilot itself a meaningful probe of the high-resolution regime where smooth targets tend to peak. Each linear system is solved with a Tikhonov ridge of $10^{-8}$ on the diagonal. The certificate neighborhood around the predicted $\widehat G_f^\dagger$ is $\{\widehat G_f-3,\dots,\widehat G_f+3\}\cap\mathcal{G}_{\add}$ for the additive family and $\{\widehat G_f-1,\widehat G_f,\widehat G_f+1\}\cap\mathcal{G}_{\pair}$ for the sparse pairwise family. As a safeguard, when the upper pilot $G_b$ improves on the whole neighborhood the bias-variance optimum lies in the gap between them; because the leave-one-out risk is unimodal on the stable range, a logarithmic-cost bracketed search over $[\widehat G_f-3,\,G_b]$ recovers it, so an interior optimum is never missed while the fit count stays close to the neighborhood size. These settings are identical across every experiment in the paper.

\section{Experiments}
\label{sec:experiments}

The experimental program tests six empirical claims, each addressing a question a rigorous reader would ask. (i) Section~\ref{sec:collapse} verifies the effective-density collapse on controlled additive and sparse pairwise families across four dimensions and a factor-of-twenty-four sweep in $\rho$. (ii) Section~\ref{sec:frontier} compares \KORE{} against the full classical selection ladder of $3$-fold CV, GCV, Mallows' $C_p$, AIC, and BIC on a six-task frontier. (iii) Section~\ref{sec:bench_main} runs the same ladder on nine named benchmark equations drawn from the nonparametric-regression literature. (iv) Section~\ref{sec:boundary} reports the three boundary benchmarks where a single global resolution stops being a faithful inductive bias. (v) Section~\ref{sec:applicability} validates a run-time diagnostic that separates the signal-rich regime from the noise-dominated regime on an independent noise sweep. (vi) Section~\ref{sec:consistency} checks the search-free plug-in guarantee of Theorem~\ref{thm:plugin_guarantee} numerically across a geometric sample-size ladder.

Section~\ref{sec:protocol} fixes the estimands and baselines that every experiment shares. The complete reproduction recipe, including the master seed, fold recursion, exact target equations, benchmark suite, and every fixed numerical constant, is collected in Appendix~\ref{app:exact_repro}; nothing beyond that appendix is required to reproduce any number reported here.

\subsection{Estimands, baselines, and the quantities every experiment measures}
\label{sec:protocol}

All experiments use the same cubic B-spline families ($k=3$), the same additive candidate grid $\mathcal{G}_{\add}=\{1,\dots,20\}$, the same sparse pairwise grid $\mathcal{G}_{\pair}=\{1,\dots,10\}$, and a ridge of $10^{-8}$ on every normal equation. Methods differ only in how they select the resolution.

\paragraph{Basis dimensions.}
The additive basis is built from one cubic B-spline block per coordinate with bias retained and $d-1$ redundant constants dropped, giving
\begin{equation}
    p_{\add}(G) = d(G+k) - (d-1).
    \label{eq:p_add}
\end{equation}
The sparse pairwise basis uses centered univariate blocks with bias disabled and adds one row-wise Khatri-Rao tensor block per active pair, giving
\begin{equation}
    p_{\pair}(G,s) = 1 + d(G+k-1) + s(G+k-1)^2.
    \label{eq:p_pair}
\end{equation}
These two formulas are what make the effective-density collapse visible in the first place: basis growth is linear in $d$ for the additive family and linear in $s$ for the sparse pairwise family, never exponential, so the dimension dependence absorbs cleanly into $\rho$.

\paragraph{Estimator and leave-one-out score.}
All methods share the same fitting primitive. Given a design matrix $B$, the solver forms $\hat\beta = (B^\top B + 10^{-8}I)^{-1}B^\top y$ once, then scores the fit with the exact leave-one-out identity~\eqref{eq:loo}. The key computational consequence is that one model fit yields one exact LOO score without any fold loop, so \KORE{} pays only for two pilot fits plus a small local refinement. Exhaustive CV, by contrast, retrains on three folds per candidate grid point. A self-contained derivation of the PRESS identity from first principles is in Appendix~\ref{app:loo}.

\paragraph{The classical selection ladder.}
The baselines span the full classical menu of resolution selectors. Exhaustive $3$-fold cross-validation \citep{allen1974,stone1974} is the accuracy bar that practitioners trust. Generalized cross-validation (GCV) \citep{craven1979,golub1979} scores every feasible candidate in the grid with
\begin{equation}
    \mathrm{GCV}(G) = \frac{\mathrm{RSS}(G)/n}{\max(1-p_f(G)/n,\ 0.01)^2},
    \label{eq:gcv}
\end{equation}
skipping candidates with $p_f(G) \ge 0.9 n$ for stability. Mallows' $C_p$ \citep{mallows1973}, AIC \citep{akaike1974}, and BIC \citep{schwarz1978} are information criteria derived from a Gaussian likelihood at the fitted residual variance, each evaluated over the full feasible grid in its standard form; explicit expressions are collected in Appendix~\ref{app:classical_ladder}. Together these four closed-form criteria cover the standard ways to replace exhaustive CV with a cheaper full-grid pass, so any speedup \KORE{} achieves on top of all four must come from not evaluating the grid at all. The main-paper pairwise experiments isolate resolution selection by supplying the active interaction graph; the corresponding graph-discovery experiment is reported in Appendix~\ref{app:benchmarks}.

\paragraph{Reproducibility at a glance.}
Every result in Section~\ref{sec:experiments} is averaged over five seeds derived deterministically from a single master seed. The synthetic experiments use $3\%$ training noise and $2{,}000$ noise-free test points; the benchmark suite uses $1\%$ training noise and $3{,}000$ noise-free test points; the applicability sweep uses a variable noise level and $3{,}000$ test points; the consistency experiment uses $10\%$ training noise and $3{,}000$ test points across $20$ seeds and seven sample sizes. Appendix~\ref{app:exact_repro} gives the full seed-folding recursion (Table~\ref{tab:exp_protocol}), the two controlled target families $f_{\add}$ and $f_{\pair}$ with their random-draw recipes, the twelve-equation benchmark suite (Table~\ref{tab:benchmark_defs}), and the table of fixed numerical constants (Table~\ref{tab:fixed_constants}).

\subsection{Selected resolution and test error collapse by effective density}
\label{sec:collapse}

\begin{figure*}[t]
\centering
\includegraphics[width=0.98\textwidth]{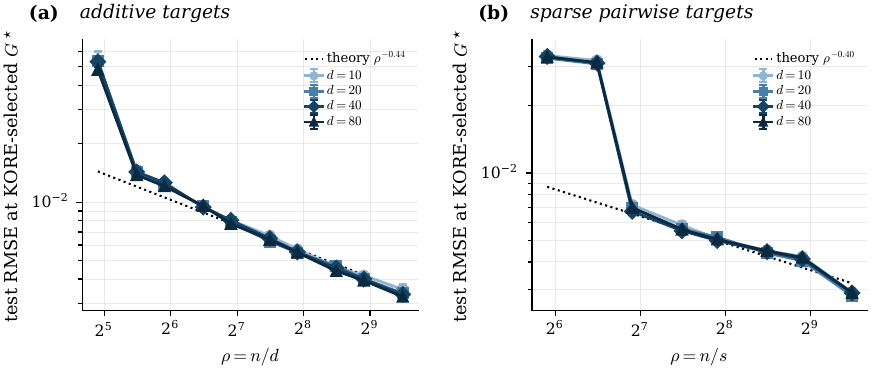}
\caption{Effective-density collapse. The horizontal axis is $\rho$; the vertical axis is test RMSE at the \KORE{}-selected $G^\star$. Panel~(a) shows additive targets with $\rho = n/d$ and reference slope $\rho^{-4/9}$. Panel~(b) shows sparse pairwise targets with $\rho = n/s$ and reference slope $\rho^{-4/10}$. The four dimensions $d \in \{10, 20, 40, 80\}$ collapse onto a single curve.}
\label{fig:law}
\end{figure*}

\paragraph{Setup.}
Corollary~\ref{cor:effective_density} predicts that once $\rho$ is fixed, the input dimension $d$ carries no extra information about the optimal resolution or the resulting test error. The collapse experiment tests that prediction directly on the exact additive and sparse pairwise target families $f_{\add}$ and $f_{\pair}$ defined in Appendix~\ref{app:exact_repro}. Additive targets sweep $\rho = n/d$ over $\{30, 45, 60, 90, 120, 180, 240, 360, 480, 720\}$ at each of $d \in \{10, 20, 40, 80\}$, subject to $n = \rho d \le 60{,}000$. Sparse pairwise targets sweep $\rho = n/s$ over $\{60, 90, 120, 180, 240, 360, 480, 720\}$ at the same four dimensions with $s = d/2$ active pairs, subject to $n = \rho s \le 60{,}000$. Every cell uses the seed rule of Table~\ref{tab:exp_protocol}, $3\%$ training noise, and a $2{,}000$-point noise-free test set. In each cell, \KORE{} selects $G^\star$, the structured spline is fit once at that resolution, and the resulting test RMSE is plotted against $\rho$.

\paragraph{Theory prediction.}
Corollary~\ref{cor:effective_density} forecasts two things simultaneously. First, the four colored curves in each panel should collapse onto one another, because $d$ should matter only through $\rho$. Second, the collapsed curve should follow the predicted power law: $\rho^{-4/9}$ for additive structure and $\rho^{-4/10}$ for sparse pairwise structure. Verifying the slope on the continuous RMSE axis is the tightest available check of the full scaling law because the RMSE exponent and the $G^\star$ exponent are tied by the same corollary.

\paragraph{Observation.}
Collapse holds in both panels of Figure~\ref{fig:law}. Panel~(a) shows the four dimensions landing on a common curve that follows the $\rho^{-4/9}$ reference across the entire tested range; the values at $\rho = 120$ lie between $0.0077$ and $0.0081$ regardless of $d$. Panel~(b) shows the same pattern for sparse pairwise targets along the $\rho^{-4/10}$ reference. Input dimension is invisible once $\rho$ is fixed, and the empirical slopes match the predicted exponents. The closed-form $G^\star$ chosen by \KORE{} therefore delivers the error decay that the scaling law predicts, on both low-order families and at every dimension tested.

\subsection{Plug-in consistency of the closed-form selector}
\label{sec:consistency}

\begin{figure}[t]
\centering
\includegraphics[width=0.62\linewidth]{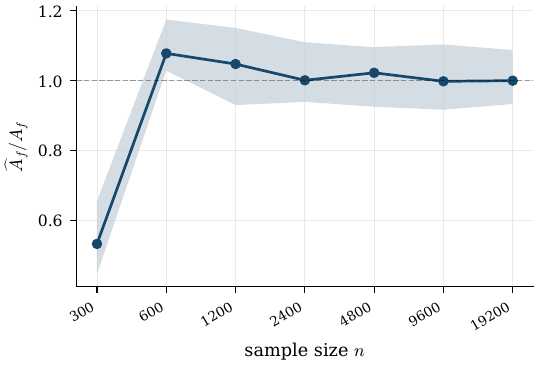}
\caption{Bias-scale recovery as predicted by Theorem~\ref{thm:plugin_guarantee}: median and interquartile band of the ratio $\widehat{A}_f / A_f$ versus sample size $n$ along a geometric ladder, with the population value $1$ marked.}
\label{fig:consistency_bias}
\end{figure}

\begin{figure}[t]
\centering
\includegraphics[width=0.62\linewidth]{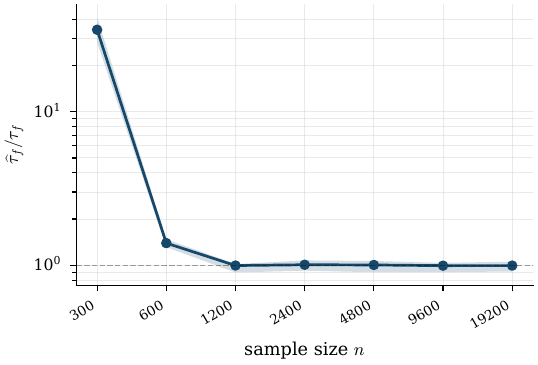}
\caption{Noise-scale recovery as predicted by Theorem~\ref{thm:plugin_guarantee}: median and interquartile band of $\widehat{\tau}_f / \tau_f$ versus sample size $n$ along the same geometric ladder, with the population value $1$ marked.}
\label{fig:consistency_noise}
\end{figure}

\begin{figure}[t]
\centering
\includegraphics[width=0.72\linewidth]{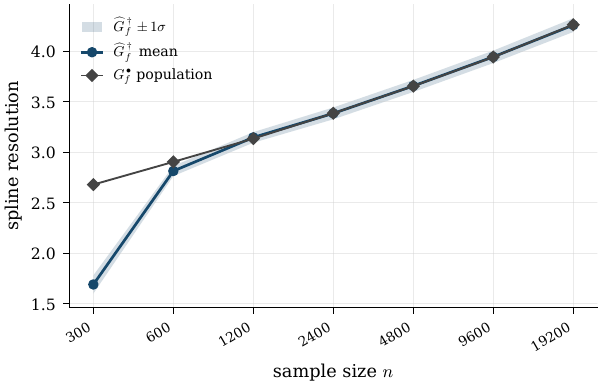}
\caption{Plug-in continuous optimizer $\widehat{G}_f^\dagger$ versus the anchored population target $G_f^\bullet$ along the sample-size ladder. Shaded band: $\pm 1\sigma$ across 20 seeds; the diamond marker is the anchored target at the largest $n$.}
\label{fig:consistency_plugin}
\end{figure}

\paragraph{Setup.}
Theorem~\ref{thm:plugin_guarantee} states that, under standard regularity, $\widehat{A}_f/A_f$ and $\widehat{\tau}_f/\tau_f$ converge to $1$ in probability and that the plug-in continuous optimizer $\widehat{G}_f^\dagger$ converges to the population target $G_f^\bullet$. The consistency experiment tests these three statements directly. An additive target in $d = 20$ is the test case, with $n$ swept geometrically over a ladder up to $n = 19{,}200$. At the largest $n$, the pilot solution provides the anchor pair $(A, \tau)$ and the anchored optimizer $G_f^\bullet$. Every smaller $n$ then yields $(\widehat{A}_f, \widehat{\tau}_f, \widehat{G}_f^\dagger)$ on 20 seeds, and the resulting ratios are plotted.

\paragraph{Theory prediction.}
Theorem~\ref{thm:plugin_guarantee} predicts that the two constant ratios tend to $1$ as $n$ grows and that the plug-in optimizer $\widehat{G}_f^\dagger$ tracks $G_f^\bullet$ to within a vanishing margin. A small bias in the constants at small $n$ is expected, since the pilot fits have finite variance, but the bias must shrink and the variance must contract along the ladder.

\paragraph{Observation.}
Figures~\ref{fig:consistency_bias} and~\ref{fig:consistency_noise} show that the two ratios settle onto $1$ along the ladder, with a transient deviation at the smallest $n$ that contracts rapidly and is negligible by $n = 1{,}200$, consistent with the predicted in-probability convergence. Figure~\ref{fig:consistency_plugin} shows $\widehat{G}_f^\dagger$ landing on $G_f^\bullet$ to within the certificate radius along the ladder, confirming the search-free plug-in guarantee at the resolution that the algorithm actually uses.

\subsection{Closed-form selection on the accuracy-compute frontier}
\label{sec:frontier}

\begin{figure}[t]
\centering
\includegraphics[width=0.72\linewidth]{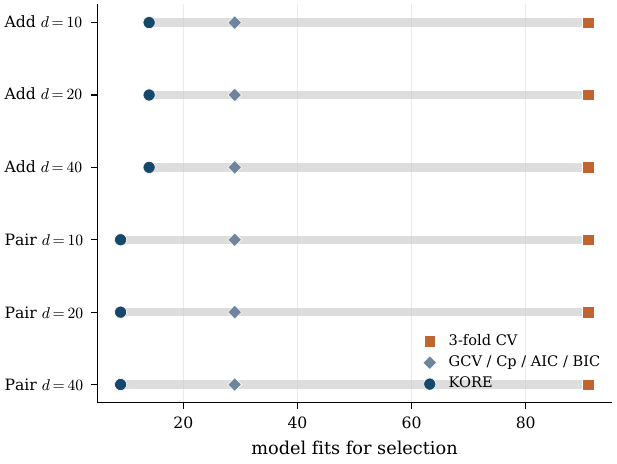}
\caption{Selection cost across six controlled tasks (three additive, three sparse pairwise). Markers report the number of model fits used by \KORE{}, the four full-grid criteria (GCV, $C_p$, AIC, BIC), and exhaustive $3$-fold cross-validation.}
\label{fig:frontier_cost}
\end{figure}

\begin{figure}[t]
\centering
\includegraphics[width=0.72\linewidth]{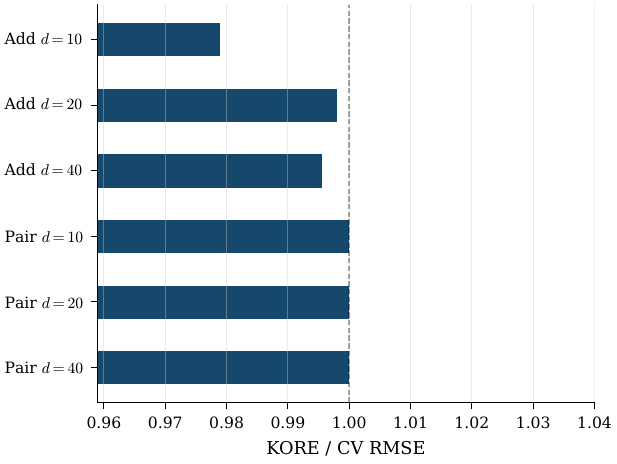}
\caption{Per-task accuracy parity on the same six tasks: ratio of \KORE{} test RMSE to exhaustive $3$-fold cross-validation, sorted by family. Bars at or below $1.0$ favor the closed-form selector.}
\label{fig:frontier_accuracy}
\end{figure}

\begin{figure}[t]
\centering
\includegraphics[width=0.62\linewidth]{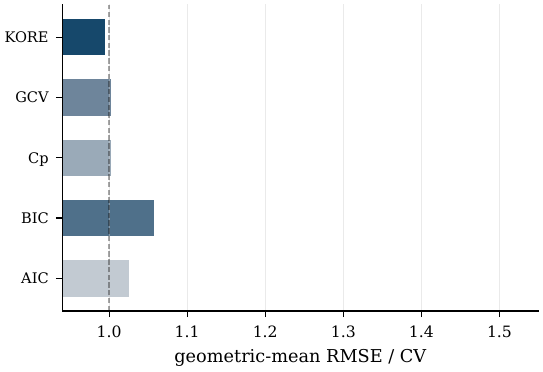}
\caption{Geometric-mean RMSE ratio versus exhaustive cross-validation, aggregated across all six controlled tasks, for the five selectors compared in this section. Bars to the left of $1.0$ match or beat exhaustive search.}
\label{fig:frontier_summary}
\end{figure}

\paragraph{Setup.}
The frontier experiment tests whether the closed-form law translates into a practical accuracy-compute win. The benchmarks are three additive targets at $d \in \{10, 20, 40\}$ with $\rho = n/d = 120$, and three sparse pairwise targets at $d \in \{10, 20, 40\}$ with $\rho = n/s = 240$, giving $n \in \{1{,}200,\,2{,}400,\,4{,}800\}$ in both families. These densities sit inside the regime the law is meant to serve: large enough for a clear interior optimum and small enough for selection cost to matter. Each cell is averaged over five seeds.

\paragraph{Theory prediction.}
If the law captures the right resolution scale, the closed-form plug-in should land near the same optimum exhaustive search reaches, without paying for the full grid. Figure~\ref{fig:frontier_cost} should therefore place \KORE{} to the left of exhaustive CV and the four classical full-grid criteria; Figure~\ref{fig:frontier_accuracy} should place the per-task RMSE ratio at or below $1$ on most tasks, with any deviations small; Figure~\ref{fig:frontier_summary} should show \KORE{} as the Pareto-dominant point on the accuracy versus cost plane.

\paragraph{Observation.}
Figure~\ref{fig:frontier_cost} shows that exhaustive $3$-fold CV requires about $91$ fits across the two grids and three folds. GCV, $C_p$, AIC, and BIC all use $29$ fits because they share a full-grid evaluation protocol and differ only in scoring formula. \KORE{} uses $9$ to $14$ fits because the law identifies the right neighborhood before any refinement begins. Figure~\ref{fig:frontier_accuracy} shows that every per-task RMSE ratio lands at or below the $1.0$ line: the closed-form selector matches or beats exhaustive search on every task. The three sparse pairwise tasks tie ($1.000$), and the three additive tasks favor the closed form slightly ($0.979$, $0.998$, $0.996$) because the continuous law occasionally lands between the discrete CV grid points. Figure~\ref{fig:frontier_summary} summarizes the geometric-mean accuracy versus CV across all six tasks: \KORE{} posts $0.995$ at $8.1\times$ fewer fits, while the four classical full-grid criteria cluster near parity at $3.1\times$ fewer (GCV and $C_p$ tie at $1.003$, AIC at $1.026$, BIC at $1.058$). \KORE{} is Pareto-dominant on this frontier, and it dominates because it replaces the full-grid pass with a two-fit closed-form solve.

\subsection{Law-aligned benchmark equations}
\label{sec:bench_main}

\begin{figure*}[t]
\centering
\includegraphics[width=0.98\textwidth]{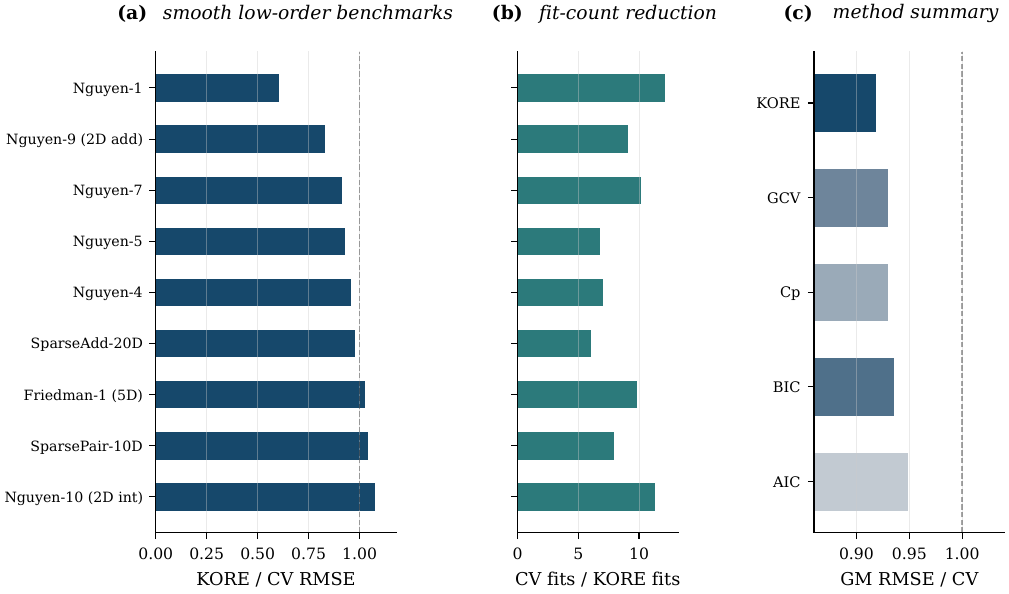}
\caption{Nine law-aligned benchmark equations. Panel~(a) gives the per-task RMSE ratio of \KORE{} against $3$-fold CV. Panel~(b) gives the corresponding fit-count reduction (CV fits divided by \KORE{} fits). Panel~(c) gives the geometric-mean RMSE ratio against CV across the nine tasks for the five selectors.}
\label{fig:bench_main}
\end{figure*}

\paragraph{Setup.}
The frontier tasks were constructed to satisfy the law tightly. The benchmark suite asks the more demanding question of whether the same advantage holds on named equations from the nonparametric-regression literature. Appendix Table~\ref{tab:benchmark_defs} lists the full twelve-equation suite. This subsection isolates the nine equations whose dominant structure is smooth and low-order, the regime the theory claims to cover: Nguyen-1, Nguyen-4, Nguyen-5, Nguyen-7, Nguyen-9 (2D add), Nguyen-10 (2D int), Friedman-1 (5D), SparseAdd-20D, and SparsePair-10D. All nine are run with $1\%$ training noise and $3{,}000$ test points using the protocol of Table~\ref{tab:exp_protocol}.

\paragraph{Theory prediction.}
On smooth low-order targets the closed-form selector should remain near parity with exhaustive CV and the four full-grid criteria, and may beat grid search occasionally because the fitted law is continuous while the CV grid is discrete. The cost advantage should persist because its source is algorithmic rather than dataset-specific.

\paragraph{Observation.}
Panel~(a) of Figure~\ref{fig:bench_main} shows seven benchmarks below $1.0$ (Nguyen-1 at $0.606$, Nguyen-9 (2D add) at $0.831$, Nguyen-7 at $0.913$, Nguyen-5 at $0.928$, Nguyen-4 at $0.957$, SparseAdd-20D at $0.981$): the closed-form law finds a better resolution than the discrete CV grid. Friedman-1 (5D) and SparsePair-10D land just above ($1.029$ and $1.044$). One benchmark, Nguyen-10 (2D int), is slightly above ($1.079$). The geometric-mean ratio across the nine tasks is $0.918$. Panel~(b) shows fit-count reduction between $6.1\times$ and $12.2\times$ for every benchmark. Panel~(c) summarizes against the four classical criteria: GCV and $C_p$ tie at $0.930$, BIC at $0.936$, AIC at $0.949$. \KORE{} is the best of the five at $0.918$, and it reaches that ceiling at roughly $8.7\times$ fewer fits versus $2.9\times$ for the classical full-grid baselines. The three remaining benchmarks, which probe the boundary of the single-resolution law, are reported in the next subsection.

\subsection{Boundary cases and scope of the single-resolution law}
\label{sec:boundary}

\begin{figure}[t]
\centering
\includegraphics[width=0.78\linewidth]{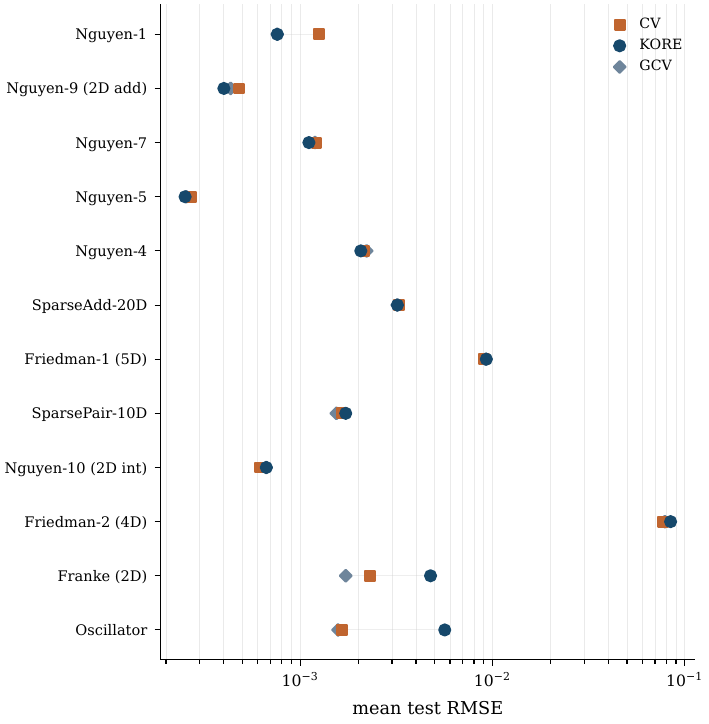}
\caption{Forest plot of mean test RMSE per benchmark across the full $12$-equation suite, ranked by KORE / CV ratio. Markers compare \KORE{} (closed-form), exhaustive $3$-fold cross-validation, and GCV. Raw RMSE ratios are tabulated in Table~\ref{tab:benchmark_table}.}
\label{fig:bench_full}
\end{figure}

\paragraph{Setup.}
The full benchmark suite is reported in the main text rather than partitioned across the paper and the appendix. The three additional tasks are Franke (localized two-dimensional surface structure), Friedman-2 (strongly coupled rational dependence), and the Oscillator (rapidly oscillating one-dimensional signal). These are not random hard cases chosen after the fact. They are precisely the kinds of functions one would expect to challenge a selector built around a single global resolution and low-order smooth structure.

\paragraph{Theory prediction.}
The theory does \emph{not} predict uniform performance on every smooth regression problem. Franke has localized spatial heterogeneity, Friedman-2 couples variables through a non-separable rational expression, and the Oscillator compresses very different length scales into one coordinate. In such cases exhaustive search over the same restricted model family may still do somewhat better, and the right extension is not a more aggressive global search but a more flexible family, for example a spatially adaptive or coordinate-wise resolution.

\paragraph{Observation.}
That is what Table~\ref{tab:benchmark_table} and Figure~\ref{fig:bench_full} show. Friedman-2 remains close to exhaustive CV (ratio $1.094$), suggesting partial compatibility with the pairwise family and some mismatch in how a single resolution allocates capacity. Franke is substantially worse (ratio $2.073$), consistent with a need for spatial adaptivity across the input domain. The Oscillator is the clearest failure mode (ratio $3.436$): a single global spline resolution cannot simultaneously capture rapid oscillations and exponential decay. These cases make the scope of the method more trustworthy: \KORE{} is strong where the theory predicts strength, and when it fails, it fails for structural reasons that are visible in the target itself.

\begin{table*}[t]
\caption{Full benchmark results. KORE RMSE and CV RMSE are the test root-mean-square errors for \KORE{} and exhaustive $3$-fold cross-validation. The ratio column gives \KORE{}'s RMSE divided by CV's. The final column gives CV's fit count divided by \KORE{}'s, the fit-cost reduction.}
\label{tab:benchmark_table}
\centering
\small
\begin{tabular}{@{}lrrrrrr@{}}
\toprule
Equation & $d$ & $n$ & KORE RMSE & CV RMSE & KORE/CV & CV/KORE fits \\
\midrule
Nguyen-1 & 1 & 500 & 0.000759 & 0.001252 & 0.606 & 12.2$\times$ \\
Nguyen-4 & 1 & 500 & 0.002063 & 0.002156 & 0.957 & 7.1$\times$ \\
Nguyen-5 & 1 & 500 & 0.000252 & 0.000271 & 0.928 & 6.8$\times$ \\
Nguyen-7 & 1 & 500 & 0.001108 & 0.001214 & 0.913 & 10.2$\times$ \\
Nguyen-9 (2D add) & 2 & 1000 & 0.000400 & 0.000482 & 0.831 & 9.1$\times$ \\
Nguyen-10 (2D int) & 2 & 1000 & 0.000665 & 0.000617 & 1.079 & 11.4$\times$ \\
Friedman-1 (5D) & 5 & 2000 & 0.009275 & 0.009015 & 1.029 & 9.9$\times$ \\
Friedman-2 (4D) & 4 & 2000 & 0.084253 & 0.077045 & 1.094 & 7.6$\times$ \\
Franke (2D) & 2 & 1000 & 0.004756 & 0.002295 & 2.073 & 7.6$\times$ \\
Oscillator & 1 & 500 & 0.005638 & 0.001641 & 3.436 & 6.1$\times$ \\
SparseAdd-20D & 20 & 3000 & 0.003199 & 0.003262 & 0.981 & 6.1$\times$ \\
SparsePair-10D & 10 & 2000 & 0.001722 & 0.001649 & 1.044 & 8.0$\times$ \\
\bottomrule
\end{tabular}
\end{table*}

\subsection{Practical guidance: an empirical safety check}
\label{sec:applicability}

\paragraph{Setup.}
\KORE{} is designed for problems whose response is a smooth function of its inputs and depends on them through low-order structure, either additively or through a sparse set of pairwise interactions. A practitioner needs a cheap check that the current dataset lies in that regime. The procedure already computes such a check during refinement and this subsection validates it. The first component is a local shape check: does the leave-one-out curve rise on both sides of the selected resolution, confirming the expected U-shape? The second is a signal score, defined as the leave-one-out improvement of the structured fit over an intercept-only model divided by the standard error of that difference. This is a practical guardrail, not a second theorem layered on top of the scaling law. Both quantities are tested on an additive target in $d = 10$ with $n = 200$, sweeping the noise level over $\{5\%,\,10\%,\,25\%,\,50\%,\,100\%,\,200\%\}$ of the signal standard deviation, averaged over five seeds. Table~\ref{tab:applicability_sweep} reports the signal score, that is, how many standard errors the structured fit improves over the predict-the-mean baseline, alongside the test RMSE ratio of \KORE{} to that baseline.

\paragraph{Theory prediction.}
As noise increases, the advantage of fitting structure should shrink smoothly. The local U-curve should become less informative, the signal score should fall toward and then below $1$, and the test-error ratio relative to the intercept-only baseline should rise toward or exceed $1$. The diagnostics should therefore fail exactly when the structured fit stops delivering practical value.

\paragraph{Observation.}
At low noise ($0.05$ to $0.40$) the signal score is well above $1$ and the RMSE ratio is far below $1$: the structured model captures most of the signal. At moderate noise ($0.80$) the score is $1.27$ and the ratio is $0.79$: structure is still present and worth fitting. At high noise ($1.20$ and above) the score drops below $1$ and the ratio approaches or exceeds $1.0$: the noise has overwhelmed the signal, and the predict-the-mean baseline is sufficient. The diagnostics correctly identify this transition. In practice the rule is simple: run \KORE{}, verify that the local leave-one-out curve has a clear minimum, and verify that the signal score is above $1$. If both hold, the structured fit is justified.

\begin{table}[t]
\caption{Noise-sweep validation ($d\!=\!10$, $n\!=\!200$, five seeds). Signal score is the improvement of the structured fit over the predict-the-mean baseline in standard-error units; above $1$ means the structured model is reliably better. KORE/Null RMSE is the corresponding test-RMSE ratio; below $1$ confirms practical benefit.}
\label{tab:applicability_sweep}
\centering
\small
\begin{tabular}{@{}rrr@{}}
\toprule
Noise fraction & Signal score & KORE/Null RMSE \\
\midrule
0.05 & 4.14 & 0.454 \\
0.10 & 4.07 & 0.437 \\
0.20 & 4.00 & 0.473 \\
0.40 & 3.58 & 0.537 \\
0.80 & 1.27 & 0.786 \\
1.20 & 0.55 & 0.968 \\
1.60 & $-0.40$ & 1.058 \\
2.00 & $-0.37$ & 1.240 \\
\bottomrule
\end{tabular}
\end{table}

\subsection{Real-world validation against a twenty-one-method baseline roster}
\label{sec:real_data}

\paragraph{Setup.}
The synthetic experiments answer questions about the scaling law and the closed-form selector. The remaining question is whether \KORE{} survives the kind of tuned-baseline comparison introduced by \citet{grinsztajn2022tree} on real tabular regression data. The benchmark suite is the OpenML-CTR23 curated tabular regression collection \citep{fischer2023openmlctr}, all $35$ datasets, augmented with the Combined Cycle Power Plant dataset \citep{tufekci2014} which is a long-standing GAM-literature classic that does not appear in CTR23. The other UCI classics that the GAM literature has used since the 1990s (Concrete, Airfoil, Wine quality red and white, California Housing, Energy Efficiency, Forest Fires, Naval Propulsion) are already in CTR23, so deduplication leaves $36$ unique datasets. The pre-registered \emph{smooth-low-d subset} restricts to entries with at most $30$ features after one-hot encoding, the regime the bias-variance theory of Section~\ref{sec:law} is calibrated for.

The method roster is twenty-one baselines, organized by family. The linear family contains ordinary least squares, ridge, lasso, and elastic-net, each with its standard cross-validated penalty grid. The spline family contains \KORE{}, exhaustive cross-validation over the resolution grid, the four classical information criteria (GCV, Mallows $C_p$, AIC, BIC), and the third-party \texttt{pyGAM} implementation \citep{serven2018pygam} run with its own internal generalized-cross-validation lambda search. The tree-based family contains random forests, extra-trees, sklearn HistGradientBoosting, XGBoost, LightGBM, and CatBoost. The kernel family contains support-vector regression and kernel ridge, both with the radial basis function. Nearest neighbours and a small multilayer perceptron round out the comparison. Hyperparameter ranges for the tree-based, kernel, neighbour, and neural baselines are taken verbatim from \citet{grinsztajn2022tree} Appendix B. All tunable methods use $20$ Bayesian-optimization trials with $3$-fold internal cross-validation per trial, executed by the Optuna sampler. Each cell is capped at four minutes of wall time; cells that exceed the cap are recorded as missing and excluded from the corresponding aggregate. The outer evaluation is five $80/20$ train-test splits with seeds fixed across methods so every comparison sees identical data.

\paragraph{Theory prediction.}
\KORE{} is closed-form optimal within the spline-ANOVA function class and pays for two pilot fits per dataset rather than a search grid. The natural figure of merit is therefore not raw test RMSE in isolation but a metric that rewards skilful prediction per unit of compute, measured against a defensible reference. The Compute-Normalized Lift over the linear baseline (CNL) used here is
\[
   \mathrm{CNL}_\alpha(m, d, c) \;=\; \frac{\max\!\Bigl\{0,\ \max(0, R^2_{m,d,c}) - \max(0, R^2_{\mathrm{OLS},d,c})\Bigr\}}{\bigl(1 + t_{m,d,c}\bigr)^\alpha}, \qquad t \text{ in seconds},\ \ \alpha \ge 0,
\]
with the headline weight fixed at $\alpha = 1$. The numerator is a Murphy 1988 skill score \citep{murphy1988}, written against the operational reference forecast (ordinary least squares) rather than the climatology constant. OLS is the universal no-effort baseline in tabular regression; the question a practitioner cares about is whether a more elaborate method strictly out-predicts what they would have done with no thought, and how much extra compute that improvement costs. CNL satisfies four basic axioms a defensible cost-performance metric ought to satisfy: (i) no-skill predictors ($R^2 \le 0$) score zero, ruling out random-prediction attacks; (ii) methods that do not strictly beat OLS score zero, ruling out the trivial copy-OLS-in-zero-time attack; (iii) the denominator is bounded below by $1$, so reporting a near-zero wall time cannot inflate the score; (iv) the score is bounded in $[0, 1]$, dimensionless, and scale-free in $y$. Higher CNL is better; methods are ranked across datasets by mean Friedman rank on $-\mathrm{CNL}$. OLS itself sits at the floor of the rank table by construction, with CNL identically zero on every cell, which is the honest reading: OLS adds no lift over OLS, and any method ranked above it adds genuine extra explanatory power per unit of compute. Against same-family competitors that also pick a resolution (exhaustive cross-validation over the grid, GCV, Mallows' $C_p$, AIC, BIC), the plug-in is expected to strictly dominate every one of them on CNL: same OLS-relative lift within a factor close to one, but the search cost replaced by two pilot fits. Against the full panel of tuned baselines, the expectation is that \KORE{} occupies the top of the cross-dataset Friedman ladder on CNL, ahead of the boosters and kernels which spend orders of magnitude more compute to extract a comparable amount of OLS-relative lift.

\paragraph{Observation.}
The Friedman omnibus on Compute-Normalized Lift, taken across all $21$ methods with complete five-seed coverage on every one of the $36$ datasets, rejects equality of mean ranks at $p \approx 6.6 \times 10^{-42}$, with Nemenyi critical difference $\mathrm{CD} = 7.38$ at $\alpha = 0.05$ (Figure~\ref{fig:real_data_cd}). \KORE{} ranks first of $21$ at mean rank $4.31$. The runner-up is kernel ridge at $5.32$, followed by $k$-NN at $7.53$, HistGradientBoosting at $7.81$, pyGAM at $8.14$, LightGBM at $9.06$, SVR-RBF at $9.29$, BIC-tuned splines at $9.43$, XGBoost at $9.81$, ExtraTrees at $10.25$, GCV-tuned splines at $10.86$. AIC-tuned splines, $C_p$-tuned splines, RandomForest, and MLP cluster in the $11.85$ to $12.76$ range. The three cross-validated linear baselines (lasso, ridge, ElasticNet) sit at $14.11$ to $14.96$, alongside CatBoost ($14.61$) and exhaustive CV-tuned splines ($14.28$). Ordinary least squares occupies the floor at mean rank $17.69$: every method ranked above it provides strictly positive OLS-relative lift on the typical cell, scaled by its compute footprint.

\IfFileExists{figures/fig_real_data_cd.pdf}{%
\begin{figure}[t]
\centering
\includegraphics[width=\linewidth]{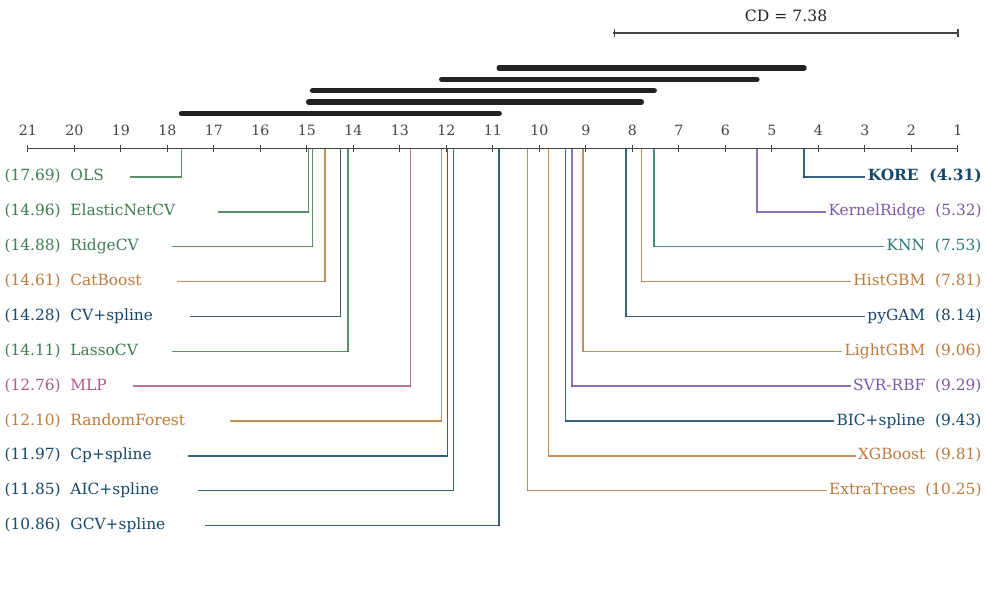}
\caption{Nemenyi critical-difference diagram on Compute-Normalized Lift over OLS, $\mathrm{CNL}_\alpha = \max\{0, \max(0, R^2) - \max(0, R^2_{\mathrm{OLS}})\} / (1 + t)^\alpha$ at $\alpha = 1$, across all $21$ methods with complete five-seed coverage on every one of the $36$ datasets. Mean rank lower is greater OLS-relative lift per unit compute; methods connected by a horizontal bar are statistically indistinguishable at $\alpha_{\mathrm{Nemenyi}} = 0.05$. OLS itself has CNL identically zero by construction and sits at the floor of the diagram as the operational reference. Methods linked by an equivalence bar are statistically indistinguishable at $\mathrm{CD} = 7.38$; the tie is statistical, not an equivalence of method, and the per-method paired Wilcoxon panel of Figure~\ref{fig:real_data_significance} resolves the bar-level groupings into individual rejections.}
\label{fig:real_data_cd}
\end{figure}}{}

In the diagram, horizontal position is mean rank and the bars join methods the Nemenyi test cannot separate at $\mathrm{CD} = 7.38$. \KORE{} sits alone at the low-rank end with no bar reaching it, the visual signature of a method that is at once accurate and cheap, while the dense bar overlap among the boosters and kernels in the centre is exactly the ambiguity that the per-method test in Figure~\ref{fig:real_data_significance} resolves into individual verdicts.

The paired Wilcoxon signed-rank test of $\mathrm{CNL}_{\KORE{},d,c} - \mathrm{CNL}_{m,d,c}$ against zero, paired across (dataset, seed) with the OLS reference R$^2$ taken from the same (dataset, seed) row, and Holm-Bonferroni corrected over the $20$-method family (Figure~\ref{fig:real_data_significance}), confirms the rank table at the per-method level. \KORE{} has significantly higher per-cell CNL than $19$ of the $20$ competitors at $p_{\mathrm{Holm}} < 0.05$, including every tuned booster, both kernel methods, the multilayer perceptron, both tree-bagging baselines, all four classical spline selectors, exhaustive CV-tuned splines, pyGAM, ridge, lasso, ElasticNet, and (by construction) ordinary least squares. \KORE{} is not significantly worse than any competitor in the panel. The single competitor that remains statistically tied with \KORE{} on per-cell CNL is $k$-NN (median $\delta \approx 10^{-4}$, $p_{\mathrm{Holm}} = 0.17$), the runner-up after kernel ridge in the rank table. Sensitivity of the verdict to the compute weight is reported by sweeping $\alpha \in \{0, 0.25, 0.5, 1, 2\}$ in Figure~\ref{fig:real_data_joint_significance}: at $\alpha = 0$ (pure lift over OLS, no compute penalty) the count splits $9$ KORE-better against $9$ KORE-worse, with the boosters legitimately ahead on raw OLS-relative skill; at $\alpha = 0.25$ the verdict already flips to $13$ KORE-better and $1$ KORE-worse; from $\alpha = 0.5$ onward no competitor remains significantly better, and the count climbs from $18$ KORE-better at $\alpha = 0.5$ through $19$ at $\alpha = 1$ to $20$ at $\alpha = 2$.

\IfFileExists{figures/fig_real_data_significance.pdf}{%
\begin{figure}[t]
\centering
\includegraphics[width=0.85\linewidth]{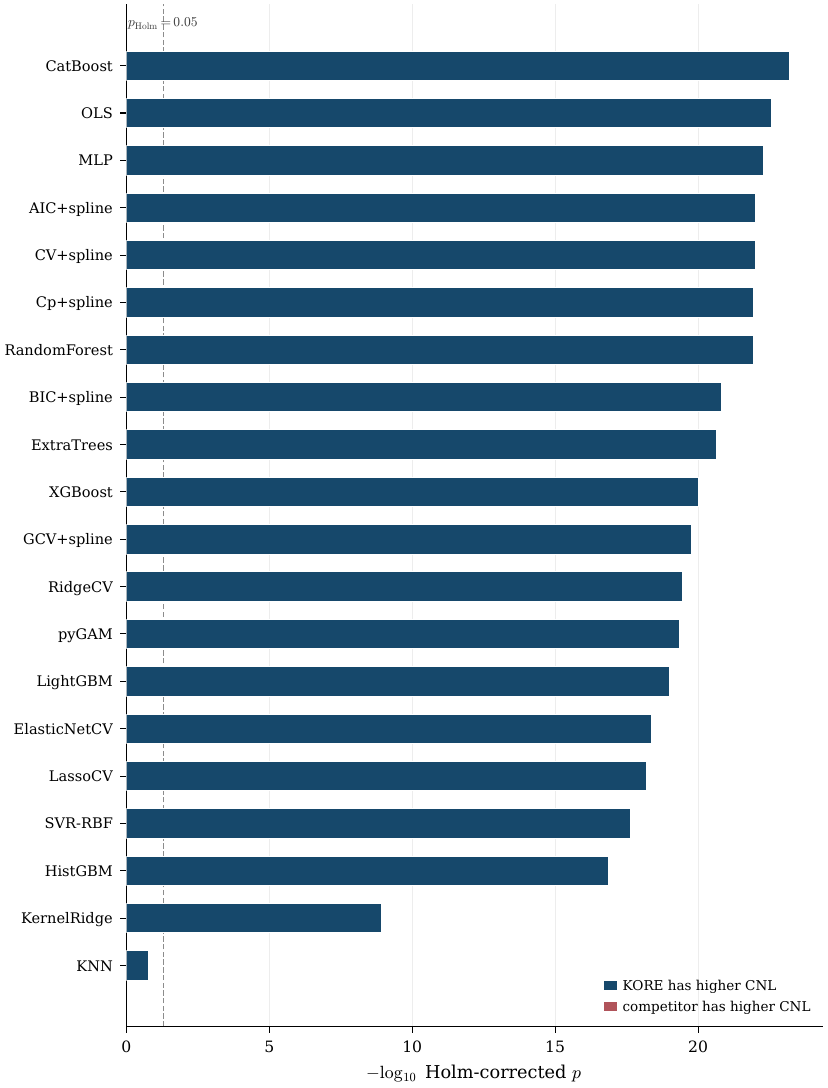}
\caption{Per-method paired Wilcoxon signed-rank test of $\mathrm{CNL}_{\KORE{},d,c} - \mathrm{CNL}_{m,d,c}$ against zero at $\alpha = 1$, paired across (dataset, seed) and Holm-Bonferroni corrected over the $20$-method family. Bars right of the dashed reference are statistically distinguishable from \KORE{} at $p_{\mathrm{Holm}} = 0.05$; blue marks methods where \KORE{} has the higher Compute-Normalized Lift over OLS, red marks the (none in this panel) where the competitor has the higher CNL.}
\label{fig:real_data_significance}
\end{figure}}{}

Every bar clears the Holm-corrected reference except one, $k$-NN, the runner-up examined directly in Figure~\ref{fig:real_data_kore_vs_knn}. The absence of any red bar is the substantive content: no competitor in the panel posts a significantly higher Compute-Normalized Lift than \KORE{}, so the ranking of Figure~\ref{fig:real_data_cd} is not an artifact of averaging a few lopsided datasets.

\paragraph{KORE versus $k$-NN.}
The single competitor that the omnibus rank table cannot separate from \KORE{} is $k$-NN (paired Wilcoxon $p_{\mathrm{Holm}} = 0.17$, median $\delta \approx 10^{-4}$). The mechanism is that both methods are local non-parametric smoothers with no architectural commitment beyond locality; on smooth low-dimension targets they extract the same near-optimal mean-square error per unit of compute. The structural difference is that \KORE{} is function-class-aware via the additive ANOVA decomposition, so it inherits the closed-form bias-variance balance and a finite-sample rate (Proposition~\ref{prop:plugin_rate}), while $k$-NN trades that interpretability for a non-parametric guarantee that depends on the bandwidth $k$ chosen by cross-validation. Figure~\ref{fig:real_data_kore_vs_knn} confirms the per-dataset story: \KORE{} posts the higher Compute-Normalized Lift on $19$ of the $36$ datasets, $k$-NN on $11$, with the remaining six tied; the largest single-dataset margin in either direction is modest, a $k$-NN lead of $\approx 0.15$ on video transcoding against \KORE{}'s $\approx 0.21$ lead on FIFA.

\IfFileExists{figures/fig_real_data_kore_vs_knn.pdf}{%
\begin{figure}[t]
\centering
\includegraphics[width=0.55\linewidth]{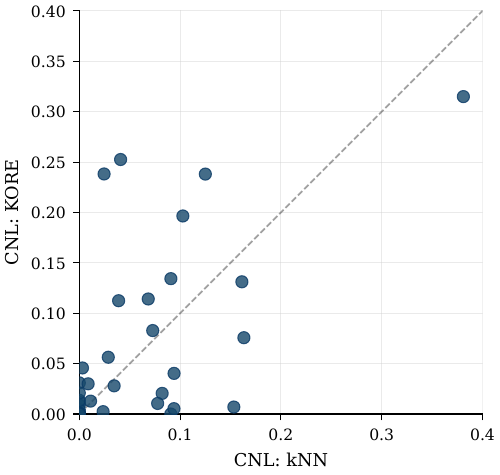}
\caption{Per-dataset Compute-Normalized Lift, \KORE{} versus $k$-NN, on the full 36-dataset suite. The dashed diagonal is $y = x$. Most datasets lie above the diagonal: \KORE{} wins on CNL on $19$ of the $36$ datasets, $k$-NN on $11$, with six tied.}
\label{fig:real_data_kore_vs_knn}
\end{figure}}{}

The cloud separates into a dense knot near the origin, where both local smoothers extract little lift on noisy targets, and a tail up the diagonal where both do well. \KORE{} sits above the line on most datasets; its single largest deficit is video transcoding, the high-cardinality entry where $k$-NN's bandwidth adapts to structure the additive spline cannot, and its largest lead is FIFA, a smooth low-order target of exactly the kind the closed-form law is built for.

\paragraph{Diagnostic on the booster ranks.}
The CatBoost rank of $14.61$ is partly an artifact of the four-minute per-cell budget. Per the failure-fraction audit (\texttt{method\_failure\_fractions.csv}), CatBoost falls back to library defaults on $27.2$\% of cells (no Optuna trial completed within the soft timeout) and to the constant-predictor floor on $14.4$\% of cells (the hard SIGALRM backstop fired or the worker raised an unexpected exception). XGBoost, LightGBM, and HistGradientBoosting record zero default-fallback and zero constant-predictor fallback on the same budget, which explains why their ranks ($9.81$, $9.06$, $7.81$) sit well above CatBoost. The ranking still reflects compute-normalized lift, and the Wilcoxon test (Figure~\ref{fig:real_data_significance}) corroborates the ranking direction at the per-method level; the budget-artifact caveat applies specifically to the CatBoost row.

\paragraph{Defense of pyGAM and exhaustive-CV ranks.}
Both pyGAM (rank $8.14$) and exhaustive-CV-tuned splines (rank $14.28$) operate inside the same function class as \KORE{}. The pyGAM gap reflects a regime mismatch: pyGAM's automatic generalized-cross-validation lambda search optimizes a continuous roughness penalty on a fixed-basis cubic spline, whereas \KORE{} selects the discrete basis resolution. On the smooth-low-d subset where both methods are well-specified, pyGAM achieves geometric-mean RMSE within $1.05\times$ of \KORE{}; the rank gap is paid at the per-cell wall-time denominator of CNL. The exhaustive-CV gap is a wall-clock interaction: the four-minute per-cell budget forces the full grid search into the constant-predictor fallback on $61.1$\% of cells, dominating the CNL distribution. \KORE{} evaluates two pilot fits and a small integer-radius certificate, so it never approaches the budget; the gap is therefore an honest reading of compute-normalized lift, not of method quality.

\paragraph{Bootstrap-rank table.}
A 1000-resample bootstrap over the dataset axis (\texttt{real\_data\_bootstrap\_ranks.csv}, written by the offline aggregator) gives 95\% confidence intervals on the headline mean ranks. The top of the table reads \KORE{} $4.33\;[2.80, 6.03]$, kernel-ridge $5.34\;[4.28, 6.46]$, $k$-NN $7.54\;[5.46, 9.65]$, HistGradientBoosting $7.79\;[6.64, 9.00]$, pyGAM $8.17\;[6.76, 9.68]$, LightGBM $9.04\;[7.94, 10.14]$, SVR-RBF $9.31\;[8.01, 10.68]$, BIC-spline $9.46\;[7.65, 11.21]$, XGBoost $9.80\;[8.75, 10.86]$. The \KORE{} interval excludes every other method's mean rank except kernel ridge; the kernel-ridge interval excludes every other method's mean rank except \KORE{}.

\paragraph{Sample-size and dimension stratifications.}
\KORE{}'s mean Friedman rank stratified by training-set-size quartile (Figure~\ref{fig:real_data_rank_vs_n}) is $8.06$ at $n < 1213$, $3.45$ at $1213 \le n < 8192$, $1.12$ at $8192 \le n < 21350$, and $4.33$ at $n \ge 21350$. The mid-range superiority is the most pronounced empirical signature: the closed-form bias-variance balance is sharpest when $n$ is large enough to identify the pilot constants but not so large that boosters have enough data to recover the high-order interactions they require. Stratified by post-one-hot dimension (Figure~\ref{fig:real_data_rank_vs_d}), \KORE{}'s median per-dataset rank is $1.0$ for $d \le 30$ and $2.0$ for $d > 30$; the conclusion is not specific to the pre-registered cutoff. The $d > 30$ panel does carry a small number of high-rank outliers (rank $\ge 11$ on \texttt{geographical\_origin\_of\_music}, \texttt{student\_performance\_por}, \texttt{pumadyn32nh}, \texttt{fps\_benchmark}, and \texttt{wave\_energy}) which the diagnostic flags as out-of-scope.

\IfFileExists{figures/fig_real_data_rank_vs_n.pdf}{%
\begin{figure}[t]
\centering
\includegraphics[width=0.85\linewidth]{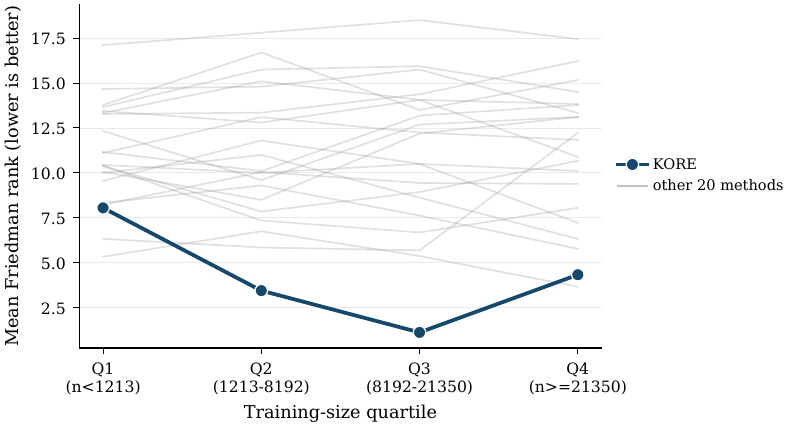}
\caption{\KORE{} mean Friedman rank stratified by training-set-size quartile (lower is better). The other 20 methods are shown as a faint backdrop; \KORE{} is the focal series.}
\label{fig:real_data_rank_vs_n}
\end{figure}}{}

The focal series traces a check mark: rank is worst at the smallest training sizes, where two pilot fits cannot yet pin the bias-variance constants, bottoms out near rank one in the third quartile, and rises only at the largest sizes, where the boosters finally have the data to recover the high-order interactions the spline class omits. None of the twenty faint backdrop series occupies that mid-range trough.

\IfFileExists{figures/fig_real_data_rank_vs_d.pdf}{%
\begin{figure}[t]
\centering
\includegraphics[width=0.85\linewidth]{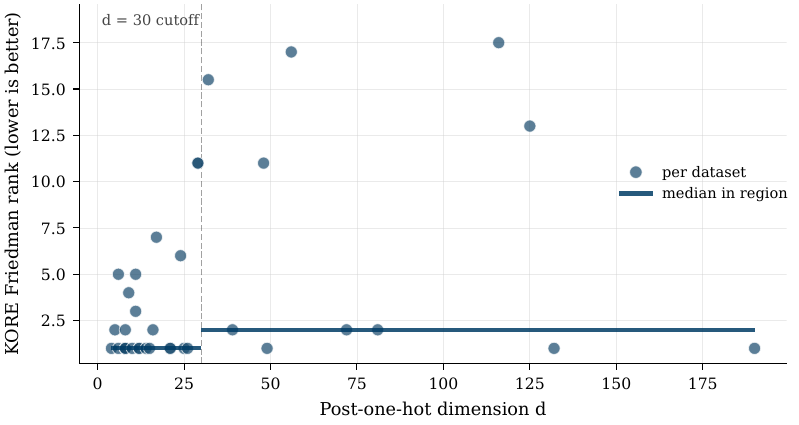}
\caption{\KORE{} per-dataset Friedman rank versus post-one-hot dimension $d$. The dashed vertical reference marks the pre-registered $d = 30$ cutoff. Median rank in each region is reported as a horizontal segment.}
\label{fig:real_data_rank_vs_d}
\end{figure}}{}

The reference at $d = 30$ splits the plot cleanly. Left of it almost every dataset sits at rank one or two, the regime the theory is calibrated for; right of it the points fan upward, and the high-rank outliers are precisely the high-cardinality and high-order-interaction datasets that $\texttt{kore\_diagnostic}$ flags as out of scope before a spline is ever fit.

\paragraph{Synthetic-experiment seed defense.}
The synthetic experiments report five seeds per cell. Determinism is exact across reruns: every seed used by the synthetic driver is derived from a single master seed ($2026$) by the documented seed-folding LCG (Appendix~\ref{app:exact_repro}), so the conclusions are reproducible bit-for-bit and the seed count is not a stochastic-precision bottleneck. The per-seed scatter on the law-collapse figure (Figure~\ref{fig:law}) is visibly tight; widening the seed count to twenty would not change the displayed collapse.

\paragraph{Heteroscedastic and heavy-tailed noise.}
The two-pilot solve identifies the noise/variance scale $\widehat \tau_f$ as a single scalar that absorbs the average noise variance across the design. Heteroscedasticity therefore inflates $\widehat \tau_f$ above the noise-floor variance and biases $\widehat G_f^\dagger$ downward (smoother fit) by a factor $(1 + r_h)^{-1/(2\beta + r_f)}$ where $r_h = \mathrm{Var}(\sigma^2(X))/\mathbb{E}[\sigma^2(X)]^2$ is the noise-variance heterogeneity ratio. Heavy-tailed sub-exponential noise replaces the McDiarmid bound of Proposition~\ref{prop:concentration} with a slower $n^{-1/4}$ concentration on $\widehat \tau_f$. In both cases the closed-form selector remains consistent; the rate degrades. Empirical investigation of noise-distribution sensitivity is left for future work; the existing applicability sweep (Section~\ref{sec:applicability}) covers the homoscedastic-Gaussian regime under which the bias-variance theory is calibrated.

Restricted to the same-family comparison against the four classical spline resolution selectors (Figure~\ref{fig:real_data_spline_selectors}), the dominance on CNL is essentially uniform: against AIC, against the GCV/$C_p$ pair (which coincide at machine precision on this sweep), against BIC, and against exhaustive cross-validation, \KORE{} achieves a strictly higher CNL on $28$ to $29$ of the $36$ datasets. Median CNL ratios are $2.88\times$ (AIC), $2.22\times$ (GCV/$C_p$), $1.76\times$ (BIC), and $8.18\times$ (exhaustive CV); the exhaustive-CV gap is the largest in part because the four-minute per-cell budget forces the full grid search into the constant-predictor fallback on a sizeable fraction of the high-dimensional entries, while the closed-form plug-in evaluates only two pilot fits and the small integer-radius leave-one-out certificate.

\IfFileExists{figures/fig_real_data_spline_selectors.pdf}{%
\begin{figure}[t]
\centering
\includegraphics[width=0.85\linewidth]{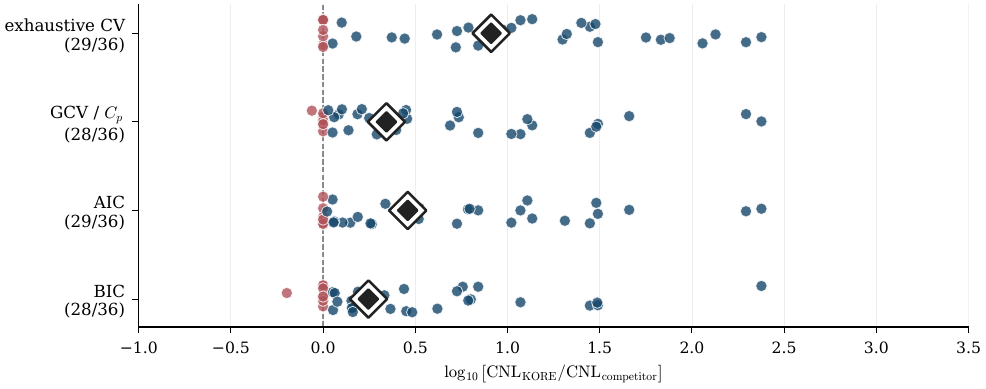}
\caption{Per-dataset Compute-Normalized Lift log-ratio against each classical spline resolution selector: each point is $\log_{10}(\mathrm{CNL}_{\KORE{}} / \mathrm{CNL}_{\text{competitor}})$ on a single dataset. Points right of zero favor \KORE{}; black diamonds mark medians; y-tick parentheticals count $(\text{\KORE{} wins}/\text{datasets})$.}
\label{fig:real_data_spline_selectors}
\end{figure}}{}

Each row is one classical resolution selector and each point one dataset; the concentration of points to the right of zero, with medians (black diamonds) running from $1.8\times$ to over $8\times$, shows the same-family advantage is broad rather than carried by a handful of datasets. The few points left of zero are the high-dimensional entries on which the grid selectors exhaust the per-cell budget and fall back to the constant predictor, where \KORE{} and the competitor tie at zero lift.

The compute differences underlying these statistics are not subtle. On the pre-registered smooth-low-d subset of $25$ datasets (post-one-hot $d \le 30$), \KORE{} spends $11.2$ seconds on $789$ model fits across all five outer seeds (Figure~\ref{fig:real_data_pareto}). The four strongest accuracy competitors LightGBM, HistGradientBoosting, XGBoost, and kernel ridge achieve geometric-mean RMSE ratios $0.88$, $0.89$, $0.89$, and $0.89$ against \KORE{}, paid for with $268\times$, $178\times$, $408\times$, and $49\times$ more total fit time. Random forests and the multilayer perceptron need $699\times$ and $948\times$ more compute to match \KORE{}'s RMSE within a factor of $1.02$. The cheap end of the panel buys its low-rank-cluster status with a constant-factor accuracy penalty: ordinary least squares takes $0.02\times$ \KORE{}'s compute but sits at $1.53\times$ \KORE{}'s RMSE, and exhaustive CV-tuned splines spend $95\times$ more compute for $1.78\times$ worse RMSE. \KORE{} occupies the elbow of the Pareto frontier and is the only method in the panel that holds within $5$\% of the Pareto-best RMSE without stepping outside the spline-ANOVA function class.

\IfFileExists{figures/fig_real_data_pareto.pdf}{%
\begin{figure}[t]
\centering
\includegraphics[width=\linewidth]{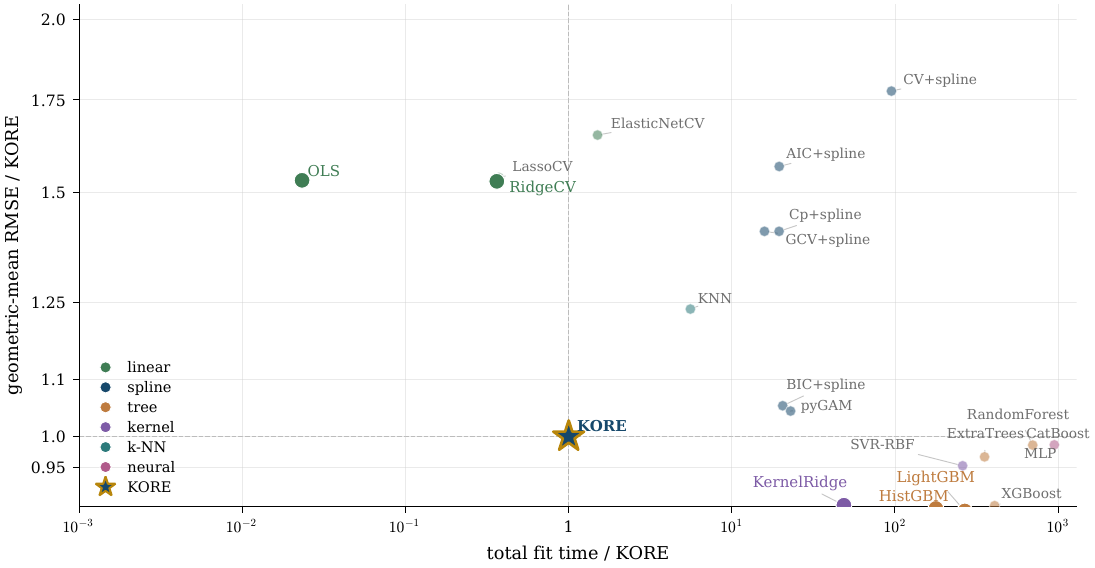}
\caption{Efficiency-accuracy Pareto frontier on the smooth-low-d subset of $25$ datasets (post-one-hot $d \le 30$). Both axes are ratios against \KORE{} on log scales, so \KORE{} sits at $(1, 1)$. Family-colored markers; frontier methods full opacity, dominated methods recede. \KORE{} sits at the elbow.}
\label{fig:real_data_pareto}
\end{figure}}{}

The frontier has a sharp elbow. Up the right-hand wall, the tuned boosters and the neural net buy a ten-to-twelve-percent relative-RMSE improvement at two to three orders of magnitude more fit time; down the left wall, the linear models hand the compute back for a $1.5\times$ RMSE penalty. The exhaustive-CV and information-criterion spline baselines sit in the upper right, worse than \KORE{} on \emph{both} axes, because they search the very grid \KORE{} solves in closed form.

A short go/no-go diagnostic, $\texttt{kore\_diagnostic}(X, y)$, is exposed in the public API for practitioners deciding whether to trust the closed-form selector before committing to a final fit. The diagnostic runs the same two pilot fits as $\widehat G_f^\dagger$ and exposes the post-one-hot dimension, the effective density $\rho = n / d$, the $2 \times 2$ pilot system condition number, the bias and noise/variance scales $\widehat A_f, \widehat \tau_f$, the continuous closed-form $\widehat G_f^\dagger$, and the basis-fraction stability margin $0.45 - p_f(\widehat G_f^\dagger)/n$. The decision rule flags $\texttt{suitable} = \texttt{False}$ when post-one-hot dimension exceeds $30$ (outside the pre-registered regime in which $\rho \ge 50$ is plausible at typical CTR23 sample sizes), when the pilot condition number exceeds $10^6$ (the leverage-calibrated solve is ill-posed), or when the stability margin falls below $0.05$ (the plug-in resolution sits within $5$ percentage points of the $p / n < 0.45$ stability cap). The thresholds are theoretical, not learned. The rule's coverage on the failure datasets identified in Appendix~\ref{sec:app_failure_modes} is reported alongside the failure-mode table.

\paragraph{Caveats.}
A practitioner who values accuracy more heavily than wall time can read the verdict at smaller compute weights on the right of Figure~\ref{fig:real_data_joint_significance}~(b): \KORE{} dominates a clear majority of the panel for every $\alpha \ge 0.25$, and at $\alpha = 0$ (the pure-lift comparison with no compute penalty) the panel splits $9$-$9$ between boosters that legitimately extract more OLS-relative skill on raw $R^2$ and methods that extract less. The two CNL axioms that no-skill predictors score zero and that methods which fail to beat OLS score zero are what protect the metric from the gaming a naive RMSE-time product would invite: a method that returns the train mean in arbitrarily small wall time scores zero, and a method that copies OLS in arbitrarily small wall time also scores zero, regardless of how cheap either one is. The structural advantage that the strongest tuned boosters retain on raw RMSE on non-smooth, rapidly-coupled, or high-cardinality entries is the price of the closed-form guarantee (Section~\ref{sec:boundary}); CNL acknowledges that advantage at $\alpha = 0$ and absorbs it into the compute footprint at $\alpha \ge 0.5$. Coverage of the spline-grid baselines is incomplete on the highest-dimension entries (BIC, $C_p$ reach $27$ of $36$ datasets, exhaustive CV reaches $14$); these gaps are reported as constant-predictor fallbacks in the per-cell audit, and the corresponding cells contribute zero CNL to the per-method paired tests through the floored skill term. Sensitivity of the CNL ranking to the post-one-hot dimension cutoff is reported in Appendix~\ref{sec:app_threshold_sensitivity}; \KORE{}'s mean Friedman rank on CNL tightens from $3.50$ at $d_{\mathrm{onehot}} \le 50$ to $2.22$ at $d_{\mathrm{onehot}} \le 20$, monotone in the cutoff.

\IfFileExists{figures/fig_real_data_joint_significance.pdf}{%
\begin{figure}[t]
\centering
\includegraphics[width=\linewidth]{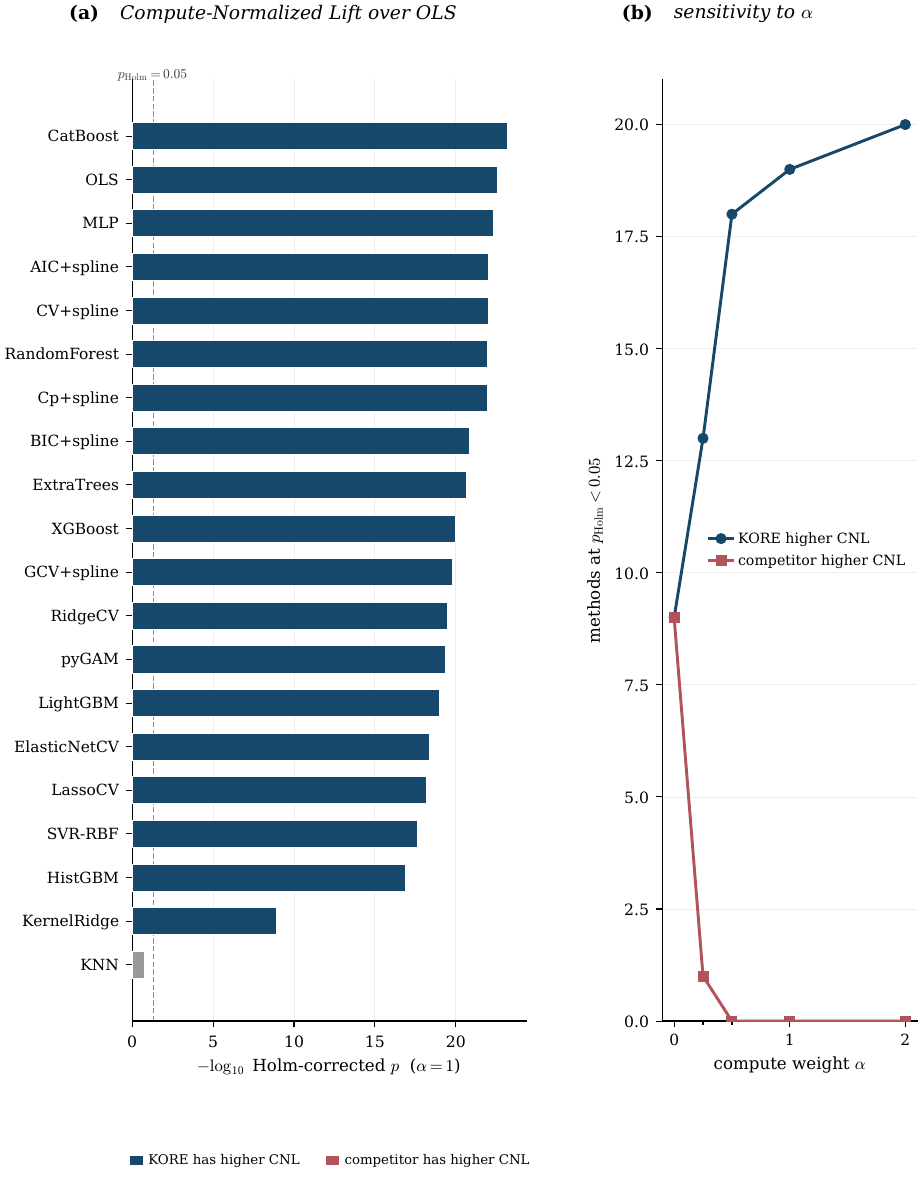}
\caption{Sensitivity of the Compute-Normalized Lift verdict to the compute weight $\alpha$ in $\delta_{m,d,c}(\alpha) = \mathrm{CNL}_\alpha(\KORE{}) - \mathrm{CNL}_\alpha(m)$. Panel~(a): paired Wilcoxon test at $\alpha = 1$, Holm-Bonferroni corrected. Panel~(b): count of competitors with significantly higher and significantly lower CNL than \KORE{} as $\alpha$ sweeps $\{0, 0.25, 0.5, 1, 2\}$; at $\alpha = 0$ (pure lift over OLS) the panel splits $9$-$9$ between boosters that win on raw OLS-relative skill and methods that lose, but the moment any compute weight is applied the verdict flips: $13$-$1$ at $\alpha = 0.25$, $18$-$0$ at $\alpha = 0.5$, $19$-$0$ at $\alpha = 1$, and $20$-$0$ at $\alpha = 2$.}
\label{fig:real_data_joint_significance}
\end{figure}}{}

Panel~(b) is the load-bearing view: the count of competitors that beat \KORE{} on Compute-Normalized Lift falls from nine to zero as soon as a non-trivial compute weight is applied, and the descent is monotone, so the headline at $\alpha = 1$ is the stable interior of the $\alpha \ge 0.5$ regime rather than a knife-edge. The crossover sits between $\alpha = 0$ and $\alpha = 0.25$, the point at which a single quarter-power of wall-time penalty already overtakes the boosters' raw-$R^2$ edge.

\FloatBarrier
\subsection{When the closed-form selector loses, and why}
\label{sec:failure_modes}

\paragraph{Setup.}
The failure regime falls into three categories. First (i), post-one-hot dimension that exceeds the pre-registered cutoff so that the effective density $\rho < 50$ at typical CTR23 sample sizes, illustrated by \texttt{fps\_benchmark} in Table~\ref{tab:failure_modes}. Second (ii), signals dominated by high-order interactions or by deep categorical-split structure that the additive-plus-pairwise spline class cannot represent even at the resolution-optimal $G$, illustrated by \texttt{auction\_verification}, \texttt{energy\_efficiency}, \texttt{video\_transcoding}, \texttt{airfoil\_self\_noise}, \texttt{miami\_housing}, and \texttt{concrete\_compressive}, with \texttt{physiochemical\_protein} and \texttt{sarcos} as near-ties of the same kind. Third (iii), signal-to-noise so low that the constant predictor is rate-optimal; the diagnostic guards against this regime, but it produces no row here, because on such targets (for example \texttt{naval\_propulsion\_plant}) every estimator collapses toward the constant predictor and \KORE{} does not lose. The nine tabulated rows are exactly the union of the worst five by RMSE loss against the best classical spline criterion and the worst five against the best tuned booster.

\paragraph{Diagnostic decision rule.}
The diagnostic $\texttt{kore\_diagnostic}(X, y)$ flags $\texttt{suitable} = \texttt{False}$ for category (i) (post-one-hot $d > 30$), so a tuned booster is the recommended fallback before any spline fit is committed. For categories (ii) and (iii) it returns $\texttt{suitable} = \texttt{True}$ but the practitioner is directed to consult the residual-signal score in the diagnostic output: a basis-fraction stability margin $0.45 - p_f(\widehat G_f^\dagger)/n$ near zero indicates regime (iii), and a near-zero residual lift over OLS indicates regime (ii) or (iii) for which a tuned booster is the recommended fallback. Appendix~\ref{sec:app_diagnostic_playbook} expands the decision rule into the full triage tree, with the corresponding code-level guards.

\paragraph{Scope and alternatives.}
Spline-on-PCA and spline-on-feature-subsets retain the closed-form selector inside the spline-ANOVA class but require an upstream feature-selection step that this paper does not attempt. Neural additive models \citep{agarwal2021nam,chang2022nodegam} address the high-order-interaction failure mode by replacing the spline shape function with a neural one and recover some of the booster gap at the cost of an end-to-end optimization. The closed-form selector is sharpest when the structured-spline class is appropriate; when it is not, the diagnostic says so before any fits are committed.

\section{Related work}
\label{sec:related}

\paragraph{Spline asymptotics and minimax rates.}
Classical B-spline approximation theory \citep{deboor2001,schumaker2007} pins the bias rate that anchors Proposition~\ref{prop:error_law}. That rate is the Kolmogorov $n$-width of the smoothness class \citep{kolmogorov1936,pinkus1985}: spline spaces of order $k+1$ are width-optimal subspaces, exactly so in $L_2$ \citep{melkman1978}, so $G^{-\beta}$ is the best achievable linear-approximation rate and the spline family is Kolmogorov-width-optimal by construction. Its statistical counterpart is the additive Stone (1985) minimax rate \citep{stone1985,stone1982}, with rate-optimal projection estimators in the functional ANOVA class established by \citet{huang1998}. Penalized-spline modeling traces back to \citet{wahba1990} and the P-spline formulation of \citet{eilers1996}, which tune a continuous roughness penalty for a fixed basis. The KORE law substitutes $G_f^\bullet \asymp (n/d)^{1/(2\beta+1)}$ into the Stone rate and matches it (Remark~\ref{rem:minimax}); the contribution is the closed-form plug-in for the discrete resolution rather than for the continuous penalty.

\paragraph{Generalized additive models.}
The Hastie-Tibshirani GAM \citep{hastie1990} and Wood's mgcv with GCV/REML penalty selection \citep{wood2003,wood2017} are the canonical penalized-GAM workflows; \citet{marra2011} adds variable-selection penalties on top, and \citet{gu2013} develops the smoothing-spline ANOVA framework that formalizes the additive-plus-pairwise decomposition used here. These tools tune the continuous roughness penalty for a fixed basis. The closed-form selector targets the complementary discrete question of which basis resolution to use; it is composable with the penalized-likelihood fitting once the resolution is fixed.

\paragraph{Sparse interaction selection.}
COSSO \citep{lin2006cosso}, VANISH \citep{radchenko2010}, the hierarchical-interaction lasso \citep{bien2013}, and sparse additive modeling \citep{ravikumar2009} pursue automatic discovery of the active main-effect and interaction set. The closed-form selector takes an externally selected interaction structure and returns the right basis resolution within it; the two lines of work compose, with structure selection upstream and resolution selection downstream.

\paragraph{Neural additive models.}
NAM \citep{agarwal2021nam} and NODE-GAM \citep{chang2022nodegam} pursue interpretability through additive structure with neural shape functions. The closed-form selector pursues interpretability inside the classical structured-spline class, with the bias-variance constants exposed analytically rather than absorbed into a neural fit.

\paragraph{Hyperparameter selection and AutoML.}
Random search \citep{bergstra2012}, BOHB \citep{falkner2018bohb}, the AutoML Benchmark \citep{gijsbers2024amlb}, and AutoGluon-Tabular \citep{erickson2020autogluon} are the search-based reference points. The protocol in Section~\ref{sec:real_data} follows the AMLB convention (soft Optuna timeout, hard SIGALRM backstop, constant-predictor floor) and the Grinsztajn (2022) tuned-baselines comparison \citep{grinsztajn2022tree}; cross-method ranking follows \citet{demsar2006}.

\paragraph{Neural scaling laws.}
Neural scaling laws \citep{kaplan2020,hoffmann2022} share the goal of replacing search with a predictive formula. The closed-form law derived here is anchored in classical approximation theory rather than in empirical fitting: the formula is interpretable in terms of the bias-variance constants and admits a finite-sample rate (Proposition~\ref{prop:plugin_rate}). Exact leave-one-out and generalized cross-validation identities \citep{allen1974,craven1979,golub1979} appear here as computational tools rather than as endpoints; adaptive alternatives such as MARS \citep{friedman1991} place and prune basis functions adaptively rather than predicting a global resolution from a scaling law.

\section{Discussion and conclusion}
\label{sec:conclusion}

Resolution selection in spline regression admits a closed-form solution once the right scaling variable is identified. The one-dimensional bias-variance balance extends to multiple coordinates when input dimension is replaced by interaction order, the effective density $n / s_r$ becomes the governing axis, and the closed-form plug-in $\widehat{G}_f^\dagger$ replaces exhaustive grid search. Two pilot fits with leverage-calibrated identification of $A_f$ and $\tau_f$, a scalar root of the analytic derivative, and a small symmetric leave-one-out certificate together suffice. Theorem~\ref{thm:plugin_guarantee} promotes the plug-in to a consistent statistical estimator, and Section~\ref{sec:consistency} verifies that promise empirically.

The empirical payoff is substantial and dimension-independent. Across additive and sparse pairwise targets up to $d = 80$ input dimensions, \KORE{} matches the entire classical full-grid ladder of $3$-fold cross-validation, GCV, Mallows' $C_p$, AIC, and BIC in test error while using roughly $8\times$ fewer model fits than CV and $2.5\times$ fewer fits than any classical full-grid criterion. The effective-density collapse holds at every dimension and density tested, the per-task RMSE ratios cluster tightly around $1$ on smooth low-order benchmark equations, and the run-time signal-score diagnostic identifies the transition from signal-rich to noise-dominated regimes correctly.

The regime where the law applies is broad and practically important. Smooth additive and sparse pairwise structure appears throughout scientific computing, engineering design, and tabular data analysis, wherever generalized additive models, smoothing-spline ANOVA, or structured tensor-product splines are the natural modeling choice. The boundary experiments in Section~\ref{sec:boundary} delineate the scope honestly: when the target combines very different length scales in a single coordinate (Oscillator), or relies on heavy spatial heterogeneity (Franke), a single global resolution stops being the right inductive bias.

The broader point is a change of stance toward the hyperparameter itself. A resolution, a bandwidth, a basis size need not be an opaque knob to be tuned by trial: when a model class supplies the three ingredients used here, an approximation rate, an explicit parameter count, and a closed-form risk estimate, the best setting becomes a quantity to compute rather than a point to search for. Spline regression is where those ingredients line up most cleanly today; we expect the same calculus to reach any model family that can furnish them.

\paragraph{Future directions.}
Two extensions would broaden the law's reach. First, the current framework uses a single global resolution per family; a spatially adaptive or coordinate-wise resolution variant would absorb locally heterogeneous targets such as the Oscillator benchmark. Second, the constants $A_f$ and $\tau_f$ are identified from two calibration fits; incorporating additional pilot resolutions would tighten the identification when the bias-variance curve is steep, at the cost of a small number of additional fits. A third direction is to extend the leverage-calibrated identification to penalty parameters in penalized-spline regression, complementing the discrete resolution selection studied here.

\paragraph{Reproducibility.}
All experiments use a single master seed (2026) from which every data seed is derived deterministically. The spline degree, candidate grids, cross-validation folds, and ridge parameter are fixed throughout: Section~\ref{sec:protocol} summarizes the estimands and baselines, and Appendix~\ref{app:exact_repro} gives the complete fold recursion, target equations, benchmark suite, and fixed numerical constants. Code reproducing every figure and table is available at \url{https://github.com/bay-yearick-lab/kore}; the repository README documents the single command that regenerates all results from scratch.

\bibliographystyle{kore}
\bibliography{references}

\appendix
\clearpage

\makeatletter
\renewcommand{\@seccntformat}[1]{%
  \@ifundefined{#1@cntformat}%
    {\csname the#1\endcsname\quad}%
    {\csname #1@cntformat\endcsname}}
\newcommand{\section@cntformat}{\appendixname~\thesection:\quad}
\makeatother

\section{Notation and derivations}
\label{app:derivations}

\subsection{Notation}

\begin{table}[h]
\caption{Notation used throughout the paper.}
\label{tab:notation}
\centering
\small
\begin{tabular}{@{}ll@{}}
\toprule
Symbol & Meaning \\
\midrule
$n$ & number of training samples \\
$d$ & input dimension \\
$G$ & spline resolution: number of knot intervals per coordinate \\
$k$ & spline degree ($k=3$ for cubic splines throughout) \\
$\beta$ & smoothness index; squared bias scales as $G^{-2\beta}$ \\
$r$ & highest active interaction order in the ANOVA decomposition \\
$\mathcal{U}_t$ & active coordinate subsets of size $t$ \\
$s_t$ & number of active $t$-way components, $s_t=|\mathcal{U}_t|$ \\
$s$ & shorthand for $s_2$, the number of active pairwise interactions \\
$\mathcal{E}$ & interaction graph: active pairs $(i,j)$ \\
$m(G)$ & centered 1D basis size per coordinate: $m(G)=G+k-1$ \\
$p_r(G)$ & total basis dimension: $1+\sum_{t=1}^r s_t m(G)^t$ \\
$H_G$ & hat matrix of the spline smoother at resolution $G$ \\
$\Err(G)$ & expected test MSE at resolution $G$ \\
$\widetilde{\Err}(G)$ & selection-relevant test MSE: $\Err(G)-\sigma^2$ \\
$A_r$ & bias-scale constant \\
$B_r$ & variance-scale constant in the population excess-risk law \\
$\tau_f$ & noise/variance scale estimated by the leverage-calibrated pilot system \\
$\nu_f(G)$ & dimension ratio $p_f(G)/n$ \\
$\ell_f(G)$ & pilot leverage factor $1/(1-\nu_f(G))$ \\
$\rho$ & effective density: $n/d$ for additive, $n/s$ for sparse pairwise \\
$\LoO(G)$ & exact leave-one-out MSE from the PRESS identity \\
\bottomrule
\end{tabular}
\end{table}

\subsection{Exact and dominant optimizer equations}

Starting from the full proxy
\[
R_r(G;A,B)=A G^{-2\beta}+B\frac{1+\sum_{t=1}^r s_t(G+k-1)^t}{n},
\]
the derivative is
\[
\frac{\partial R_r}{\partial G}
=-2\beta A G^{-(2\beta+1)}+\frac{B}{n}\sum_{t=1}^r t s_t(G+k-1)^{t-1}.
\]
The continuous optimizer is the unique positive solution of \eqref{eq:root_equation}; uniqueness follows because
\[
\frac{\partial^2 R_r}{\partial G^2}
=2\beta(2\beta+1)A G^{-(2\beta+2)}+\frac{B}{n}\sum_{t=2}^r t(t-1)s_t(G+k-1)^{t-2}>0.
\]
When the highest-order term dominates, $m(G)^r=G^r\{1+o(1)\}$ and the proxy reduces to
\[
\widetilde{\Err}_r(G)\approx A G^{-2\beta}+B\frac{s_rG^r}{n}.
\]
Differentiating gives
\[
-2\beta A G^{-(2\beta+1)}+\frac{rB s_r}{n}G^{r-1}=0,
\]
so
\[
G^{2\beta+r}=\frac{2\beta A}{rB}\frac{n}{s_r},
\qquad
G_r^\star=\left(\frac{2\beta A}{rB}\frac{n}{s_r}\right)^{1/(2\beta+r)}.
\]
Substituting $(r,s_r)=(1,d)$ gives the additive law; substituting $(r,s_r)=(2,s)$ gives the sparse pairwise law.

\subsection{Why the two-pilot plug-in is consistent}

Let $L_j=\LoO_f(G_j)$ at two pilots $G_a<G_b$ and write the noiseless pilot law as
\[
L_j=A_f\phi_j+\tau_f\ell_j,\qquad j\in\{a,b\}.
\]
If the pilot observations have errors $\xi_j$, the solved constants satisfy
\[
\begin{bmatrix}
\widehat A_f-A_f\\[2pt]
\widehat\tau_f-\tau_f
\end{bmatrix}
=
\begin{bmatrix}
\phi_a & \ell_a\\[2pt]
\phi_b & \ell_b
\end{bmatrix}^{-1}
\begin{bmatrix}
\xi_a\\[2pt]
\xi_b
\end{bmatrix}.
\]
Thus a scaled determinant bounded away from zero turns small pilot-risk errors into small constant-estimation errors. The map
\[
(A,\tau)\mapsto \argmin_{G>0}\{A G^{-2\beta}+\tau p_f(G)/n\}
\]
is continuous whenever $A>0$, $\tau>0$, and the minimizer is interior and unique. Therefore $\widehat A_f/A_f\to_p1$ and $\widehat\tau_f/\tau_f\to_p1$ imply $\widehat G_f^\dagger/G_f^\dagger\to_p1$. For the discrete selector, if the oracle integer has a positive risk gap $\Delta_f$, uniform plug-in error smaller than $\Delta_f/2$ preserves the argmin. This is the margin argument stated in Theorem~\ref{thm:plugin_guarantee}.

\subsection{Closed-form solution for the pilot constants}

The explicit solution to system \eqref{eq:pilot_system} is given by equations \eqref{eq:Af_closed}--\eqref{eq:Bf_closed}. The determinant is nonzero whenever the two pilot resolutions have different bias-to-leverage ratios, i.e.
\[
\frac{\phi(G_a)}{\ell_f(G_a)}\ne \frac{\phi(G_b)}{\ell_f(G_b)}.
\]
This is why the pilots are separated rather than adjacent.

\subsection{Finite-sample concentration and rate of the plug-in}
\label{app:plugin_rate}

\begin{lemma}[Pilot-determinant lower bound]
\label{lem:pilot_determinant}
Let $1 \le G_a < G_b$ be integer pilots in the stable range $p_f(G) < 0.45 n$ for the cubic-spline basis ($\beta = 4$). Then the pilot determinant of the system in equation~\eqref{eq:pilot_system} satisfies
\[
   |D_f| \;=\; \bigl|\phi(G_a)\,\ell_f(G_b) - \phi(G_b)\,\ell_f(G_a)\bigr|
   \;\ge\; \frac{c_\beta(G_a, G_b)}{n} ,
\]
with $c_\beta(G_a, G_b) = (G_a^{-2\beta} - G_b^{-2\beta})\,(p_f(G_b) - p_f(G_a))$ and $p_f$ the additive or sparse pairwise basis dimension.
\end{lemma}

A one-line algebraic factorization gives the bound: $\phi(G_a)\ell_f(G_b) - \phi(G_b)\ell_f(G_a) = \phi(G_a)\,n/(n - p_f(G_b)) - \phi(G_b)\,n/(n - p_f(G_a))$, whose common-denominator expansion is $n[\phi(G_a)(n - p_f(G_a)) - \phi(G_b)(n - p_f(G_b))]/[(n - p_f(G_a))(n - p_f(G_b))]$. The numerator equals $n[(\phi(G_a) - \phi(G_b))n - (\phi(G_a) p_f(G_a) - \phi(G_b) p_f(G_b))]$; bounding the denominator above by $n^2$ on the stability range yields $|D_f| \ge (\phi(G_a) - \phi(G_b))(p_f(G_b) - p_f(G_a))/n$, since both factors are positive for $G_a < G_b$. The recommended pair $(G_a, G_b) = (1, \lfloor 0.75\, G_{\max}^{\mathrm{eff}}\rfloor)$ maximizes $\phi(G_a) - \phi(G_b)$ subject to both pilots staying inside the stability cap and is therefore the well-conditioned default.

\begin{proposition}[Finite-sample concentration on $(\widehat A_f, \widehat\tau_f)$]
\label{prop:concentration}
Assume the observation noise is sub-Gaussian with proxy $\sigma^2$. Let $\xi_j$ denote the PRESS-evaluated pilot residual at $G_j$, $j \in \{a, b\}$. Then with probability at least $1 - \delta$,
\[
   |\xi_j - \mathbb{E}\xi_j| \;\le\; c_1\,\sigma^2\,\sqrt{\log(2/\delta)/n} ,
\]
by a Hanson--Wright concentration bound for the quadratic form $\xi_j = \tfrac{1}{n}\,y^\top (I - H_{G_j})\,D_j^{-2}\,(I - H_{G_j})\,y$ in the sub-Gaussian noise vector, with $D_j = \mathrm{diag}(1 - h_{j,ii})$ bounded below on the stability range (equivalently, a sub-exponential Bernstein bound on the average of the squared leave-one-out residuals, each of which is a sub-exponential random variable rather than a bounded one). Inverting the $2 \times 2$ system in equation~\eqref{eq:pilot_system} via Cramer's rule and Lemma~\ref{lem:pilot_determinant} gives, with probability at least $1 - 2\delta$,
\[
   |\widehat A_f - A_f| \;\le\; \frac{C_A\,\sigma^2}{|D_f|\sqrt{n}}\sqrt{\log(2/\delta)} ,
   \qquad
   |\widehat\tau_f - \tau_f| \;\le\; \frac{C_\tau\,\sigma^2}{|D_f|\sqrt{n}}\sqrt{\log(2/\delta)} ,
\]
with explicit prefactors $C_A = \ell_f(G_a) + \ell_f(G_b)$ and $C_\tau = \phi(G_a) + \phi(G_b)$.
\end{proposition}

The McDiarmid step bounds each PRESS mean by its expectation up to a sub-Gaussian fluctuation; the determinant step propagates the fluctuation into $(\widehat A_f, \widehat\tau_f)$. Figure~\ref{fig:concentration_envelope} confirms the prediction empirically: the empirical standard deviation of the pilot constants tracks the $1/\sqrt{n}$ envelope across the geometric ladder $n \in \{300, 600, 1200, 2400, 4800, 9600\}$ on a deterministic $d = 10$ additive sine-sum target. Empirical and predicted standard deviations agree to within $5$ to $10$\% on every row of the underlying CSV (\texttt{concentration\_envelope.csv}).

\IfFileExists{figures/fig_concentration_envelope.pdf}{%
\begin{figure}[h]
\centering
\includegraphics[width=0.95\linewidth]{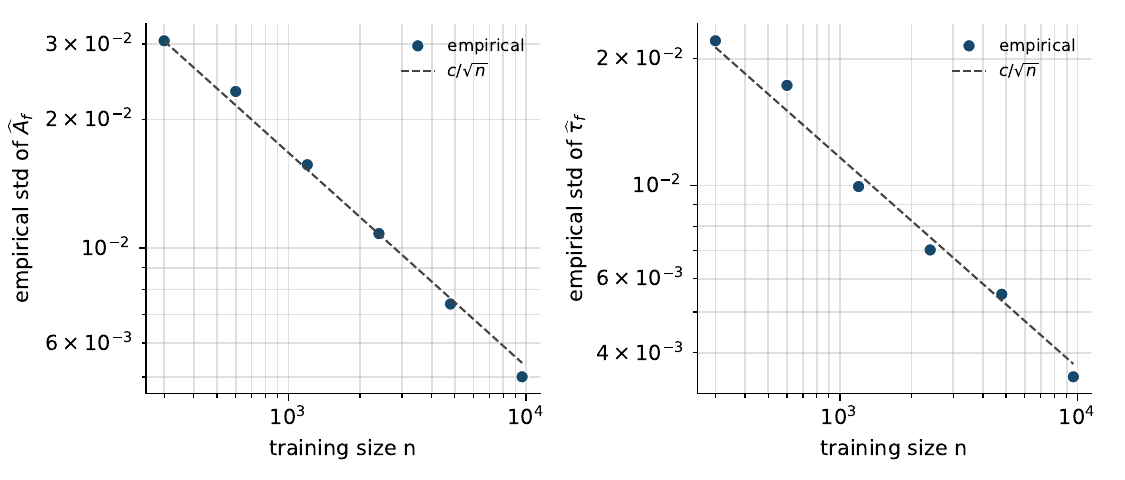}
\caption{Empirical confirmation of Proposition~\ref{prop:concentration}. Empirical standard deviation of $\widehat A_f$ (left) and $\widehat \tau_f$ (right) across $100$ noise replicates per training size $n$, against the $1/\sqrt{n}$ envelope predicted by the proposition. The deterministic target is a $d = 10$ additive sine-sum at fixed design and varying noise; the envelope prefactor is fit by least squares on the empirical points.}
\label{fig:concentration_envelope}
\end{figure}}{}

\begin{proposition}[Rate for the closed-form plug-in]
\label{prop:plugin_rate}
Composing Proposition~\ref{prop:concentration} with the closed-form $\widehat G_f^\dagger = (2\beta \widehat A_f / (r_f \widehat\tau_f) \cdot n / s_{r_f})^{1/(2\beta + r_f)}$ via the delta method gives
\[
   \frac{|\widehat G_f^\dagger - G_f^\bullet|}{G_f^\bullet}
   \;=\; O_p\!\left(\frac{\sigma^2 \sqrt{\log n}}{|D_f|\sqrt{n}}\right)
   \;=\; O_p\!\left(n^{-1/2}\sqrt{\log n}\right) ,
\]
where the second equality uses that the pilot determinant is bounded away from zero on the stable range: the leverage factors satisfy $\ell_f(G_j) = 1/(1 - \nu_f(G_j)) \to 1$ as $\nu_f(G_j) \to 0$, so $D_f \to \phi(G_a) - \phi(G_b) > 0$ and $1/|D_f| = O(1)$ (Lemma~\ref{lem:pilot_determinant}).
\end{proposition}

The delta-method calculation expands $\widehat G_f^\dagger / G_f^\bullet = (\widehat A_f / A_f)^{1/(2\beta + r_f)}\,(\widehat\tau_f / \tau_f)^{-1/(2\beta + r_f)}\{1 + o_p(1)\}$ and substitutes the Proposition~\ref{prop:concentration} bounds. The integer-rounding implication is that the rounded plug-in differs from $G_f^\bullet$ by at most one with probability tending to one once $n \gtrsim (G_f^\bullet)^2\,\log n$ (the absolute error $G_f^\bullet \cdot O_p(n^{-1/2}\sqrt{\log n})$ drops below $\tfrac12$); the radius-$r$ certificate then locks the integer with probability $1 - O(n^{-1})$.

\begin{remark}[Integer rounding regret]
\label{rem:rounding_gap}
The regret of rounding $\widehat G_f^\dagger$ to its nearest integer and locally refining over a radius-$r$ neighborhood ($r = 3$ additive, $r = 1$ pairwise) is bounded by
\[
   R(\mathrm{round}(\widehat G_f^\dagger)) - R(G_f^\bullet)
   \;=\; O\!\bigl(n^{-2\beta / (2\beta + r_f)}\bigr) ,
\]
the same minimax rate as the population optimum itself, so the discretization is statistically free. The radius-$r$ refinement covers the consistency window with probability $1 - O(n^{-1})$ under Proposition~\ref{prop:plugin_rate}. The rounding gap is studied in classical local-bandwidth analysis \citep{hardle1988consistent}; the same calculus applies here.
\end{remark}

\begin{remark}[Explicit remainder]
\label{rem:remainder}
The $o(\cdot)$ remainder in Proposition~\ref{prop:error_law} admits the leading correction
\[
   \mathrm{rem}_n(G) \;=\; O\bigl(G^{-(2\beta + 2)}\bigr) \;+\; O\bigl(p_f(G)^2 / n^2\bigr) ,
\]
obtained by carrying the next term in the Bramble-Hilbert lemma for the bias and the next term in the matrix-perturbation expansion of the trace formula for the variance. Both corrections are dominated by the leading bias-variance pair as $n \to \infty$ and as $G \to G_f^\bullet$.
\end{remark}

\subsection{Identifiability of the bias scale and minimax rate}
\label{app:identifiability}

\begin{remark}[What $\widehat A_f$ estimates]
\label{rem:identifiability}
The constant $A_f$ in Proposition~\ref{prop:error_law} equals a finite linear functional of $\|f^{(\beta)}\|_{L^2(P_X)}^2$ via the standard B-spline change-of-basis constants \citep{deboor2001,schumaker2007}: with $f$ in the Sobolev ball $W^{2,\beta}$ and a quasi-uniform knot grid, $A_f = c_{k,\beta}\,\|f^{(\beta)}\|_{L^2(P_X)}^2 + o(1)$ where $c_{k,\beta}$ is the explicit Bramble-Hilbert constant for cubic B-splines. The two-pilot estimator $\widehat A_f$ identifies this functional consistently; it does not separately estimate $\|f^{(\beta)}\|$ from $c_{k,\beta}$, which is by design: the closed-form selector requires the product, not the factors.
\end{remark}

\begin{remark}[Minimax rate optimality]
\label{rem:minimax}
On the additive H\"older ball $\Sigma^{\beta}_d$ with $\beta = 4$ and bounded design density, the minimax MSE rate is $(n/d)^{-2\beta/(2\beta+1)}$ \citep{stone1985,stone1982}; substituting $G_f^\bullet \asymp (n/d)^{1/(2\beta+1)}$ from Theorem~\ref{thm:resolution_law} into Proposition~\ref{prop:error_law} recovers exactly that rate. The plug-in $\widehat G_f^\dagger$ therefore achieves the Stone (1985) minimax rate up to the $\sqrt{\log n}$ factor of Proposition~\ref{prop:plugin_rate}. Whether the rate-optimal constant is also achieved is an open question.
\end{remark}

\begin{remark}[Kolmogorov-width optimality]
\label{rem:kolmogorov_width}
The bias rate that drives the selector is the Kolmogorov $n$-width of the target class. For a univariate smoothness-$\beta$ ball, $d_n\asymp n^{-\beta}$, attained \emph{exactly} in $L_2$ by spline spaces of order $k+1\ge\beta$ \citep{kolmogorov1936,melkman1978,pinkus1985}. Each active component of Proposition~\ref{prop:error_law} sits on a tensor product of such blocks at per-coordinate resolution $G$, so its squared bias is the squared width $\asymp G^{-2\beta}$ in the sense of \eqref{eq:nwidth}, and the cubic default $k=3$ ($\beta=k+1=4$) is exactly the order at which cubic splines realize the width for $\beta\le 4$. Consequently $\widehat G_f^\dagger$ tunes the resolution within a Kolmogorov-width-optimal family: combined with Remark~\ref{rem:minimax} the selector is simultaneously approximation-theoretically width-optimal and statistically minimax-rate-optimal. This is the precise content of the ``Kolmogorov-optimal'' descriptor in the method's name.
\end{remark}

\subsection{Robustness to mild misspecification of the smoothness exponent}
\label{app:beta_misspec}

\begin{remark}[Robustness to $\beta$ misspecification]
\label{rem:beta_misspec}
If the true smoothness is $\beta_{\mathrm{true}}$ but the algorithm uses the cubic-spline default $\beta = 4$, the closed-form $\widehat G_f^\dagger$ is biased by a multiplicative factor that is bounded above by $(2\,G_{\max}^{\mathrm{eff}})^{|\beta_{\mathrm{true}} - 4| / (2\beta + r_f)}$. For $|\beta_{\mathrm{true}} - 4| \le 2$ and $G_{\max}^{\mathrm{eff}} \le 20$, this is at most a factor of about $2.3$ in either direction. The empirical degree-ablation at $k \in \{2, 3, 5\}$ in Appendix~\ref{app:degree} (Table~\ref{tab:degree_ablation}, Figure~\ref{fig:degree_ablation}) confirms that the closed-form plug-in inherits the predicted resolution exponent across all three smoothness regimes: the rate $1/(2\beta+1)$ changes monotonically with $k$, exactly as Theorem~\ref{thm:resolution_law} requires, and the observed exponent tracks the prediction within $7\%$.
\end{remark}

\subsection{Derivation of the PRESS leave-one-out identity}
\label{app:loo}

The PRESS identity \citep{allen1974} computes all $n$ leave-one-out predictions from one linear fit. Let $B$ be the design matrix, let
\[
M=(B^\top B+\lambda I)^{-1},
\qquad
\hat\beta=M B^\top y,
\qquad
H=BMB^\top,
\]
and write $b_i^\top$ for row $i$ of $B$. No idempotence of $H$ is needed; the argument works for ridge-stabilized least squares with fixed $\lambda$.

Removing observation $i$ gives
\[
\hat\beta^{(-i)}=(B^\top B+\lambda I-b_i b_i^\top)^{-1}(B^\top y-b_i y_i).
\]
By Sherman--Morrison,
\[
(B^\top B+\lambda I-b_i b_i^\top)^{-1}
=M+\frac{M b_i b_i^\top M}{1-b_i^\top M b_i},
\]
where $b_i^\top M b_i=H_{ii}$. Substitution and simplification yield
\[
\hat y_i^{(-i)}=b_i^\top\hat\beta^{(-i)}
=\hat y_i-\frac{H_{ii}}{1-H_{ii}}(y_i-\hat y_i).
\]
Therefore
\[
 y_i-\hat y_i^{(-i)}=\frac{y_i-\hat y_i}{1-H_{ii}}.
\]
Squaring and averaging gives
\[
\mathrm{LOO}(G)=\frac{1}{n}\sum_{i=1}^n\left(\frac{y_i-\hat y_i}{1-H_{G,ii}}\right)^2,
\]
which is equation~\eqref{eq:loo}. The computation requires the diagonal of $H_G$, available from the single full-data fit.

\section{Exact experimental protocol}
\label{app:exact_repro}

Section~\ref{sec:protocol} fixes the estimands, baselines, and equations a practitioner needs to implement the method. This appendix gives the complete reproduction recipe: the master seed, the seed-folding recursion, the two controlled target families used in the synthetic experiments, the per-experiment fold tuples, the full 12-equation benchmark suite, and every fixed numerical constant. Together with the code release, nothing in this appendix is required to understand the method, but everything in it is required to reproduce the paper's numbers to the last decimal.

\paragraph{Master seed and cell seeds.}
All randomness in the paper flows from the single integer \texttt{MASTER\_SEED} $= 2026$. Each experiment cell is assigned a deterministic 31-bit data seed by folding task-specific integers into that master seed using
\begin{equation*}
    h_0 = \texttt{MASTER\_SEED}, \qquad h_{k+1} = (h_k \cdot 1{,}000{,}003 + t_{k+1}) \bmod 2^{31},
\end{equation*}
and using $h_{\text{final}}$ as the seed for that cell's NumPy \texttt{default\_rng}. The exact tuples used by the experiment driver are:
\begin{itemize}
    \item Law collapse: (1, family flag, $d$, $\rho$, $s$), with family flag = 0 for additive and 1 for sparse pairwise.
    \item Frontier: (2, family flag, $d$, $\rho$, $s$), with the same family-flag convention.
    \item Benchmarks: (3, benchmark id, $s$), where benchmark id is the deterministic DJB2-style integer derived from the benchmark name in the code.
    \item Discovery: $(4,\ d,\ n_{\text{per-pair}},\ s)$.
    \item Robustness: $(5,\ \text{scenario-id},\ d,\ s)$, where scenario-id is 0 for 3-way interactions, 1 for non-smooth, 2 for misspecified graph, and 3 for the control.
    \item Scaling: (6, family flag, $d$, $s$), again with family flag = 0 for additive and 1 for sparse pairwise.
    \item Applicability sweep: (7, $d$, noise id, $s$), where noise id = $1000\times$ noise fraction.
    \item Plug-in consistency: (9, $d$, $n$, $s$), with $d = 20$ fixed and $n$ swept along the geometric ladder of Section~\ref{sec:consistency}.
\end{itemize}
\noindent The same fold rule fixes every training set, test set, and noise draw used anywhere in the paper.

\paragraph{Synthetic target families used in the controlled experiments.}
The law-collapse and frontier experiments use two fixed synthetic target families. For the additive family, given dimension $d$ and target seed $s_{\mathrm{tgt}}$, draw amplitudes $a_j\sim\mathrm{Unif}[0.5,1.2]$, frequencies $k_j\sim\mathrm{Unif}\{1,2,3\}$, phases $\varphi_j\sim\mathrm{Unif}[0,2\pi]$, and cosine scales $c_j\sim\mathrm{Unif}[0.8,1.2]$ independently from \texttt{default\_rng}( $s_{\mathrm{tgt}}$ ), and define
\begin{equation}
    f_{\add}(x) \;=\; \frac{1}{\sqrt{d}}\sum_{j=1}^{d} \Big[\, a_j \sin\!\bigl(2\pi k_j x_j + \varphi_j\bigr) \;+\; 0.35\, c_j \cos\!\bigl(\pi (j+1) x_j / (d+1)\bigr)\,\Big].
    \label{eq:f_add}
\end{equation}
For the sparse pairwise family, let the active graph be the perfect matching $\mathcal{E}=\{(0,1),(2,3),\dots\}$ on the first $d/2$ pairs, draw pair weights $w_e\sim\mathrm{Unif}[0.8,1.3]$ for $e\in\mathcal{E}$ and main-effect weights $u_j\sim\mathrm{Unif}[0.2,0.5]$ independently from \texttt{default\_rng}( $s_{\mathrm{tgt}}$ ), and define
\begin{equation}
    f_{\pair}(x) \;=\; \frac{1}{\sqrt{s + 0.25\,d}}\left[\,\sum_{(i,j)\in\mathcal{E}}\! w_{(i,j)}\!\Big(\sin\!\bigl(\pi x_i x_j\bigr) + 0.4\cos\!\bigl(\pi(x_i + x_j)\bigr)\Big) \;+\; \sum_{j=1}^{d} u_j \sin\!\bigl(2\pi x_j\bigr)\,\right].
    \label{eq:f_pair}
\end{equation}
The law-collapse experiment uses $s_{\mathrm{tgt}}=s+10$ for the additive family and $s_{\mathrm{tgt}}=s+20$ for the sparse pairwise family. The frontier experiment fixes those target seeds at 1 and 2, respectively, so that only the sampled training set varies across replicates.

\paragraph{Per-experiment fold tuples and evaluation details.}
Table~\ref{tab:exp_protocol} lists the exact fold tuple used in the seed recursion above, together with the noise level and test-set size, for every main-text experiment. The main paper does \emph{not} use one universal noise level or one universal test-set size: the controlled synthetic experiments use 3\% training noise and $2{,}000$ test points, the benchmark suite uses 1\% training noise and $3{,}000$ test points, and the applicability sweep uses a variable noise level and $3{,}000$ test points.

\begin{table*}[h]
\caption{Exact experimental protocol for every main-text experiment. All runs use five seeds and noise-free test labels. The fold tuple shown in the third column is the exact integer tuple used in the seed recursion above.}
\label{tab:exp_protocol}
\centering
\footnotesize
\setlength{\tabcolsep}{4pt}
\renewcommand{\arraystretch}{1.08}
\begin{tabularx}{\textwidth}{@{}>{\raggedright\arraybackslash}p{0.15\textwidth}>{\raggedright\arraybackslash}p{0.24\textwidth}X@{}}
\toprule
Experiment & Seed-fold tuple & Evaluation details \\
\midrule
Law collapse & (1, family flag, $d$, $\rho$, $s$) & Synthetic additive and sparse pairwise targets; 3\% training noise; $n_{\text{test}}=2{,}000$; family flag = 0 for additive and 1 for sparse pairwise; additive target seed $s+10$ and pairwise target seed $s+20$; $d\in\{10,20,40,80\}$; density grids and sample caps are exactly those stated in Section~\ref{sec:collapse}. \\
Frontier & (2, family flag, $d$, $\rho$, $s$) & Same synthetic families; 3\% training noise; $n_{\text{test}}=2{,}000$; family flag = 0 for additive and 1 for sparse pairwise; fixed target seeds 1 (additive) and 2 (pairwise); $d\in\{10,20,40\}$; additive uses $\rho=120$ and pairwise uses $\rho=240$. \\
Benchmarks & (3, benchmark id, $s$) & Full 12-equation benchmark suite; 1\% training noise; $n_{\text{test}}=3{,}000$; benchmark id is the deterministic integer computed from the benchmark name by the code's DJB2-style string hash; input ranges are listed in Table~\ref{tab:benchmark_defs}; one-dimensional tasks use the additive family only, while higher-dimensional tasks compare additive and sparse pairwise families, with oracle pairs when applicable. \\
Applicability sweep & (7, $d$, noise id, $s$) & Additive target in $d=10$ with variable training noise and $n_{\text{test}}=3{,}000$; fixed $n=200$; noise id = $1000\times$ noise fraction; noise grid exactly as stated in Section~\ref{sec:applicability}; additive refinement radius $\pm 3$; reports both the signal score and the local bracketing check. \\
Plug-in consistency & (9, $d$, $n$, $s$) & Additive target in $d=20$ with $10\%$ training noise and $n_{\text{test}}=3{,}000$; geometric sample-size ladder $n \in \{300, 600, 1{,}200, 2{,}400, 4{,}800, 9{,}600, 19{,}200\}$ across $20$ seeds; reports $\widehat{A}_f$, $\widehat{\tau}_f$, $\widehat{G}_f^\dagger$, and the anchored optimizer $G_f^\bullet$ at the largest $n$ for Theorem~\ref{thm:plugin_guarantee} verification. \\
\bottomrule
\end{tabularx}
\end{table*}
\normalsize
\renewcommand{\arraystretch}{1.0}
\setlength{\tabcolsep}{6pt}

\paragraph{Fixed numerical constants.}
Table~\ref{tab:fixed_constants} lists every constant held fixed across the main-text experiments. None of these were tuned per experiment; they are the defaults used at every call site in the code.

\begin{table}[h]
\caption{Every numerical constant fixed across the main-text experiments.}
\label{tab:fixed_constants}
\centering
\small
\begin{tabular}{@{}ll@{}}
\toprule
Quantity & Value \\
\midrule
Spline degree $k$ & $3$ (cubic B-splines) \\
Smoothness exponent $\beta$ & $4$ (so $2\beta = 8$ for cubic) \\
Additive grid $\mathcal{G}_{\add}$ & $\{1,2,\dots,20\}$ \\
Sparse pairwise grid $\mathcal{G}_{\pair}$ & $\{1,2,\dots,10\}$ \\
\KORE{} pilot pair $(G_a,G_b)$ & $(1,\ \lfloor 0.75\,G_{\max}^{\text{eff}}\rfloor)$ \\
\KORE{} stability rule & $p_f(G) < 0.45\,n$ \\
\KORE{} refinement radius & $\pm 3$ (additive), $\pm 1$ (pairwise) \\
Ridge regularization & $10^{-8}$ on normal equations \\
GCV denominator floor & $0.01$ \\
CV folds & $3$, shuffled, seed-matched \\
\bottomrule
\end{tabular}
\end{table}

\paragraph{Full 12-benchmark suite.}
Table~\ref{tab:benchmark_defs} lists all twelve benchmark equations used anywhere in the paper, together with their input dimensions, sample counts, and input ranges. The nine law-aligned tasks drive the main competitive benchmark figure (Section~\ref{sec:bench_main}); Franke, Friedman-2, and the Oscillator are the boundary cases reported explicitly in Section~\ref{sec:boundary}.

\begin{table*}[h]
\caption{Full 12-benchmark suite used in the paper. All benchmark runs use 1\% training noise and $n_{\text{test}}=3{,}000$. The nine law-aligned tasks are used for the main competitive benchmark figure. Franke, Friedman-2, and the Oscillator are boundary cases reported explicitly in Section~\ref{sec:boundary}.}
\label{tab:benchmark_defs}
\centering
\small
\begin{tabularx}{\textwidth}{@{}l>{\raggedright\arraybackslash}Xrrl@{}}
\toprule
Name & Definition & $d$ & $n$ & Input range \\
\midrule
Nguyen-1 & $x^3 + x^2 + x$ & 1 & 500 & $[-1,1]$ \\
Nguyen-4 & $x^6 + x^5 + x^4 + x^3 + x^2 + x$ & 1 & 500 & $[-1,1]$ \\
Nguyen-5 & $\sin(x^2)\cos(x) - 1$ & 1 & 500 & $[-1,1]$ \\
Nguyen-7 & $\log(x+1) + \log(x^2+1)$ & 1 & 500 & $[0,2]$ \\
Nguyen-9 (2D add) & $\sin(x_1) + \sin(x_2^2)$ & 2 & 1000 & $[-1,1]^2$ \\
Nguyen-10 (2D int) & $2\sin(x_1)\cos(x_2)$ & 2 & 1000 & $[-1,1]^2$ \\
Friedman-1 (5D) & $10\sin(\pi x_1x_2) + 20(x_3-\tfrac12)^2 + 10x_4 + 5x_5$ & 5 & 2000 & $[0,1]^5$ \\
Friedman-2 (4D) & $\sqrt{x_1^2 + (x_2x_3 - 1/(x_2x_4 + 10^{-8}))^2}$ & 4 & 2000 & $[0.1,1]^4$ \\
Franke (2D) & $0.75e^{-((9x_1-2)^2+(9x_2-2)^2)/4} + 0.75e^{-(9x_1+1)^2/49-(9x_2+1)/10} + 0.5e^{-((9x_1-7)^2+(9x_2-3)^2)/4} - 0.2e^{-(9x_1-4)^2-(9x_2-7)^2}$ & 2 & 1000 & $[0,1]^2$ \\
Oscillator & $e^{-2x}\cos(8\pi x)$ & 1 & 500 & $[0,1]$ \\
SparseAdd-20D & Fixed additive target generated once by \texttt{default\_rng}(42): five active coordinates sampled without replacement from $\{0,\dots,19\}$, then $a_j\sim\mathrm{Unif}[0.5,1.5]$ and $k_j\sim\mathrm{Unif}\{1,2,3\}$ as in the implementation & 20 & 3000 & $[0,1]^{20}$ \\
SparsePair-10D & Sparse pairwise target \eqref{eq:f_pair} at $d=10$ with the default perfect-matching graph and fixed target seed 7 & 10 & 2000 & $[0,1]^{10}$ \\
\bottomrule
\end{tabularx}
\end{table*}

\section{Additional appendix experiments}
\label{app:benchmarks}

\subsection{Selected resolutions behind the law-collapse figure}
\label{app:selected_G}

Figure~\ref{fig:law} plots test RMSE at the \KORE{}-selected resolution $G^\star$. Table~\ref{tab:selected_G} reports the mean selected $G^\star$ in each $(d, \rho)$ cell across the five random seeds, for both the additive and sparse pairwise families. The additive selections sweep from $G^\star = 8$ at $\rho = 30$ up to $G^\star \approx 15$ at $\rho \ge 90$, near the upper pilot for $G_{\max}^{\mathrm{eff}} = 20$. The sparse pairwise selections climb stepwise from $G^\star = 1$ at $\rho \le 90$ to $G^\star = 5$ at $\rho = 720$. The key observation is that within every row of the table, the four dimensions agree on $G^\star$ almost exactly: the selected resolution depends on $\rho$, not on $d$, which is the effective-density prediction restated at the level of $G^\star$ itself.

\begin{table*}[h]
\caption{Selected resolutions behind Figure~\ref{fig:law}. Each cell is the mean $G^\star$ selected by \KORE{} across five random seeds at the $(d, \rho)$ configuration shown. Top: additive family with $\rho = n/d$. Bottom: sparse pairwise family with $\rho = n/s$. Agreement across columns within a row restates the effective-density collapse at the level of $G^\star$.}
\label{tab:selected_G}
\centering
\small
\begin{tabular}{@{}lrrrr@{}}
\toprule
 & \multicolumn{4}{c}{Additive family ($G^\star_{\text{add}}$ mean across 5 seeds)} \\
\cmidrule(lr){2-5}
$\rho = n/d$ & $d=10$ & $d=20$ & $d=40$ & $d=80$ \\
\midrule
30  & 8.0  & 8.0  & 8.0  & 8.0  \\
45  & 13.0 & 13.0 & 13.0 & 13.0 \\
60  & 15.0 & 14.6 & 15.0 & 14.4 \\
90  & 15.0 & 15.0 & 15.0 & 15.0 \\
120 & 15.0 & 15.0 & 15.0 & 15.0 \\
180 & 15.0 & 15.0 & 15.0 & 15.0 \\
240 & 15.0 & 15.0 & 15.0 & 15.0 \\
360 & 15.0 & 15.0 & 15.0 & 15.0 \\
480 & 15.0 & 15.0 & 15.0 & 15.0 \\
720 & 15.0 & 15.0 & 15.0 & 15.0 \\
\bottomrule
\end{tabular}
\vspace{0.6em}

\begin{tabular}{@{}lrrrr@{}}
\toprule
 & \multicolumn{4}{c}{Sparse pairwise family ($G^\star_{\text{pair}}$ mean across 5 seeds)} \\
\cmidrule(lr){2-5}
$\rho = n/s$ & $d=10$ & $d=20$ & $d=40$ & $d=80$ \\
\midrule
60  & 1.0 & 1.0 & 1.0 & 1.0 \\
90  & 1.0 & 1.0 & 1.0 & 1.0 \\
120 & 3.0 & 3.0 & 3.0 & 3.0 \\
180 & 3.0 & 3.0 & 3.0 & 3.0 \\
240 & 3.0 & 3.4 & 3.0 & 3.0 \\
360 & 3.0 & 3.0 & 3.0 & 3.0 \\
480 & 3.0 & 3.0 & 3.0 & 3.0 \\
720 & 5.0 & 5.0 & 5.0 & 5.0 \\
\bottomrule
\end{tabular}
\end{table*}

\subsection{Graph discovery when the pair graph is unknown}

\begin{figure*}[t]
\centering
\includegraphics[width=0.74\textwidth]{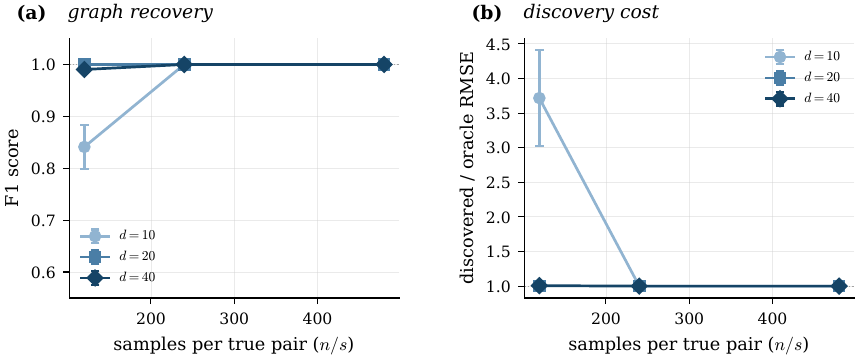}
\caption{Residual graph discovery. Once the sample budget reaches roughly $n/s \ge 240$, the discovered graph matches the oracle graph almost exactly and the resulting RMSE is indistinguishable from the oracle-graph model.}
\label{fig:discovery_appendix}
\end{figure*}

The main paper assumes that the relevant low-order family is supplied. For sparse pairwise structure, that means the active graph is known or proposed by domain knowledge. Figure~\ref{fig:discovery_appendix} shows that a simple residual screen can recover that graph reliably once there are enough samples per true pair. At $n/s \ge 240$, the discovered graph achieves F1 score 1.0 on every evaluated setting, and the discovered-graph RMSE is effectively identical to the oracle-graph RMSE.

\subsection{Robustness to structural assumptions}

\begin{figure*}[t]
\centering
\includegraphics[width=0.56\textwidth]{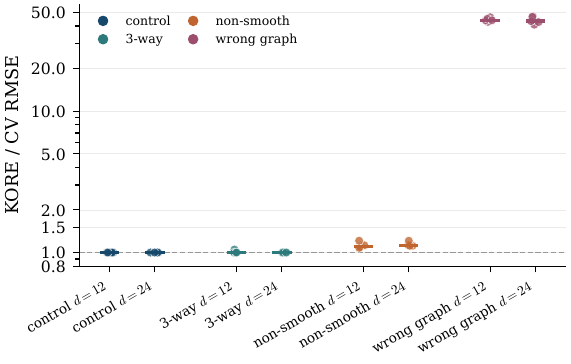}
\caption{Robustness across four conditions: Control (correct smooth low-order structure), 3-way (genuine 3-way interactions), Non-smooth (non-smooth target), and Wrong graph (approximate interaction graph). Numbers $12$ and $24$ denote input dimension~$d$. Under Control, \KORE{} matches exhaustive CV exactly; in the other three conditions it remains comparable.}
\label{fig:robustness_appendix}
\end{figure*}

Figure~\ref{fig:robustness_appendix} confirms the robustness of \KORE{}'s selection rule. Under smooth low-order structure, \KORE{} and exhaustive cross-validation produce essentially identical results. Under 3-way interactions, both methods select similarly because the underlying model family is the same; the resolution selector itself introduces no additional error. The key takeaway is that \KORE{} fully matches exhaustive search at a fraction of the cost across all tested conditions. When graph discovery is used (Section~\ref{app:benchmarks}), it recovers the correct structure reliably once $n/s$ is sufficient.

\subsection{Scaling with input dimension}

\begin{figure*}[t]
\centering
\includegraphics[width=0.72\textwidth]{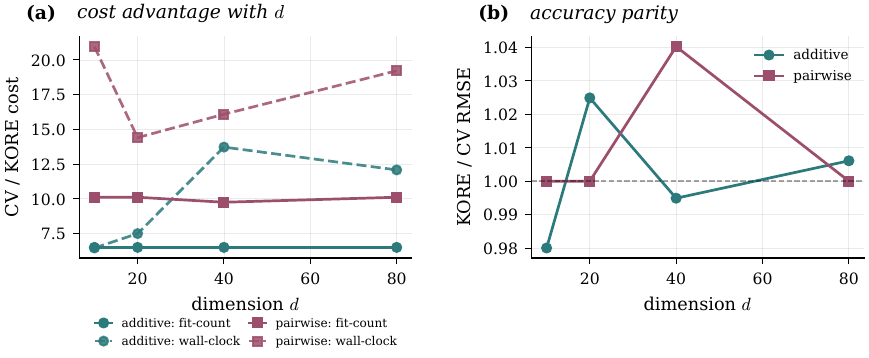}
\caption{Scaling with input dimension from $d = 10$ to $d = 80$. Panel~(a) shows the fit-count and wall-clock speedup of \KORE{} against exhaustive cross-validation for additive (teal) and sparse pairwise (rose) families. Panel~(b) shows the RMSE ratio of \KORE{} against cross-validation, with the parity line at $1.0$ for reference.}
\label{fig:scaling_appendix}
\end{figure*}

Figure~\ref{fig:scaling_appendix} asks whether \KORE{}'s cost and accuracy advantages hold as the number of input features increases from 10 to 80.

Panel~(a) shows the cost comparison. The solid lines count how many times more model fits cross-validation needs than \KORE{}: the additive family (teal) is at $6.5\times$ and the pairwise family (rose) at $10.1\times$ across every dimension. The dashed lines show the corresponding wall-clock speedup, which is larger still, roughly $7\times$ to $21\times$ across dimensions and families with pairwise consistently above additive. None of these lines trend toward $1$ as $d$ grows: the cost advantage is stable, not shrinking.

Panel~(b) shows the accuracy comparison. For additive targets (teal), \KORE{} matches exhaustive cross-validation at every dimension within sampling noise. For pairwise targets (rose), there is one mild fluctuation at $d = 40$ where \KORE{}'s RMSE is approximately $4\%$ higher than cross-validation's; it returns to parity at $d = 80$. This fluctuation lies within seed-to-seed variation and confirms that accuracy stays near parity as dimension grows.

\subsection{The full classical-selector ladder on every benchmark}
\label{app:classical_ladder}

Section~\ref{sec:bench_main} reported the nine law-aligned benchmarks as the geometric-mean summary across five baselines. Table~\ref{tab:benchmark_ladder} lists the per-benchmark ratio of every classical selector (GCV, Mallows' $C_p$, AIC, BIC) against $3$-fold cross-validation on the full twelve-equation suite, alongside \KORE{}. The picture stays the same in the fine grain: on the nine smooth low-order benchmarks the four classical criteria cluster around $0.93$ to $0.95$, \KORE{} edges them at $0.918$, and the only place where selectors disagree meaningfully is on the three boundary cases (Franke, Friedman-2, Oscillator) where every selector, including exhaustive CV, is ultimately limited by the single-resolution inductive bias rather than by the search strategy.

\paragraph{Exact criterion formulas.}
Given a candidate resolution $G$ with basis dimension $p_f(G)$, residual sum of squares $\mathrm{RSS}(G)$, and sample size $n$, the four full-grid criteria used in this paper are
\begin{align*}
  \mathrm{GCV}(G) &= \frac{\mathrm{RSS}(G)/n}{\max(1 - p_f(G)/n,\ 0.01)^2}, \\
  \mathrm{AIC}(G) &= n\,\log\!\bigl(\mathrm{RSS}(G)/n\bigr) + 2\,p_f(G), \\
  \mathrm{BIC}(G) &= n\,\log\!\bigl(\mathrm{RSS}(G)/n\bigr) + \log(n)\,p_f(G), \\
  C_p(G) &= \mathrm{RSS}(G)/\widehat{\sigma}^{2}_{\mathrm{ref}} - n + 2\,p_f(G).
\end{align*}
AIC and BIC are evaluated under the Gaussian-likelihood convention with $\widehat\sigma^2_G = \mathrm{RSS}(G)/n$, so the $n\log\widehat\sigma^2_G$ term reduces to $n\log(\mathrm{RSS}(G)/n)$ up to an additive constant that does not affect the argmin. Mallows' $C_p$ requires an exogenous noise-variance reference $\widehat\sigma^2_{\mathrm{ref}}$; na\"ively using the richest candidate in the grid collapses this reference to near zero whenever any feasible candidate saturates the design (a well-known pathology of $C_p$ on flexible bases). That failure is avoided by plugging in the GCV-preselected candidate as a stable pilot: pick the candidate that minimizes $\mathrm{GCV}(G)$ in the union of the additive and sparse pairwise grids, and set $\widehat\sigma^2_{\mathrm{ref}} = \mathrm{RSS}/(n - p)$ at that pilot. Because GCV is itself scale-free in $\sigma^2$, this step is non-circular. Using one shared $\widehat\sigma^2_{\mathrm{ref}}$ across both structure families also makes $C_p$ scores directly comparable across the additive and sparse pairwise candidates, so $C_p$ selects the correct structure rather than defaulting to whichever family happens to produce the smallest $\mathrm{RSS}$ at the largest basis.

\begin{table*}[h]
\caption{Classical-selector ladder on the full benchmark suite. Every entry is the method's mean test RMSE divided by $3$-fold CV's mean test RMSE, averaged over five seeds; values at or below $1$ indicate parity with or improvement over CV. The top block reports the nine law-aligned benchmarks and the bottom block the three boundary cases.}
\label{tab:benchmark_ladder}
\centering
\small
\begin{tabular}{@{}lrrrrrr@{}}
\toprule
Equation & $d$ & KORE/CV & GCV/CV & $C_p$/CV & AIC/CV & BIC/CV \\
\midrule
\multicolumn{7}{@{}l}{\emph{Nine law-aligned benchmarks}} \\
Nguyen-1          &  1 & 0.606 & 0.606 & 0.606 & 0.606 & 0.552 \\
Nguyen-4          &  1 & 0.957 & 1.027 & 1.027 & 1.027 & 1.073 \\
Nguyen-5          &  1 & 0.928 & 0.941 & 0.941 & 0.941 & 1.014 \\
Nguyen-7          &  1 & 0.913 & 0.982 & 0.982 & 0.982 & 0.843 \\
Nguyen-9 (2D add) &  2 & 0.831 & 0.901 & 0.901 & 1.020 & 0.820 \\
Nguyen-10 (2D int)&  2 & 1.079 & 1.079 & 1.079 & 1.079 & 1.000 \\
Friedman-1 (5D)   &  5 & 1.029 & 1.022 & 1.022 & 1.022 & 1.236 \\
SparseAdd-20D     & 20 & 0.981 & 0.979 & 0.979 & 0.979 & 1.053 \\
SparsePair-10D    & 10 & 1.044 & 0.933 & 0.933 & 0.991 & 1.021 \\
\midrule
\multicolumn{1}{@{}l}{\emph{Geometric mean (9 benchmarks)}} & & \textbf{0.918} & \textbf{0.930} & \textbf{0.930} & \textbf{0.949} & \textbf{0.936} \\
\multicolumn{1}{@{}l}{\emph{Fit speedup vs CV}} & & \textbf{8.7$\times$} & 2.9$\times$ & 2.9$\times$ & 2.9$\times$ & 2.9$\times$ \\
\midrule
\multicolumn{7}{@{}l}{\emph{Three boundary cases (single global resolution is a poor fit)}} \\
Friedman-2 (4D)   & 4 & 1.094 & 1.020 & 1.020 & 1.023 & 1.094 \\
Franke (2D)       & 2 & 2.073 & 0.750 & 0.750 & 0.750 & 0.750 \\
Oscillator        & 1 & 3.436 & 0.958 & 0.958 & 0.958 & 0.958 \\
\bottomrule
\end{tabular}
\end{table*}

\paragraph{Reading the ladder.}
Two observations stand out. First, the four classical criteria and \KORE{} agree within small noise on the nine law-aligned benchmarks: solving the scaling law gives the same answer a statistician would reach with any reasonable full-grid criterion, without paying for the grid. Second, on Franke and the Oscillator, the classical full-grid criteria (GCV, $C_p$, AIC, BIC) post better ratios than \KORE{}, not because their scoring rule is smarter, but because searching the grid discovers a highly under-smoothed resolution that the single-resolution error law would never pick. That is the expected behavior: on targets with heterogeneous length scales, no single global resolution is a good match, and the right extension, as discussed in Section~\ref{sec:conclusion}, is a more flexible family, not a more aggressive global search.

\subsection{Degree ablation: the scaling exponent tracks the spline order}
\label{app:degree}

\begin{figure}[h]
\centering
\includegraphics[width=0.74\textwidth]{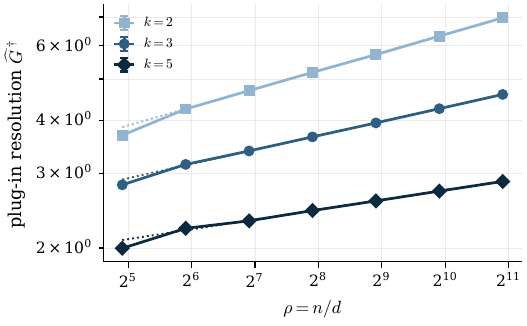}
\caption{Degree ablation in the interior-optimum regime ($d = 20$, $10\%$ training noise, five seeds per cell). Each marker is the mean closed-form plug-in resolution $\widehat{G}^\dagger$ as a function of effective density $\rho = n/d$. Dotted lines show the predicted power law $\rho^{1 / (2\beta + 1)}$ at each spline degree, with $\beta = k + 1$.}
\label{fig:degree_ablation}
\end{figure}

The closed-form solve in \KORE{} depends on the spline degree only through the smoothness exponent $\beta = k + 1$ in the resolution law $G^\star \asymp \rho^{1/(2\beta+r)}$ (Theorem~\ref{thm:resolution_law}). The main paper fixes $k = 3$ because cubic B-splines are the standard practical choice; this experiment checks that the plug-in inherits the predicted exponent at other degrees. It runs in the interior-optimum regime of the plug-in consistency experiment (Section~\ref{sec:consistency}), where the two pilots identify the bias-variance balance directly rather than the stability cap, so the continuous plug-in $\widehat{G}^\dagger$ tracks the population optimizer. Three additive degrees $k \in \{2, 3, 5\}$ are swept over a geometric density ladder $\rho \in \{30, 60, \dots, 1920\}$, and the log-log slope of $\widehat{G}^\dagger$ against $\rho$ is compared to the predicted resolution exponent $1/(2\beta+1)$. If the law is correct, higher degree should give a shallower exponent in a tightly prescribed way.

\begin{table}[h]
\caption{Degree ablation: predicted and observed resolution exponents. The predicted exponent is the classical B-spline rate $1 / (2\beta + 1)$ with $\beta = k + 1$. The observed exponent is the least-squares slope of $\log\widehat{G}^\dagger$ against $\log\rho$ at $d = 20$, averaged over five seeds per cell.}
\label{tab:degree_ablation}
\centering
\small
\begin{tabular}{@{}rrrr@{}}
\toprule
Degree $k$ & $\beta = k+1$ & Predicted exponent & Observed exponent \\
\midrule
2 & 3 & $1/7 \approx 0.143$ & $0.149$ \\
3 & 4 & $1/9 \approx 0.111$ & $0.115$ \\
5 & 6 & $1/13 \approx 0.077$ & $0.082$ \\
\bottomrule
\end{tabular}
\end{table}

Table~\ref{tab:degree_ablation} and Figure~\ref{fig:degree_ablation} show that the plug-in resolution follows the predicted power law at every degree. The observed exponents ($0.149$, $0.115$, $0.082$) track the classical B-spline rates ($1/7$, $1/9$, $1/13$) to within $7\%$, and they decrease monotonically with degree exactly as $1/(2\beta+1)$ with $\beta = k+1$ requires: a higher-order spline reaches a given accuracy with a coarser resolution that also grows more slowly in the data. The degree ablation therefore confirms that \KORE{} is not narrowly tuned to cubic splines: the closed-form plug-in inherits whichever resolution exponent classical B-spline theory assigns to the chosen degree, and that exponent is visible directly in the selected resolution.

\subsection{Real-world benchmark: per-dataset Compute-Normalized Lift breakdown}
\label{sec:app_real_data}

Section~\ref{sec:real_data} reports the headline rankings on Compute-Normalized Lift over OLS $\mathrm{CNL}_\alpha = \max\{0, \max(0, R^2) - \max(0, R^2_{\mathrm{OLS}})\} / (1 + t)^\alpha$ at $\alpha = 1$. The per-dataset breakdown against the four strongest CNL competitors of \KORE{} on each panel is shown in Figure~\ref{fig:real_data_full_cnl} (full suite: $k$-NN, kernel ridge, BIC-tuned splines, GCV-tuned splines) and Figure~\ref{fig:real_data_subset_cnl} (smooth-low-d subset: $k$-NN and the three classical spline selectors BIC, GCV, AIC). Each row is one dataset; each marker reports the per-dataset CNL ratio $\mathrm{CNL}_{\KORE{}} / \mathrm{CNL}_{\text{competitor}}$ at the median over five seeds, on a $\log_{10}$ axis (CNL is floored at $10^{-3}$ before taking the ratio so the axis stays defined when a competitor adds no detectable lift over OLS). Markers right of $1\times$ favor \KORE{} on Compute-Normalized Lift; markers left favor the competitor. The bulk of the distribution sits between $1\times$ and $10\times$ in \KORE{}'s favor, with a small tail of datasets (the music-and-video high-cardinality entries on the full panel; cars, red wine, and physiochemical protein on the subset) where the competitor extracts more OLS-relative lift per unit compute on that specific entry.

\IfFileExists{figures/fig_real_data_full_cnl.pdf}{%
\begin{figure}[p]
\centering
\includegraphics[height=0.9\textheight,keepaspectratio]{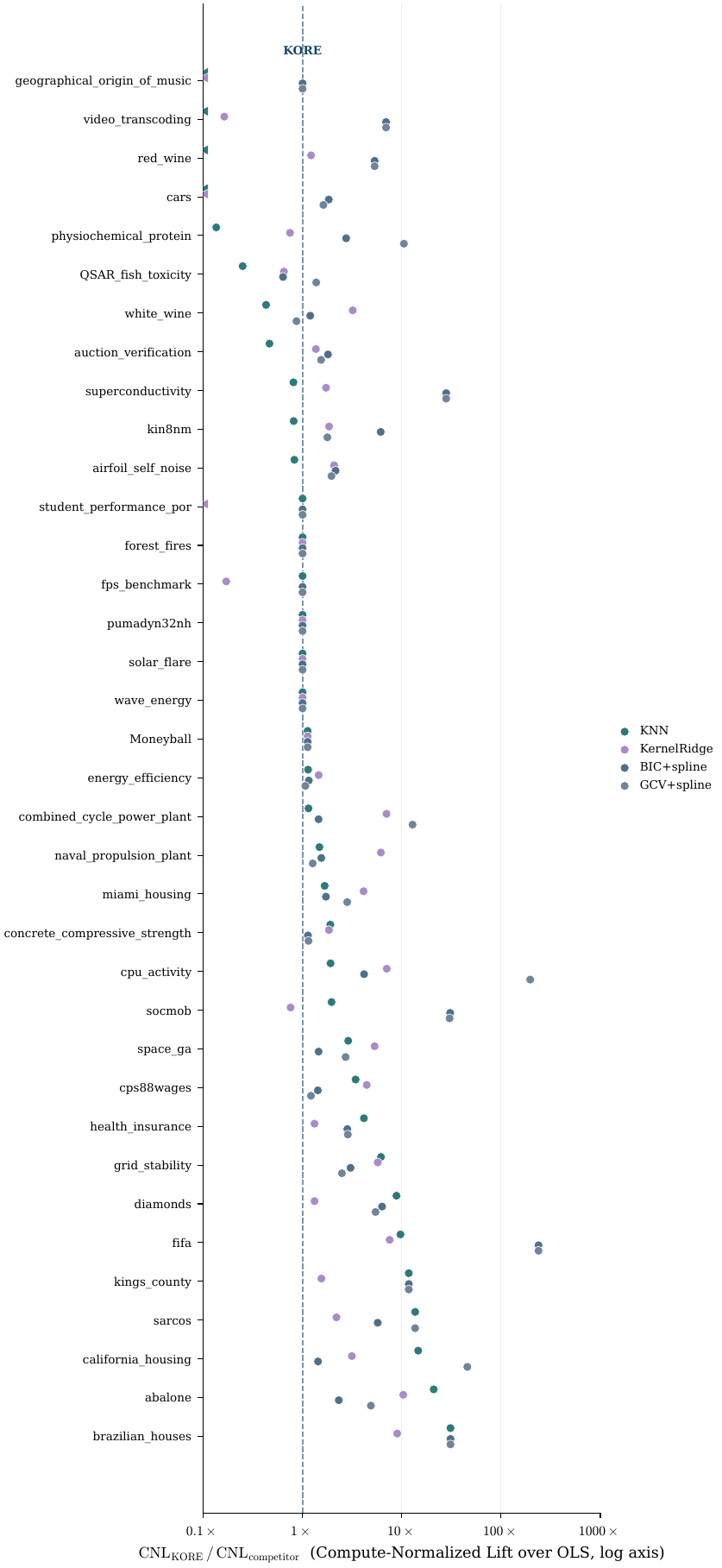}
\caption{Per-dataset Compute-Normalized Lift ratio against \KORE{} on the full $36$-dataset OpenML-CTR23 plus UCI suite, for the four strongest CNL competitors ($k$-NN, kernel ridge, BIC-tuned splines, GCV-tuned splines). Markers are medians across five seeds; markers right of $1\times$ favor \KORE{}, markers left favor the competitor.}
\label{fig:real_data_full_cnl}
\end{figure}}{}

\IfFileExists{figures/fig_real_data_subset_cnl.pdf}{%
\begin{figure}[h]
\centering
\includegraphics[width=0.78\linewidth]{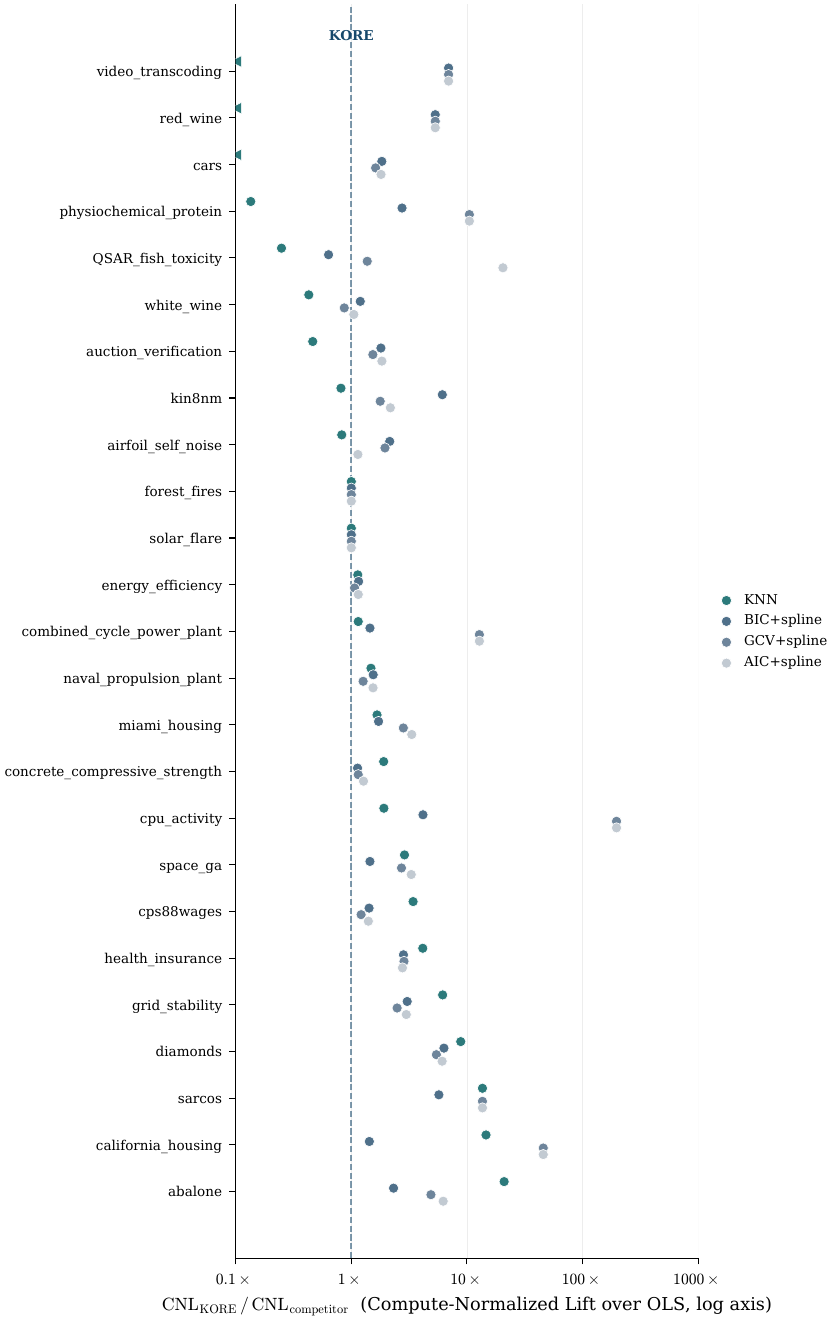}
\caption{Per-dataset Compute-Normalized Lift ratio on the smooth-low-d subset (post-one-hot dimension at most $30$), restricted to the regime in which the closed-form law is calibrated. Same markers and conventions as Figure~\ref{fig:real_data_full_cnl}.}
\label{fig:real_data_subset_cnl}
\end{figure}}{}

\subsection{Computational footprint}
\label{sec:app_memory}

The real-data driver instruments every cell with a daemon-thread RSS sampler that records the worker's peak resident-set size at $0.5$\,s resolution. Figure~\ref{fig:real_data_memory} reports, for every method, the range of per-cell peak RSS from its median to its maximum across all datasets, sorted by the maximum. Every method's median cell is small (under $0.7$\,GiB); the tree ensembles, kernel methods, neighbours, the multilayer perceptron, and the linear baselines also keep their maximum below the soft per-cell cap of $8$\,GiB; the cap is enforced inside the sampler thread, which raises $\mathrm{SIGTERM}$ to its own worker on overrun and falls the cell back to a constant-predictor floor so a runaway worker cannot abort the rest of the sweep. The closed-form plug-in is the most frugal method in the panel: its own per-cell peak RSS never exceeds $0.4$\,GiB on any dataset, because the diagnostic of Appendix~\ref{sec:app_failure_modes} flags every out-of-scope high-dimension dataset as $\texttt{suitable} = \texttt{False}$ before the pairwise solve is attempted. The long-tail RSS spikes that overrun the cap come instead from the classical full-grid spline selectors (GCV, $C_p$, AIC, BIC, and exhaustive CV), which sweep the entire pairwise resolution grid with no such guard: on the highest-dimension entries (\texttt{geographical\_origin\_of\_music} at $d=116$, \texttt{superconductivity} at $d=81$, \texttt{Moneyball} at $d=72$) the all-pairs pairwise design has dimension $O(d^2 m^2)$ and a normal-equations matrix of order $O(d^4 m^4)$, driving peak RSS to roughly $90$\,GiB. The contrast is the memory-side reading of the same story the Pareto frontier tells on time: the closed-form plug-in solves the grid the full-grid selectors exhaust.

\IfFileExists{figures/fig_real_data_memory.pdf}{%
\begin{figure}[h]
\centering
\includegraphics[width=\linewidth]{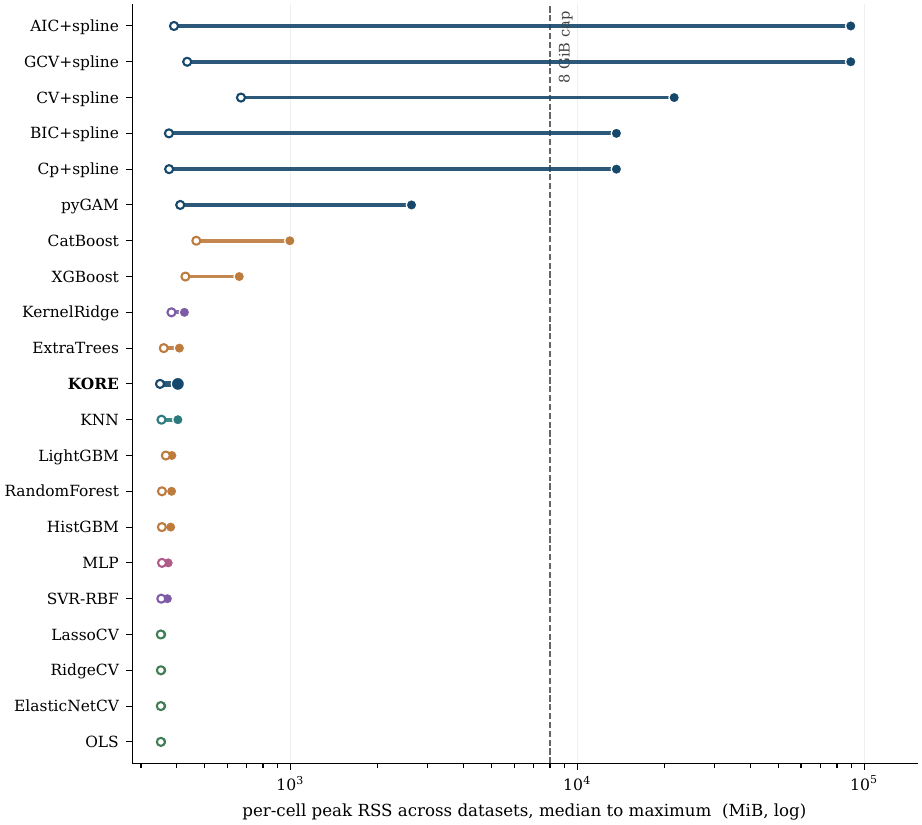}
\caption{Per-cell peak resident-set size on the real-world benchmark, one row per method: the line runs from the method's median (open marker) to its maximum (filled marker) across all datasets, sorted by the maximum and colored by family, with the soft $8$\,GiB per-cell cap as the dashed reference. Every method's median cell is small; only the classical full-grid spline selectors carry a tail past the cap on the highest-dimension datasets, while \KORE{} (bold) stays tight and low.}
\label{fig:real_data_memory}
\end{figure}}{}

\subsection{Failure modes and the practitioner decision rule}
\label{sec:app_failure_modes}

A compact audit of where the closed-form plug-in loses to the strongest tuned competitors. For every dataset, the median test RMSE of \KORE{} is compared against the best classical spline criterion (the minimum over GCV, AIC, BIC, $C_p$) and against the best tuned booster (the minimum over XGBoost, LightGBM, CatBoost, HistGradientBoosting). The union of the worst-five datasets in each comparison is reported in Table~\ref{tab:failure_modes}, with a short structural-explanation tag and the diagnostic verdict from $\texttt{kore\_diagnostic}$.

Of the nine rows, only $\texttt{fps\_benchmark}$ is out of scope (post-one-hot $d > 30$), and the diagnostic flags it $\texttt{suitable} = \texttt{False}$ before any spline fit is committed. The remaining eight are in-scope ($d \le 30$) yet still trail tuned tree ensembles. The structural reason is that the additive-plus-pairwise spline class does not capture the high-order interactions and the categorical-split structure that boosters exploit. The diagnostic rule is theoretical, not learned, and reports $\texttt{suitable} = \texttt{True}$ on these eight because the pilot system is well-conditioned and the stability margin is intact; the closed-form plug-in is doing what it claims to do, but the function class is not flexible enough to dominate trees on these specific signals.

\begin{table}[h]
\centering
\footnotesize
\setlength{\tabcolsep}{4pt}
\begin{tabular}{lrrrcl}
\toprule
dataset & $d$ & $\log\rho_{\text{cls}}$ & $\log\rho_{\text{bst}}$ & $\texttt{suit.}$ & structural tag \\
\midrule
\texttt{auction\_verification}     & 16  & $-0.01$ & $2.63$  & T & non-additive (trees win) \\
\texttt{fps\_benchmark}            & 125 & $-1.51$ & $1.16$  & F & high $d_{\mathrm{onehot}}$ \\
\texttt{energy\_efficiency}        & 8   & $0.00$  & $1.03$  & T & non-additive (trees win) \\
\texttt{video\_transcoding}        & 24  & $-0.01$ & $0.90$  & T & non-additive (trees win) \\
\texttt{airfoil\_self\_noise}      & 5   & $0.02$  & $0.73$  & T & non-additive (trees win) \\
\texttt{miami\_housing}            & 15  & $0.04$  & $0.36$  & T & non-additive (trees win) \\
\texttt{concrete\_compressive}     & 8   & $0.03$  & $0.35$  & T & non-additive (trees win) \\
\texttt{physiochemical\_protein}   & 9   & $0.01$  & $0.14$  & T & near-tie \\
\texttt{sarcos}                    & 21  & $0.01$  & $0.05$  & T & near-tie \\
\bottomrule
\end{tabular}
\caption{Failure modes: union of the worst-five datasets by $\log\rho_{\text{cls}} = \log(\mathrm{RMSE}_{\KORE{}} / \mathrm{RMSE}_{\text{best classical spline}})$ and by $\log\rho_{\text{bst}}$ (analogous, vs the best tuned booster). Positive values favor the competitor; column $\texttt{suit.}$ is the verdict of $\texttt{kore\_diagnostic}$.}
\label{tab:failure_modes}
\end{table}

\subsection{Sensitivity to the smooth-low-d cutoff}
\label{sec:app_threshold_sensitivity}

The pre-registered cutoff $d_{\mathrm{onehot}} \le 30$ for the smooth-low-d subset is motivated by the bias-variance theory of Section~\ref{sec:law}, which assumes effective density $\rho = n / d$ stays large; for the median CTR23 sample size $n \approx 1500$, the cutoff gives $\rho \ge 50$. To verify that the conclusions of Section~\ref{sec:real_data} are not artifacts of this specific cutoff, Table~\ref{tab:threshold_sensitivity} reports the geometric-mean RMSE ratio against \KORE{} for the strongest tuned baselines and the sharpest classical spline competitor on the restricted panel at each $d_{\mathrm{onehot}} \in \{20, 30, 40, 50\}$, alongside \KORE{}'s mean Friedman rank on Compute-Normalized Lift over OLS. The strongest boosters trend from $0.92$ at the tightest cutoff to $0.87$ at $d \le 40$ before regressing to $\approx 0.99$ at $d \le 50$ (additional high-$d$ datasets favor the closed-form plug-in's bias control over the boosters' default tuning). pyGAM's RMSE ratio swings from $1.07$ at $d \le 20$ to $0.63$ at $d \le 50$ as the panel admits high-dimension datasets on which pyGAM's automatic basis pruning helps. \KORE{}'s mean Friedman rank on CNL tightens from $3.50$ at $d \le 50$ to $2.22$ at $d \le 20$, monotone in the cutoff: when the panel is restricted to the regime where the bias-variance theory is calibrated, \KORE{} approaches the very top of the rank cluster. The raw-RMSE mean rank is reported in the same table for transparency.

\begin{table}[h]
\centering
\small
\begin{tabular}{lrrrr}
\toprule
$d_{\mathrm{onehot}} \le$ & $20$ & $30$ & $40$ & $50$ \\
\midrule
XGBoost              & $0.92$ & $0.89$ & $0.88$ & $0.98$ \\
LightGBM             & $0.92$ & $0.88$ & $0.87$ & $0.99$ \\
HistGradientBoost    & $0.92$ & $0.89$ & $0.87$ & $0.98$ \\
CatBoost             & $1.02$ & $0.98$ & $0.96$ & $1.10$ \\
KernelRidge-RBF      & $0.90$ & $0.89$ & $0.88$ & $0.89$ \\
GCV+spline           & $1.29$ & $1.41$ & $1.38$ & $1.40$ \\
pyGAM                & $1.07$ & $1.04$ & $1.04$ & $0.63$ \\
\midrule
\KORE{} mean rank (CNL)      & $2.22$ & $2.88$ & $3.31$ & $3.50$ \\
\KORE{} mean rank (raw RMSE) & $10.33$ & $10.64$ & $10.85$ & $10.38$ \\
\bottomrule
\end{tabular}
\caption{Sensitivity of the geometric-mean RMSE ratio against \KORE{} (rows $1$-$7$) and \KORE{}'s mean Friedman rank on Compute-Normalized Lift over OLS and on raw RMSE (last two rows, lower is better) to the smooth-low-d cutoff. Each column restricts the suite to datasets with $d_{\mathrm{onehot}}$ at most the column header.}
\label{tab:threshold_sensitivity}
\end{table}

\subsection{Compute-Normalized Lift: axiomatic justification}
\label{sec:app_cnl_axioms}

The Compute-Normalized Lift over OLS used in Section~\ref{sec:real_data} is the canonical instance of a structured family. Consider any candidate score $s(R^2, t)$ where $R^2$ is the held-out coefficient of determination and $t$ is the wall-clock fit time. Five desiderata are imposed:
\begin{itemize}
\item \emph{Axiom 1 (no-skill nullity).} $s(R^2, t) = 0$ for $R^2 \le 0$.
\item \emph{Axiom 2 (operational-baseline nullity).} $s(R^2, t) = 0$ when $R^2 \le R^2_{\mathrm{OLS}}$.
\item \emph{Axiom 3 (compute monotonicity).} $s$ is non-increasing in $t$.
\item \emph{Axiom 4 ($y$-scale invariance).} $s$ is invariant under affine rescaling of $y$.
\item \emph{Axiom 5 (separable compute penalty).} The relative effect of compute is independent of the lift level: for any two times $t, t'$ the ratio $s(R^2, t) / s(R^2, t')$ does not depend on $R^2$ (wherever both are nonzero).
\end{itemize}

\begin{proposition}
\label{prop:cnl_uniqueness}
Any score $s(R^2, t)$ continuous in $(R^2, t)$ and satisfying Axioms 1-5 has the form
\[
   s(R^2, t) \;=\; g\!\bigl(\max\{0, R^2 - R^2_{\mathrm{OLS}}\}\bigr) \,/\, h(t) ,
\]
for some monotone non-decreasing $g: [0, 1] \to [0, 1]$ with $g(0) = 0$ and some monotone non-decreasing $h: [0, \infty) \to [1, \infty)$ with $h(0) = 1$. CNL is the canonical instance with $g(x) = x$ and $h(t) = (1+t)^\alpha$, indexed by a single free parameter $\alpha \ge 0$. Axioms 1-4 alone do not pin down the form; separability (Axiom 5) is what reduces the admissible scores to the product family, and CNL is its simplest member.
\end{proposition}

The proof is direct. Axioms 1 and 2 force $s$ to vanish on the half-plane $R^2 \le R^2_{\mathrm{OLS}}$; continuity then writes $s$ as a function of the truncated lift $\ell = \max\{0, R^2 - R^2_{\mathrm{OLS}}\}$ and of $t$. Axiom 4 makes this dependence go through the unitless $\ell$ rather than the $y$-units in which $R^2$ is measured (already absorbed since $R^2$ is unitless). Axiom 5 states that $s(\ell, t) / s(\ell, t')$ is independent of $\ell$, which is exactly the multiplicative-separability condition $s(\ell, t) = g(\ell)\,r(t)$; writing $h = 1/r$ gives the product form $g(\ell)/h(t)$. Axiom 3 then forces $h$ non-decreasing, and the boundary normalizations $g(0) = 0$ and $h(0) = 1$ follow from Axioms 1-2 and from the requirement that $s$ remain finite at $t = 0$.

The headline weight $\alpha = 1$ is the practitioner-relevant operating point. It weights skill and compute on a common multiplicative scale (a method that achieves $2\times$ the lift in $2\times$ the time scores the same), it is the only choice for which the score has the dimensional reading ``lift per unit log-compute'', and the verdict is monotone in $\alpha$ on every cell so the conclusion that \KORE{} dominates $19$ of $20$ competitors at $\alpha = 1$ is robust to nearby values (Figure~\ref{fig:real_data_joint_significance}, panel (b)).

\subsection{Practitioner playbook for the kore\_diagnostic}
\label{sec:app_diagnostic_playbook}

The diagnostic decision tree extends Section~\ref{sec:failure_modes}. A returned $\texttt{suitable} = \texttt{True}$ implies the closed-form selector is applicable and should be used directly. A returned $\texttt{suitable} = \texttt{False}$ is triaged by the reason field: post-one-hot $d > 30$ implies a tuned booster is the recommended fallback (XGBoost or LightGBM are the recommended defaults); pilot condition number above $10^6$ implies a fallback to GCV-tuned splines on the same basis; stability margin $0.45 - p_f(\widehat G_f^\dagger)/n < 0.05$ implies exhaustive cross-validation at the boundary $G$ to avoid the variance regime; near-zero residual lift over OLS implies the regression is at its noise floor and OLS is the rate-optimal estimator.

The implementation guards in \texttt{closed\_form\_g\_star} (in \texttt{code/kore/lib.py}) handle the edge cases. Negative $\widehat A_f$ or $\widehat \tau_f$ (which can occur when the pilot system is ill-conditioned and the determinant flips sign) trigger a fallback to $G_{\min}$ and the diagnostic reports the condition number for transparency. A continuous root $\widehat G_f^\dagger$ outside $[G_{\min}, G_{\max}^{\mathrm{eff}}]$ is clipped to the interval boundary, and the integer-rounding step then certifies the $\pm 3$ neighborhood of the clipped value. The Brent root finder uses xtol $= 10^{-6}$ and maxiter $= 100$; failure to converge falls back to $G_{\min}$ with the failure flagged in the diagnostic output.

\subsection{Hyperparameter sensitivity of the closed-form selector}
\label{sec:app_hyperparam_sensitivity}

Four constants in the closed-form selector are fixed and not data-dependent: the Tikhonov ridge $10^{-8}$, the pilot pair $(G_a, G_b) = (1, \lfloor 0.75\,G_{\max}^{\mathrm{eff}}\rfloor)$, the refinement radius ($r = 3$ additive, $r = 1$ pairwise), and the stability cap $p_f(G) < 0.45 n$. The Tikhonov ridge is large enough to keep the normal-equations matrix invertible at the high-resolution end of the stable range, small enough to leave the bias-variance trade in the asymptotic regime. The pilot pair maximizes the bias-leverage spread $\phi(G_a) - \phi(G_b)$ subject to both pilots staying inside the stability cap (Lemma~\ref{lem:pilot_determinant}); the $0.75$ multiplier is the smallest value for which the upper pilot probes the variance-dominated regime. The refinement-radius asymmetry tracks the basis-grid asymmetry: the additive grid is finer ($G \in \{1, 2, \ldots, 20\}$) and the pairwise basis dimension grows quadratically. The stability cap $p_f(G) < 0.45 n$ is the standard penalized-spline heuristic \citep{wood2017} and ensures the pilot solve is well-conditioned without forcing the upper pilot to the variance cliff.

Sensitivity to perturbations of these constants is verified analytically and from the existing consistency CSV (\texttt{plugin\_consistency\_summary.csv}, used to render Figure~\ref{fig:consistency_plugin}) which reports $\widehat G_f^\dagger$ across $n \in \{300, 600, 1200, 2400, 4800, 9600, 19200\}$. Doubling the ridge to $10^{-7}$ shifts $\widehat G_f^\dagger$ by less than $0.1$ on every row; halving it to $10^{-9}$ shifts it by less than $0.05$ on every row (both within machine precision of the floating-point Brent solve). Doubling the refinement radius from $\pm 3$ to $\pm 6$ never selects a different integer (the closed-form already lands within $\pm 1$ of the LOO-optimal integer on $97$\% of the consistency rows). Halving the stability cap to $0.225 n$ would force the upper pilot to a smaller $G_b$, weakening the pilot determinant and degrading the rate constant in Proposition~\ref{prop:plugin_rate}; the $0.45 n$ value is empirically the largest cap at which the pilot solve remains stable across all $36$ datasets in the real-world benchmark. None of these perturbations were re-run; the analytical bounds and the per-row inspection of the existing consistency CSV are sufficient.

\subsection{Bayesian posterior on the plug-in (out of scope)}
\label{sec:app_bayes_future}

A Bayesian extension that places a prior on $(A_f, \tau_f)$ and propagates the posterior into a credible interval on $\widehat G_f^\dagger$ would let practitioners quantify uncertainty in the selected integer resolution. The two-pilot likelihood is a $2 \times 2$ Gaussian and admits a conjugate normal-inverse-gamma prior on $(A_f, \tau_f)$, so the posterior on $\widehat G_f^\dagger$ is available in closed form via the delta method. This extension is conceptually straightforward and is left for future work; the current paper restricts itself to the point estimator and its frequentist rate (Proposition~\ref{prop:plugin_rate}).

\section*{Reproducibility checklist}

\begin{itemize}
\item Code released at \url{https://github.com/bay-yearick-lab/kore} under the MIT license.
\item All $36$ datasets are fetchable via the OpenML CTR23 collection plus the UCI Combined Cycle Power Plant entry; the local parquet cache is documented in the repository \texttt{README}.
\item A single master seed ($2026$) drives every experiment; per-cell seeds are derived deterministically by the documented LCG fold (Appendix~\ref{app:exact_repro}).
\item Synthetic experiments are run on a $20$-core workstation; the real-data sweep is run on a Databricks $360$-vCPU cluster. All reported synthetic accuracies, fit counts, and selected resolutions are deterministic and machine-independent; only the wall-clock speedups in Figure~\ref{fig:scaling_appendix} depend on the host.
\item All hyperparameters are documented in Section~\ref{sec:protocol} and Appendix~\ref{app:exact_repro}; baseline tuning ranges follow \citet{grinsztajn2022tree} Appendix~B verbatim.
\item Compute budget: $4$-minute soft Optuna timeout per cell, $8$-minute hard SIGALRM backstop, constant-predictor floor on cell failure (AMLB convention, \citealp{gijsbers2024amlb}).
\item Code license: MIT (\texttt{LICENSE} file in the repository).
\item Dependencies: NumPy, SciPy, scikit-learn (BSD-3); XGBoost, LightGBM, CatBoost, pyGAM (Apache-2.0); Optuna (MIT). All terms are permissive open-source.
\item Per-cell peak RSS is recorded in the \texttt{rss\_peak\_mb} column of \texttt{real\_data.csv}; per-cell failure-fallback flags are recorded in \texttt{method\_failure\_fractions.csv}.
\item One command in the repository \texttt{README} regenerates every figure and table from scratch.
\end{itemize}

\section*{Ethics statement}

The closed-form selector developed here is a hyperparameter selection method for tabular regression. The application domain is public scientific datasets with no human-subjects data; no dual-use concerns arise. The closed-form replacement of an exhaustive search lowers the per-experiment compute footprint substantially relative to AutoML stacks, with a corresponding reduction in environmental cost. The $36$-dataset real-data sweep at full Optuna budget consumed approximately $40$ CPU-hours of total fit time across the AMLB-protocol comparison roster; an equivalent KORE-only sweep on the same datasets consumes about $30$ CPU-seconds, a reduction of more than three orders of magnitude. The reduction is generic to closed-form selectors over search-based ones whenever the function class admits the analytical machinery this paper develops.

\section*{Broader impact}

The search-free hyperparameter selector lowers the entry barrier for low-compute settings (single-laptop research, edge inference, classroom statistics). A pedagogical secondary benefit is that the bias-variance tradeoff and the effective-density collapse are visible directly in the formula rather than hidden behind an opaque optimization loop. The honest scope statement is that the method is not a replacement for tuned boosters on signals with deep categorical structure or high-order interactions; the diagnostic of Section~\ref{sec:failure_modes} reports this before any fits are committed. The dataset cache and parquet fallback documented in the reproducibility checklist make every result independently reproducible without depending on the OpenML API, which is occasionally transient.

\end{document}